\documentclass{article}

\usepackage[final,nonatbib]{neurips_2022}

\usepackage[utf8]{inputenc} %
\usepackage[T1]{fontenc}    %
\usepackage{hyperref}       %
\usepackage{url}            %
\usepackage{booktabs}       %
\usepackage{amsfonts}       %
\usepackage{nicefrac}       %
\usepackage{microtype}      %
\usepackage{xcolor}         %
\usepackage{amsmath}
\usepackage{amssymb}
\usepackage{microtype}
\usepackage{graphicx}
\usepackage{pgffor} %
\usepackage{xstring} %
\usepackage{todonotes}
\usepackage{wrapfig}
\usepackage{tabularx} %
\usepackage{standalone}
\usepackage{caption}
\usepackage{subcaption}
\usepackage{wasysym} %
\usepackage{amssymb} %
\usepackage{tabularx} %
\usepackage{listings} %
\usepackage{placeins}
\usepackage{afterpage}
\usepackage[toc,page,header]{appendix}
\usepackage{minitoc}

\usepackage{sidecap} %
\usepackage{etoc} %

\newcommand{\surya}[1]{{\textcolor{purple}{[\textbf{SG:} #1]}}}
\newcommand{\ari}[1]{{\textcolor{blue}{[\textbf{AM:} #1]}}}
\newcommand{\ben}[1]{{\textcolor{orange}{[\textbf{BS:} #1]}}}
\newcommand{\robert}[1]{{\textcolor{red}{[\textbf{RG:} #1]}}}
\newcommand{\shashank}[1]{{\textcolor{brown}{[\textbf{SS:} #1]}}}
\newcommand{\bo}[1]{\mbox{$\mathbf{#1}$}}

\renewcommand{\surya}[1]{}
\renewcommand{\shashank}[1]{}
\renewcommand{\ben}[1]{}
\renewcommand{\robert}[1]{}
\renewcommand{\ari}[1]{}

\title{Beyond neural scaling laws: \\ beating power law scaling via data pruning}

\author{%
   Ben Sorscher\thanks{\small{work done during an internship at Meta AI (FAIR)}}\textsuperscript{$\ast$1} \\
   \And
   Robert Geirhos\textsuperscript{$\ast$2} \\
   \And
   Shashank Shekhar\textsuperscript{3} \\
   \AND 
   Surya Ganguli\textsuperscript{1,3$\S$} \\
   \And
   Ari S. Morcos\textsuperscript{3$\S$}
   \AND
   \\ \multicolumn{1}{c}{\textsuperscript{$\ast$}{\small{equal contribution}}}
   \\ \multicolumn{1}{c}{\textsuperscript{1}{\small{Department of Applied Physics, Stanford University}}}
   \\ \multicolumn{1}{c}{\textsuperscript{2}{\small{University of T\"{u}bingen}}}
   \\ \multicolumn{1}{c}{\textsuperscript{3}{\small{Meta AI (FAIR)}}}
   \\ \multicolumn{1}{c}{\textsuperscript{$\S$}{\small{Joint senior authors}}
}}

\begin{document}
\doparttoc %
\faketableofcontents %
\part{} %

\maketitle

\begin{abstract}
Widely observed neural scaling laws, in which error falls off as a power of the training set size, model size, or both, have driven substantial performance improvements in deep learning.  However, these improvements through scaling alone require considerable costs in compute and energy. Here we focus on the scaling of error with dataset size and show how in theory we can break beyond power law scaling and potentially even reduce it to exponential scaling instead if we have access to a high-quality data pruning metric that ranks the order in which training examples should be discarded to achieve any pruned dataset size. We then test this improved scaling prediction with pruned dataset size empirically, and indeed observe better than power law scaling in practice on ResNets trained on CIFAR-10, SVHN, and ImageNet. Next, given the importance of finding high-quality pruning metrics, we perform the first large-scale benchmarking study of ten different data pruning metrics on ImageNet. We find most existing high performing metrics scale poorly to ImageNet, while the best are computationally intensive and require labels for every image. We therefore developed a new simple, cheap and scalable self-supervised pruning metric that demonstrates comparable performance to the best supervised metrics. Overall, our work suggests that the discovery of good data-pruning metrics may provide a viable path forward to substantially improved neural scaling laws, thereby reducing the resource costs of modern deep learning.
\end{abstract}

\section{Introduction}
Empirically observed neural scaling laws \cite{Hestness2017-yq,Kaplan2020-ti,Henighan2020-jf,rosenfeld2020a,Gordon2021-az,Hernandez2021-ix,Zhai2021-dl,Hoffmann2022-gw} in many domains of machine learning, including vision, language, and speech, demonstrate that test error often falls off as a power law with either the amount of training data, model size, or compute. Such power law scaling has motivated significant societal investments in data collection, compute, and associated energy consumption. However, power law scaling is extremely weak and unsustainable. For example, a drop in error from $3\%$ to $2\%$ might require an {\it order of magnitude} more data, compute, or energy. In language modeling with large transformers, a drop in cross entropy loss from about 3.4 to 2.8 nats\footnote{However, note that nats is on a logarithmic scale and and small improvements in nats can lead to large improvements in downstream tasks.} requires {\it 10 times} more training data (Fig.~1 in \cite{Kaplan2020-ti}). Also, for large vision transformers, an additional $2$ {\it billion} pre-training data points (starting from $1$ billion) leads to an accuracy gain on ImageNet of a few percentage points (Fig.~1 in \cite{Zhai2021-dl}). Here we ask whether we might be able to do better. For example, can we achieve exponential scaling instead, with a good strategy for selecting training examples? Such vastly superior scaling would mean that we could go from $3\%$ to $2\%$ error by only adding a few carefully chosen training examples, rather than collecting $10\times$ more random ones.

\begin{figure*}[t!]
\includegraphics[width=\linewidth]{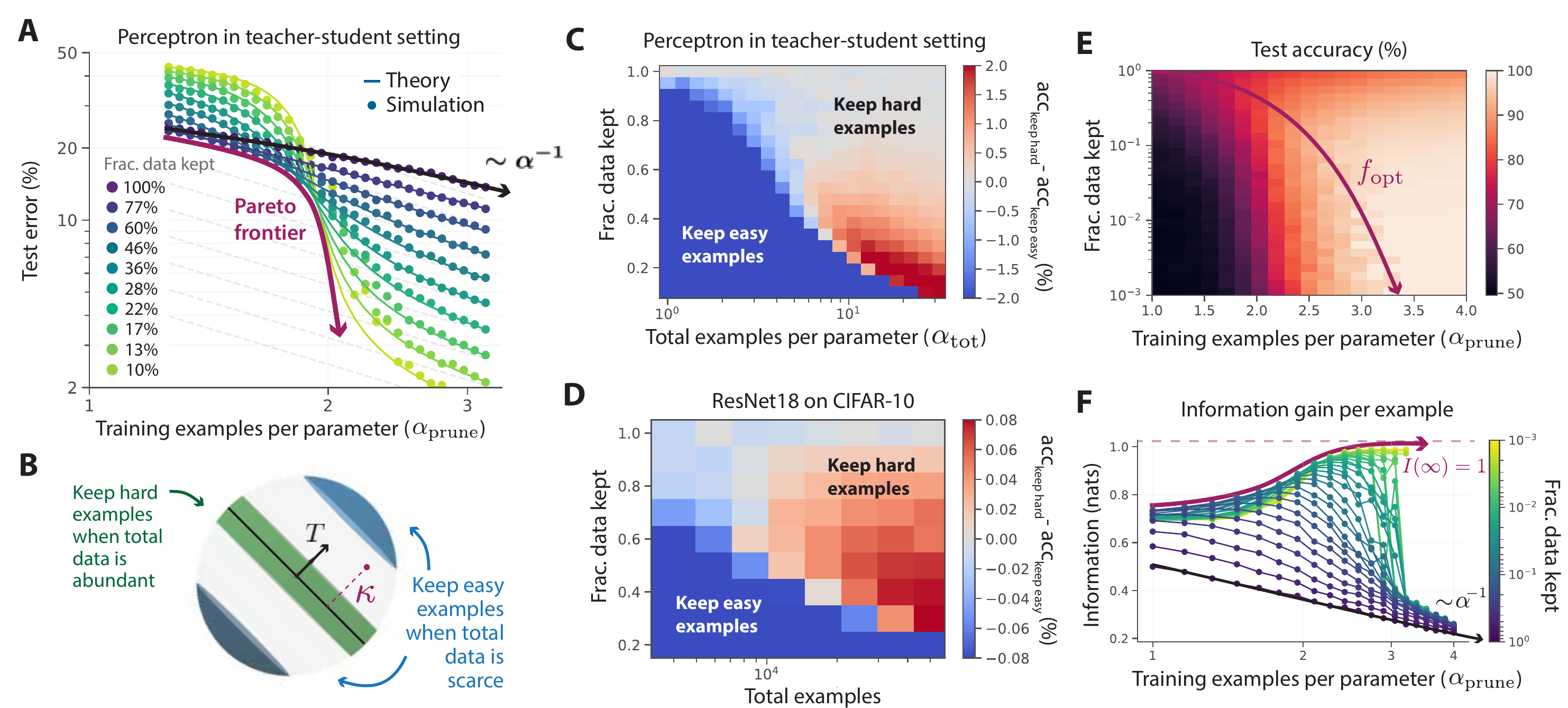}
\centering
\caption{Our analytic theory of data pruning predicts that power law scaling of test error with respect to dataset size can be beaten. \textbf{A:}~Test error as a function of $\alpha_\text{prune}=f\alpha_\text{tot}$ with $\theta=0$. We observe an excellent match between our analytic theory (solid curves) and numerical simulations (dots) of perceptron learning at parameters N=200 (here: N=200 constant throughout figure). The red curve indicates the Pareto optimal test error $\varepsilon$ achievable from a tradeoff between $\alpha_\text{tot}$ and $f$ at fixed $\alpha_\text{prune}$. \textbf{B:}~We find that when data is abundant (scarce) corresponding to  large (small) $\alpha_\text{tot}$, the better pruning strategy is to keep the hard (easy) examples. \textbf{C:}~Color indicates difference in test error in keeping hard versus easy examples, revealing the change in strategy in (B). \textbf{D:}~We tested this prediction on a ResNet18 trained on CIFAR-10, finding remarkably the same shift in optimal pruning strategy under the EL2N metric. \textbf{E:}~Test accuracy as a function of $f$ and $\alpha_\text{prune}$. For every fixed $\alpha_\text{prune}$, there is an optimal $f_\text{opt}$ (purple curve). \textbf{F:} $I(\alpha_\text{prune})$ for different $f$.} 
\label{fig:theory}
\end{figure*}

Focusing on scaling of performance with training dataset size, we demonstrate that exponential scaling is possible, both in theory and practice. The key idea is that power law scaling of error with respect to data suggests that many training examples are highly redundant. Thus one should in principle be able to prune training datasets to much smaller sizes and train on the smaller pruned datasets without sacrificing performance. Indeed some recent works \cite{Toneva2019-hj,Paul2021-ci,Chitta2021-se} have demonstrated this possibility by suggesting various metrics to sort training examples in order of their difficulty or importance, ranging from easy or redundant examples to hard or important ones, and pruning datasets by retaining some fraction of the hardest examples. However, these works leave open fundamental theoretical and empirical questions: When and why is successful data pruning possible? What are good metrics and strategies for data pruning? Can such strategies beat power law scaling?  Can they scale to ImageNet? Can we leverage large {\it unlabeled} datasets to successfully prune labeled datasets? We address these questions through both theory and experiment. Our main contributions are:

\begin{enumerate}
    \item Employing statistical mechanics, we develop a new analytic theory of data pruning in the student-teacher setting for perceptron learning, where examples are pruned based on their teacher margin, with large (small) margins corresponding to easy (hard) examples. Our theory quantitatively matches numerical experiments and reveals two striking predictions:
    \begin{enumerate}
    \item The optimal pruning strategy changes depending on the amount of initial data; with abundant (scarce) initial data, one should retain only hard (easy) examples. 
    \item Exponential scaling is possible with respect to pruned dataset size provided one chooses an increasing Pareto optimal pruning fraction as a function of initial dataset size.
    \end{enumerate}
    \item We show that the two striking predictions derived from theory hold also in practice in much more general settings. Indeed we empirically demonstrate signatures of exponential scaling of error with respect to pruned dataset size for ResNets trained from scratch on SVHN, CIFAR-10 and ImageNet, and Vision Transformers fine-tuned on CIFAR-10.
    \item Motivated by the importance of finding good quality metrics for data pruning, we perform a large scale benchmarking study of 10 different data pruning metrics at scale on ImageNet, finding that most perform poorly, with the exception of the most compute intensive metrics.
    \item We leveraged self-supervised learning (SSL) to developed a new, cheap {\it unsupervised} data pruning metric that does {\it not} require labels, unlike prior metrics. We show this unsupervised metric performs comparably to the best supervised pruning metrics that require labels and much more compute. This result opens the door to the exciting possibility of leveraging pre-trained foundation models to prune new datasets even before they are labeled. 
\end{enumerate}

Overall these results shed theoretical and empirical insights into the nature of data in deep learning and our ability to prune it, and suggest our current practice of collecting extremely large datasets may be highly inefficient. Our initial results in beating power law scaling motivate further studies and investments in not just inefficently collecting large amounts of random data, but rather, intelligently collecting much smaller amounts of carefully selected data, potentially leading to the creation and dissemination of {\it foundation datasets}, in addition to foundation models \cite{Bommasani2021-mu}. 

\section{Background and related work} \label{sec:related_work}

Our work brings together $3$ largely disparate strands of intellectual inquiry in machine learning: (1) explorations of different metrics for quantifying differences between individual training examples; (2) the empirical observation of neural scaling laws; and (3) the statistical mechanics of learning. 

\subsection{Pruning metrics: not all training examples are created equal}
Several recent works have explored various metrics for quantifying individual differences between data points. To describe these metrics in a uniform manner, we will think of all of them as ordering data points by their difficulty, ranging from ``easiest'' to ``hardest.''  When these metrics have been used for data pruning, the hardest examples are retained, while the easiest ones are pruned away. 

\paragraph{EL2N scores.} For example \cite{Paul2021-ci} trained small ensembles (of about $10$) networks for a very short time (about $10$ epochs) and computed for every training example the average $L_2$ norm of the error vector (EL2N score).
Data pruning by retaining only the hardest examples with largest error enabled training from scratch on only $50\%$ and $75\%$ of CIFAR-10 and CIFAR-100 respectively without any loss in final test accuracy. However the performance of EL2N on ImageNet has not yet been explored. 

\paragraph{Forgetting scores and classification margins.} \cite{Toneva2019-hj} noticed that over the entire course of training, some examples are learned early and never forgotten, while others can be learned and unlearned (i.e.\ forgotten) repeatedly. They developed a forgetting score which measures the degree of forgetting of each example. Intuitively examples with low (high) forgetting scores can be thought of as  easy (hard) examples. 
\cite{Toneva2019-hj} explored data pruning using these metrics, but not at ImageNet scale. 

\paragraph{Memorization and influence.} \cite{Feldman2020-yv} defined a memorization score for each example, corresponding to how much the probability of predicting the correct label for the example increases when it is present in the training set relative to when it is absent (also see \cite{harutyunyan2021estimating}); a large increase means the example must be memorized (i.e.\ the remaining training data do not suffice to correctly learn this example). Additionally \cite{Feldman2020-yv} also considered an influence score that quantifies how much adding a particular example to the training set increases the probability of the correct class label of a test example. 
Intuitively, low memorization and influence scores correspond to easy examples that are redundant with the rest of the data, while high scores correspond to hard examples that must be individually learned.
\cite{Feldman2020-yv} did not use these scores for data pruning as their computation is expensive. We note since memorization explicitly approximates the increase in test loss due to removing each individual example, it is likely to be a good pruning metric (though it does not consider interactions). 
\paragraph{Ensemble active learning.} Active learning iterates between training a model and selecting new inputs to be labeled \cite{Settles2009-mo,Bordes2005-nb,Emam2021-wa,Sener2017-on,Karamcheti2021-bs}. In contrast, we focus on data pruning: one-shot selection of a data subset sufficient to train to high accuracy from scratch. A variety of coreset algorithms (e.g.\ \cite{Mirzasoleiman2020-fy}) have been proposed for this, but their computation is expensive, and so data-pruning has been less explored at scale on ImageNet. An early clustering approach \cite{Birodkar2019-on} allowed training on $90\%$ of ImageNet without sacrificing accuracy. Notably \cite{Chitta2021-se} reduced this to $80\%$ by training a large ensemble of networks on ImageNet and using ensemble uncertainty to define the difficulty of each example, with low (high) uncertainty corresponding to easy (hard) examples.  We will show how to achieve similar pruning performance without labels or the need to train a large ensemble.

\paragraph{Diverse ensembles (DDD).} \cite{meding2022trivial} assigned a score to every ImageNet image, given by the number of models in a diverse ensemble (10 models) that misclassified the image. Intuitively, low (high) scores correspond to easy (hard) examples. The pruning performance of this metric remains unexplored. 

\paragraph{Summary.} We note: (1) only one of these metrics has tested well for its efficacy in data pruning at scale on ImageNet; (2) {\it all} of these metrics require label information; (3) there is no theory of when and why data pruning is possible for any of these metrics; and (4) none of these works suggest the possibility of exponential scaling.  We thus go beyond this prior work by benchmarking the data pruning efficacy of not only these metrics but also a new unsupervised metric we introduce that does not require label information, all at scale on ImageNet. We also develop an analytic theory for data-pruning for the margin metric that predicts not only the possibility of exponential scaling but also the novel finding that retaining easy instead of hard examples is better when data is scarce.

\subsection{Neural scaling laws and their potential inefficiency} Recent work \cite{Hestness2017-yq,Kaplan2020-ti,Henighan2020-jf,rosenfeld2020a,Gordon2021-az,Hernandez2021-ix,Zhai2021-dl,Hoffmann2022-gw} has demonstrated that test loss $\mathcal L$ often falls off as a power law with different resources like model parameters ($N$), number of training examples ($P$), and amount of compute ($C$).
However, the exponents $\nu$ of these power laws are often close to $0$, suggesting potentially inefficient use of resources. 
For example, for large models with lots of compute, so that the amount of training data constitutes a performance bottleneck, the loss scales as $\mathcal L \approx P^{-\nu}$. 
Specifically for a large transformer based language model, $\nu=0.095$, which implies {\it an order of magnitude} increase in training data drops cross-entropy loss by only about $0.6$ nats (Fig.~1 in \cite{Kaplan2020-ti}).  In neural machine translation experiments $\nu$ varies across language pairs from $0.35$ to $0.48$ (Table 1 in \cite{Gordon2021-az}). Interestingly, \cite{Hoffmann2022-gw} explored a fixed computation budget $C$ and optimized jointly over model size $N$ and training set size $P$, revealing that scaling both $N$ and $P$ commensurately as $C$ increases is compute optimal, and can yield smaller high performing models (trained on more data) than previous work.  Nevertheless, for a transformer based language model, a $100\times$ increase in compute, corresponding to $10\times$ increases in {\it both} model size and training set size, leads to a drop in cross-entropy loss of only about $0.5$ nats (Fig.~2 in \cite{Hoffmann2022-gw}). Similar slow scaling holds for large vision transformers where adding $2$ billion pre-training images reduces ImageNet performance by a few percentage points (Fig.~1 in \cite{Zhai2021-dl}). While all of these results constitute significant improvements in performance, they do come at a substantial resource cost whose fundamental origin arises from power law scaling with small exponents. Recent theoretical works \cite{JMLR:v23:20-1111,DBLP:journals/corr/abs-2102-06701,DBLP:phd/us/Rosenfeld21} have argued that the power law exponent is governed by the dimension of a data manifold from which training examples are uniformly drawn. Here we explore whether we can beat power law scaling through careful data selection.

\subsection{Statistical mechanics of perceptron learning} Statistical mechanics has long played a role in analyzing machine learning problems (see e.g.\ \cite{Engel2001-sp,Advani2013-en,Bahri2020-mi,Zdeborova2016-vk} for reviews). One of the most fundamental applications is perceptron learning in the student-teacher setting \cite{Gardner1988-wr,Seung1992-ob}, in which random i.i.d.~Gaussian inputs are labeled by a teacher perceptron to construct a training set.  The test error for another student perceptron learning from this training set then scales as a power law with exponent $-1$ for such data.  Such perceptrons have also been analyzed in an active learning setting where the learner is free to design {\it any} new input to be labeled \cite{Freund1992-uv,Zhou2019-ha}, rather than choose from a fixed set of inputs, as in data-pruning.  Recent work \cite{Cui2021-kp} has analyzed this scenario but focused on message passing algorithms that are tailored to the case of Gaussian inputs and perceptrons, and are hard to generalize to real world settings. In contrast we analyze margin based pruning algorithms that are used in practice in diverse settings, as in \cite{Toneva2019-hj,Paul2021-ci}.

\section{An analytic theory of data pruning}

To better understand data pruning, we employed the replica method from statistical mechanics \cite{Mezard1987-pc} to develop an analytic theory of pruning for the perceptron in the student-teacher setting \cite{Engel2001-sp} (see App.~\ref{app:theory} for detailed derivations of all results).
Consider a training dataset of $P$ examples $\{\bo{x}^\mu, y^\mu\}_{\mu=1,\ldots,P}$ 
where $\bo{x}^\mu \in \mathbb{R}^N$ are i.i.d.~zero mean unit variance random Gaussian inputs and 
$y^\mu = \text{sign}(\bo{T} \cdot \bo{x}^\mu)$ 
are labels generated by a teacher perceptron with weight vector $\bo{T} \in \mathbb{R}^N$.  
We work in the high dimensional statistics limit where $N,P \rightarrow \infty$ but the ratio $\alpha_\text{tot} = \frac{P}{N}$ of the number of total training examples to parameters remains $O(1)$.  
We then consider a pruning algorithm used in \cite{Toneva2019-hj,Paul2021-ci}, namely: 
(1) train a probe student perceptron for very few epochs on the training data, obtaining weights $\bo{J}_\text{probe}$; 
(2) compute the margin $m^\mu = \bo{J}_\text{probe} \cdot (y^\mu \bo{x}^\mu)$ of each training example, where large (small) margins correspond to easy (hard) examples; 
(3) construct a pruned dataset of size $P_\text{prune} = f P$, where $f$ is the fraction of examples kept, by retaining the $P_\text{prune}$ hardest examples, 
(4) train a new perceptron to completion on the smaller dataset with a smaller ratio $\alpha_\text{prune} = \frac{P_\text{prune}}{N}$ of examples to parameters.  

We are interested in the test error $\varepsilon$ of this final perceptron as a function of $\alpha_\text{tot}$, $f$, and the angle $\theta$ between the probe student $\bo{J}_\text{probe}$ and the teacher $\bo{T}$.  Our theory approximates $\bo{J}_\text{probe}$ as simply a random Gaussian vector conditioned to have angle $\theta$ with the teacher $\bo{T}$.  Under this approximation we obtain an analytic theory for $\varepsilon(\alpha_\text{tot},f,\theta)$ that is asymptotically exact in the high dimensional limit (App.~\ref{app:theory}).  We first examine results when $\theta=0$, so we are pruning training examples according to their veridical margins with respect to the teacher (Fig.~\ref{fig:theory}A). We find two striking phenomena, each of which constitute predictions in real-world settings that we will successfully confirm empirically.

\paragraph{The best pruning strategy depends on the amount of initial data.} First, we note the test error curve for $f=1$ in Fig.~\ref{fig:theory}A corresponding to no pruning, or equivalently to {\it randomly} pruning a larger dataset of size $\alpha_\text{tot}$ down to a size $\alpha_\text{prune}$, exhibits the well known classical perceptron learning power law scaling $\varepsilon \propto \alpha_\text{prune}^{-1}$.  
Interestingly though, for small $\alpha_\text{tot}$, keeping the hardest examples performs {\it worse} than random pruning (lighter curves above darkest curve for small $\alpha_\text{prune}$ in Fig.~\ref{fig:theory}A).  However, for large $\alpha_\text{tot}$, keeping the hardest examples performs {\it substantially better} than random pruning (lighter curves below darkest curve for large $\alpha_\text{prune}$ in Fig.~\ref{fig:theory}A).  It turns out keeping the {\it easiest} rather than hardest examples is a better pruning strategy when $\alpha_\text{tot}$ is small (Fig.~\ref{fig:theory}C). If one does not have much data to start with, it is better to keep the easiest examples with largest margins (i.e.\ the blue regions of Fig.~\ref{fig:theory}B) to avoid overfitting. The easiest examples provide coarse-grained information about the target function, while the hard examples provide fine-grained information about the target function which can prevent the model from learning if one starts with lots of data. In cases where overfitting is less of an issue, it is best to keep the hardest examples with smallest margin that provide more information about the teacher's decision boundary (i.e.\ the green region of Fig.~\ref{fig:theory}B). Intuitively, in the limited data regime, it is challenging to model outliers since the basics are not adequately captured; hence, it is more important to keep easy examples so that the model can get to moderate error. However, with a larger dataset, the easy examples can be learned without difficulty, making modeling outliers the fundamental challenge. 

Fig.~\ref{fig:theory}C reveals which pruning strategy is best as a joint function of $\alpha_\text{tot}$ and $f$. Note the transition between optimal strategies becomes sharper at small fractions $f$ of data kept. This transition between optimal pruning strategies can be viewed as a prediction in more general settings.  To test this prediction we trained a ResNet18 on pruned subsets of the CIFAR-10 dataset (Fig.~\ref{fig:theory}D), and observed strikingly similar behavior, indicating the prediction can hold far more generally, beyond perceptron learning. Interestingly, \cite{Toneva2019-hj,Paul2021-ci} missed this transition, likely because they started pruning from large datasets. 

\paragraph{Pareto optimal data pruning can beat power law scaling.} A second prediction of our theory is that when keeping a {\it fixed} fraction $f$ of the hardest examples as $\alpha_\text{tot}$ increases 
(i.e.\ constant color curves in Fig.~\ref{fig:theory}A), 
the error initially drops exponentially in $\alpha_\text{prune} = f \alpha_\text{tot}$, 
but then settles into the universal power law $\varepsilon \propto \alpha_\text{prune}^{-1}$ 
for all fixed $f$. 
Thus there is no asymptotic advantage to data pruning at a fixed $f$.  
However, by pruning more aggressively (smaller $f$) when given more initial data (larger $\alpha_\text{tot}$), 
one can achieve a Pareto optimal test error as a function of pruned dataset size $\alpha_\text{prune}$ 
that remarkably traces out at least an exponential scaling law (Fig.~\ref{fig:theory}A, purple curve). 
Indeed our theory predicts for each $\alpha_\text{prune}$ a Pareto optimal point in $\alpha_\text{tot}$ and $f$ (subject to $\alpha_\text{prune}= f \alpha_\text{tot}$), yielding for every fixed $\alpha_\text{prune}$ an optimal $f_\text{opt}$, plotted in Fig.~\ref{fig:theory}E. Note $f_\text{opt}$ decreases with $\alpha_\text{prune}$ indicating more aggressive pruning (smaller $f_\text{opt}$) of original datasets of larger size $\alpha_\text{tot}$ is required to obtain larger Pareto optimal pruned datasets of size $\alpha_\text{prune}$. We will test this striking scaling prediction in Fig.~\ref{fig:scaling_practice}.

\begin{figure}[t!]
\includegraphics[width=0.9\linewidth]{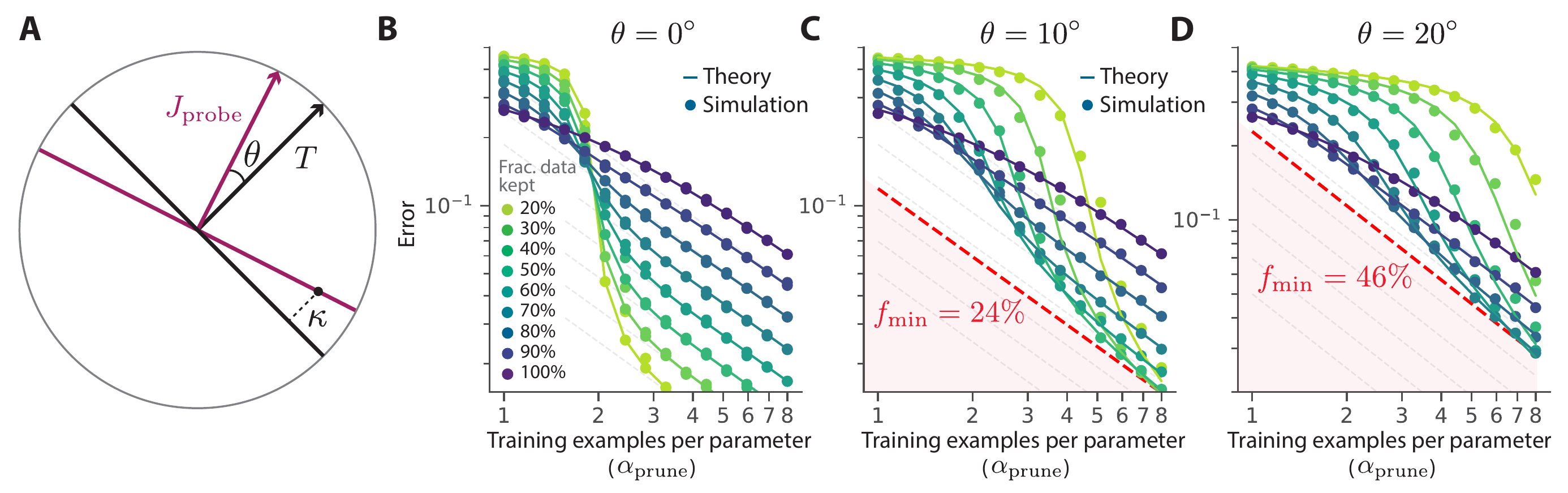}
\centering
\caption{Data pruning with an imperfect metric. \textbf{A:} Weight vectors and decision boundaries for a teacher (black) and probe student (red) separated by angle $\theta$. 
The black point has margin $0$ ($\kappa$) w.r.t. the probe (teacher). 
\textbf{B--D:} Test error as a function of $\alpha_\text{prune}$ for different $f$ and different $\theta$.}
\label{fig:teacher_student_overlap}
\end{figure}

\paragraph{Beating power law scaling: an information-theoretic perspective.} Classical randomly selected data generates slow power law error scaling because each extra training example provides less new information about the correct decision boundary than the previous example. 
More formally, let $S(\alpha_\text{tot})$ denote the typical entropy of the posterior distribution over student perceptron weights consistent with a training set of size $\alpha_\text{tot}$.  
The information gain $I(\alpha_\text{tot})$ due to additional examples beyond $\alpha_\text{tot}$ can be defined as the rate at which the posterior entropy is reduced: 
$I(\alpha_\text{tot}) = -\frac{d}{d\alpha_\text{tot}} S(\alpha_\text{tot})$. 
In classical perceptron learning $I(\alpha_\text{tot})$ decays to zero as a power law in $\alpha_\text{tot}$, 
reflecting a vanishing amount of information per each new example, leading to the slow power law decay of test error $\varepsilon \propto \alpha_\text{tot}^{-1}$. However, data pruning can increase the information gained per example by pruning away the uninformative examples. To show this, we generalized the replica calculation of the posterior entropy $S$ and information gain $I$ 
from random datasets of size $\alpha_\text{tot}$ to pruned datasets of size $\alpha_\text{prune}$ (App.~\ref{app:theory}).
We plot the resulting information gain $I(\alpha_\text{prune})$ for different $f$ in Fig.~\ref{fig:theory}F. 
For any fixed $f$, $I(\alpha_\text{prune})$ will eventually decay as a power law as $\alpha_\text{prune}^{-1}$. However, by more aggressively pruning (smaller $f$) datasets of larger size $\alpha_\text{tot}$, $I(\alpha_\text{prune})$ can converge to a finite value $I(\infty)=1$ nat/example, 
resulting in larger pruned datasets only adding useful non-redundant information.  
Since each new example under Pareto optimal data pruning conveys finite information about the target decision boundary, as seen in Fig.~\ref{fig:theory}F, the test error can decay at least exponentially in pruned dataset size as in Fig.~\ref{fig:theory}A. Classical results \cite{Seung1992-ob} have shown that training examples chosen by maximizing the disagreement of a committee of student perceptrons can provide an asymptotically finite information rate, leading to exponential decay in test error. Intriguingly, the Pareto-optimal data pruning strategy we study in this work leads to \textit{faster} than exponential decay, because it includes (partial) information about the target function provided by the probe student (Fig. \ref{fig:scaling_experiments}).

\paragraph{An imperfect pruning metric yields a cross over from exponential to power law scaling.} We next examine the case of nonzero angle $\theta$ between the probe student $\bo{J}_\text{probe}$ and the teacher $\bo{T}$, such that the ranking of training examples by margin is no longer completely accurate (Fig.~\ref{fig:teacher_student_overlap}A).  
Retaining the hard examples with smallest margin with respect to the probe student will always result in pruned datasets lying near the probe's decision boundary. But if $\theta$ is large, such examples might be far from the teacher's decision boundary, and therefore could be less informative about the teacher (Fig.~\ref{fig:teacher_student_overlap}A). 
As a result our theory, confirmed by simulations, predicts that under nonzero angles $\theta$, the Pareto optimal lower envelope of test error over both $\alpha_\text{tot}$ and $f$ initially scales exponentially as a function of $\alpha_\text{prune} = f \alpha_\text{tot}$ but then crosses over to a power law (Fig.~\ref{fig:teacher_student_overlap}BCD). Indeed, at any given nonzero $\theta$, our theory reveals that as $\alpha_\text{tot}$ (and therefore $\alpha_\text{prune}$) becomes large, one cannot decrease test error any further by retaining less than a minimum fraction $f_\text{min}(\theta)$ of all available data. For example when $\theta=10^\circ$ ($\theta=20^\circ$) one can do no better asymptotically than pruning down to 24\% ($46\%$) of the total data (Fig.~\ref{fig:teacher_student_overlap}CD). As $\theta$ approaches $0$, $f_\text{min}(\theta)$ approaches $0$, indicating that one can prune extremely aggressively to arbitrarily small $f$ while still improving performance, leading to at least exponential scaling for arbitrarily large $\alpha_\text{prune}$ in Fig.~\ref{fig:teacher_student_overlap}B.  However, for nonzero $\theta$, the lack of improvement for $f < f_\text{min}(\theta)$ at large $\alpha_\text{prune}$ renders aggressive pruning ineffective.  This result highlights the importance of finding high quality pruning metrics with $\theta \approx 0$. Such metrics can delay the cross over from exponential to power law scaling as pruned dataset size $\alpha_\text{prune}$ increases, by making aggressive pruning with very small $f$ highly effective.  Strikingly, in App.~Fig.~\ref{fig:svhn_teacher_student_overlap} we demonstrate this cross-over in a real-world setting by showing that the test error on SVHN is bounded below by a power law when the dataset is pruned by a probe ResNet18 under the EL2N metric, trained for 4 epochs (weak pruning metric) but not a probe ResNet18 trained for 40 epochs (strong pruning metric). \surya{Ben: I love this observation and connection to this theory section - if you can still add please do!}

\begin{figure}[t!]
\includegraphics[width=\linewidth]{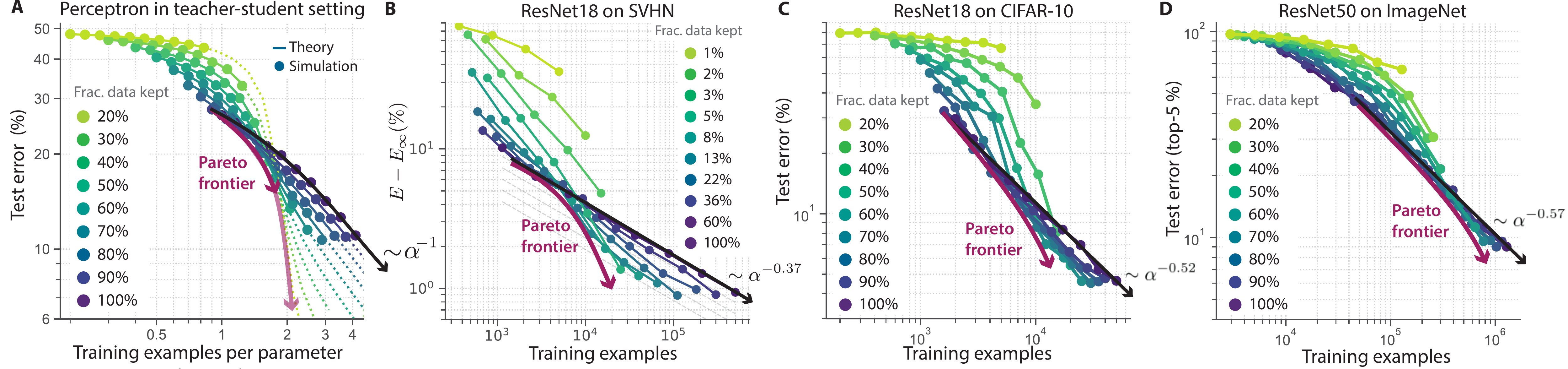}
\centering
\caption{Beating power law scaling in practice. \textbf{A--D:} Curves of test error against pruned dataset size in $4$ settings. Pruning scores were EL2N \cite{Paul2021-ci} for CIFAR-10 and SVHN and memorization \cite{Feldman2020-yv} for ImageNet. See App.~\ref{app:training_details} for all pruning/training details and App.~\ref{app:additional_scaling_experiments} for similar ImageNet plots with EL2N. Note solid curves reflect performance with a fixed total dataset size; if we prune more aggressively with even larger datasets, scaling could improve further (e.g., dashed lines in \textbf{A}). Error curves with no data pruning ($f=1$) are labeled with their best-fit power law scaling $\sim \alpha^{-\nu}$. (Note that for SVHN in \textbf{B} an asymptotic constant error $E(P \rightarrow \infty) = 1.1\%$ is subtracted from each of the curves to visualize the power law scaling more clearly.)
}
\label{fig:scaling_practice}
\end{figure}

\section{Data pruning can beat power law scaling in practice}

Our theory of data pruning for the perceptron makes three striking predictions which can be tested in more general settings, such as deep neural networks trained on benchmark datasets: (1) relative to random data pruning, keeping only the hardest examples should \textit{help} when the initial dataset size is large, but \textit{hurt} when it is small; (2) data pruning by retaining a fixed fraction $f$ of the hardest examples should yield power law scaling, with exponent equal to that of random pruning, as the initial dataset size increases; (3) the test error optimized over both initial data set size and fraction of data kept can trace out a Pareto optimal lower envelope that beats power law scaling of test error as a function of pruned dataset size, through more aggressive pruning at larger initial dataset size.  We verified all three of these predictions on ResNets trained on SVHN, CIFAR-10, and ImageNet using varying amounts of initial dataset size and fractions of data kept under data pruning (compare theory in Fig.~\ref{fig:scaling_practice}A with deep learning experiments in Fig.~\ref{fig:scaling_practice}BCD). In each experimental setting we see better than power law scaling at larger initial data set sizes and more aggressive pruning. Moreover we would likely see even better scaling with even larger initial datasets (as in Fig.\ref{fig:scaling_practice}A dashed lines).  

\begin{SCfigure}
\caption{Data pruning improves transfer learning. \textbf{A}: CIFAR-10 performance of a ViT pre-trained on all of ImageNet21K and fine-tuned on different pruned subsets of CIFAR-10 under the EL2N metric. \textbf{B:} CIFAR-10 performance of ResNet50s pre-trained on different pruned subsets of ImageNet1K and fine-tuned on all of CIFAR-10.}

\includegraphics[width=0.55\linewidth]{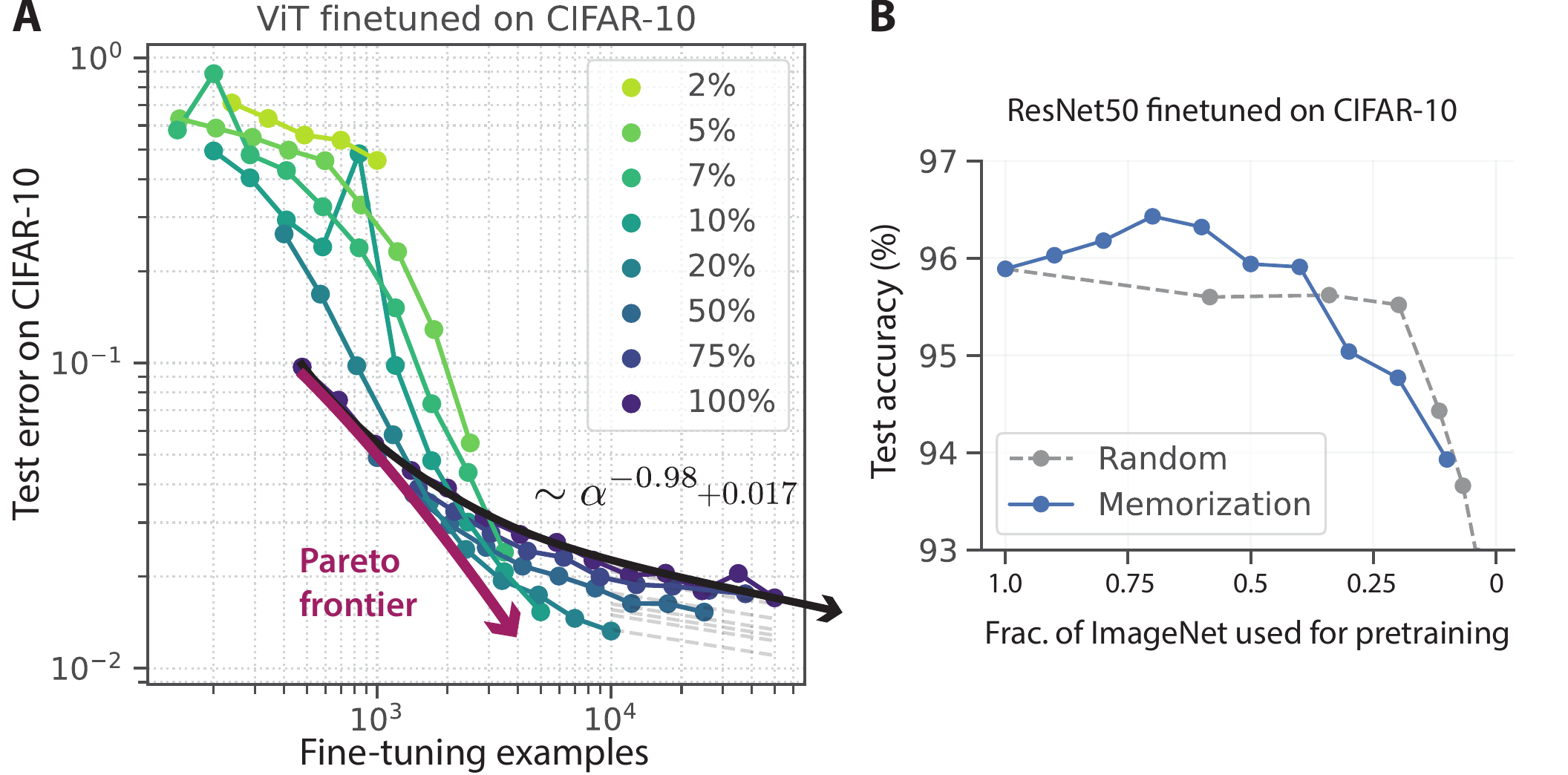}
\label{fig:transfer}
\end{SCfigure}

\paragraph{Data pruning improves transfer learning.} Modern foundation models are pre-trained on a large initial dataset, and then transferred to other downstream tasks by fine-tuning on them.  We therefore examined whether data-pruning can be effective for both reducing the amount of fine-tuning data and the amount of pre-training data. To this end, we first analyzed a vision transformer (ViT) pre-trained on ImageNet21K and then fine-tuned on different pruned subsets of CIFAR-10. %
Interestingly, pre-trained models allow for far more aggressive data pruning; fine-tuning on only 10\% of CIFAR-10 can match or exceed performance obtained by fine tuning on {\it all} of CIFAR-10 (Fig.~\ref{fig:transfer}A). Furthermore Fig.~\ref{fig:transfer}A provides a new example of beating power law scaling in the setting of fine-tuning.  Additionally, we examined the efficacy of pruning pre-training data by pre-training ResNet50s on different pruned subsets of ImageNet1K (exactly as in Fig.~\ref{fig:scaling_practice}D)  and then fine-tuning them on all of CIFAR-10. Fig.~\ref{fig:transfer}B demonstrates pre-training on as little as $50\%$ of ImageNet can match or exceed CIFAR-10 performance obtained by pre-training on all of ImageNet. Thus intriguingly pruning pre-training data on an upstream task can still maintain high performance on a different downstream task. Overall these results demonstrate the promise of data pruning in transfer learning for both the pre-training and fine-tuning phases.

\begin{figure*}[h!]
\includegraphics[width=\textwidth]{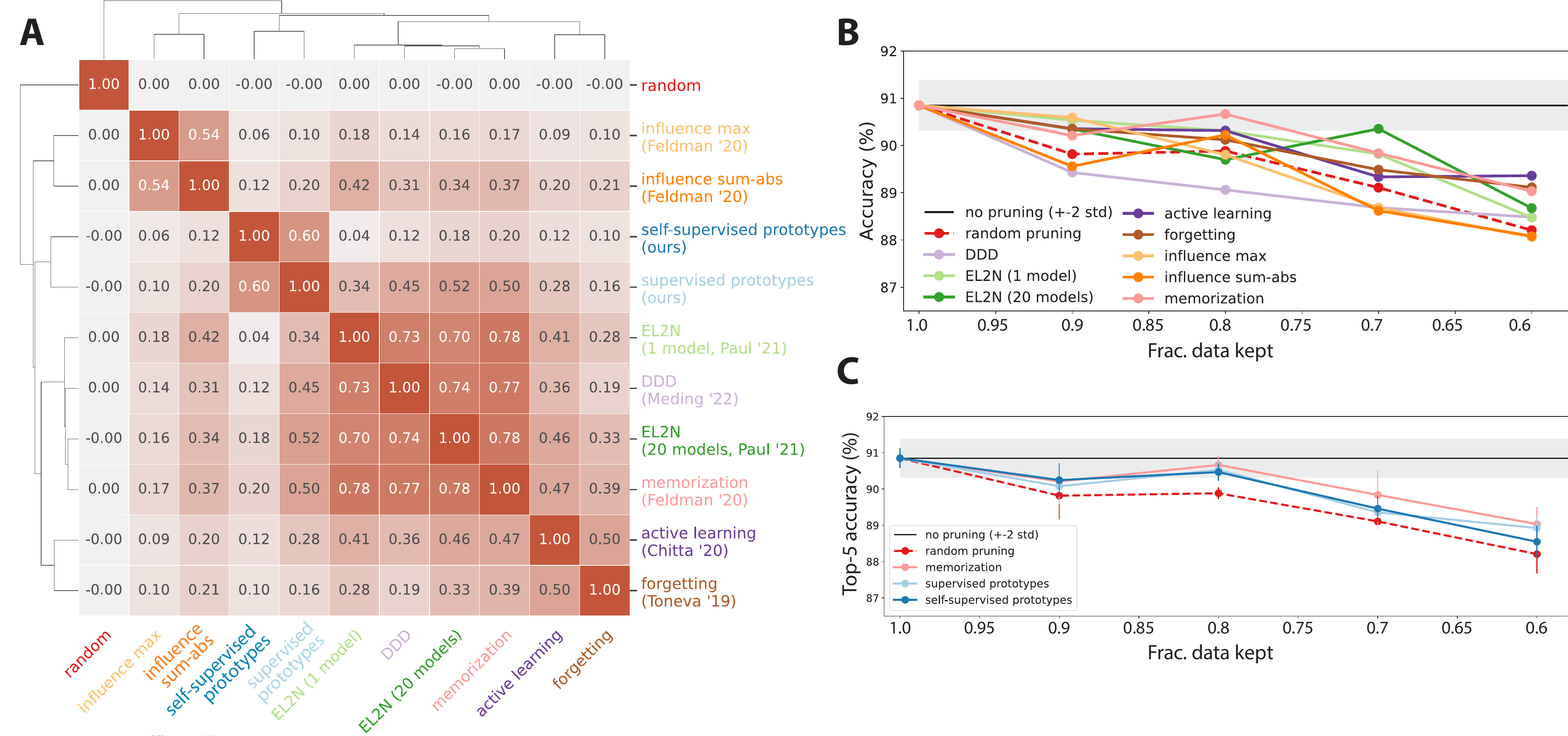}
\centering
\caption{Dataset pruning at ImageNet scale. \textbf{A:} Spearman's rank correlation between all pairs of ImageNet metric scores, along with hierarchical clustering (as provided by \texttt{seaborn.clustermap}). \textbf{B:}~Benchmarking existing supervised metrics on ImageNet (top-5 validation accuracy). \textbf{C:}~Comparing top-5 performance on ImageNet when pruning according to the best existing supervised metric (memorization) and our supervised and self-supervised prototype metrics. In all $3$ cases, training on 80\% of ImageNet approximates training on 100\%. See App.~\ref{app:training_details} for pruning and training details.}
\label{fig:ImageNet-1K_panel}
\end{figure*}

\section{Benchmarking supervised pruning metrics on ImageNet}
We note that the majority of data pruning experiments have been performed on small-scale datasets (i.e.\ variants of MNIST and CIFAR), while the few pruning metrics proposed for ImageNet have rarely been compared against baselines designed on smaller datasets. Therefore, it is currently unclear how most pruning methods scale to ImageNet and which method is best. Motivated by how strongly the quality of a pruning metric can impact performance in theory (Fig.~\ref{fig:teacher_student_overlap}), we decided to fill this knowledge gap by performing a systematic evaluation of $8$ different supervised pruning metrics on ImageNet: two variants of influence scores \cite{Feldman2020-yv}, two variants of EL2N \cite{Paul2021-ci}, DDD \cite{meding2022trivial}, memorization \cite{Feldman2020-yv}, ensemble active learning \cite{Chitta2021-se}, and forgetting \cite{Toneva2019-hj}. See Section \ref{sec:related_work} for a review of these metrics. Additionally, we include two new prototypicality metrics that we introduce in the next section.

We first asked how consistent the rankings induced by different metrics are by computing the Spearman rank correlation between each pair of metrics (Fig.~\ref{fig:ImageNet-1K_panel}A). Interestingly, we found substantial diversity across metrics, though some (EL2N, DDD, and memorization) were fairly similar with rank correlations above $0.7$. However, we observed marked performance differences between metrics: Fig~\ref{fig:ImageNet-1K_panel}BC shows test performance when a fraction $f$ of the hardest examples under each metric are kept in the training set. Despite the success of many of these metrics on smaller datasets, only a few still match performance obtained by training on the full dataset, when selecting a significantly smaller training subset (i.e.\ about $80\%$ of ImageNet). Nonetheless, most metrics continue to beat random pruning, with memorization in particular demonstrating strong performance (Fig.~\ref{fig:ImageNet-1K_panel}C). We note that data pruning on ImageNet may be more difficult than data pruning on other datasets, because ImageNet is already carefully curated to filter out uninformative examples.

We found that all pruning metrics amplify class imbalance, which results in degraded performance. To solve this we used a simple $50\%$ class balancing ratio for all ImageNet experiments. Further details and baselines without class balancing are shown in App.~\ref{app:class_imbalance}. Metric scores, including baselines, are available from \url{https://github.com/rgeirhos/dataset-pruning-metrics}.

\section{Self-supervised data pruning through a prototypicality metric}
Fig.~\ref{fig:ImageNet-1K_panel} shows many data pruning metrics do not scale well to ImageNet, while the few that do require substantial amounts of compute. Furthermore, all these metrics require labels, thereby limiting their ability to prune data for large-scale foundation models trained on massive unlabeled datasets \cite{Bommasani2021-mu}. Thus there is a clear need for simple, scalable, self-supervised pruning metrics.

To compute a self-supervised pruning metric for ImageNet, we perform $k$-means clustering in the embedding space of an ImageNet pre-trained self-supervised model (here: SWaV \cite{caron_swav}), and define the difficulty of each data point by the cosine distance to its nearest cluster centroid, or prototype. Thus easy (hard) examples are the most (least) prototypical. Encouragingly, in Fig.~\ref{fig:ImageNet-1K_panel}C, we find our self-supervised prototype metric matches or exceeds the performance of the best supervised metric, memorization, until only 70--80\% of the data is kept, despite the fact that our metric does not use labels and is much simpler and cheaper to compute than many previously proposed supervised metrics. See App.~Fig.~\ref{fig:SSL_metric_scaling} for further scaling experiments using the self-supervised metric.

\begin{figure*}[h!]
\includegraphics[width=\textwidth]{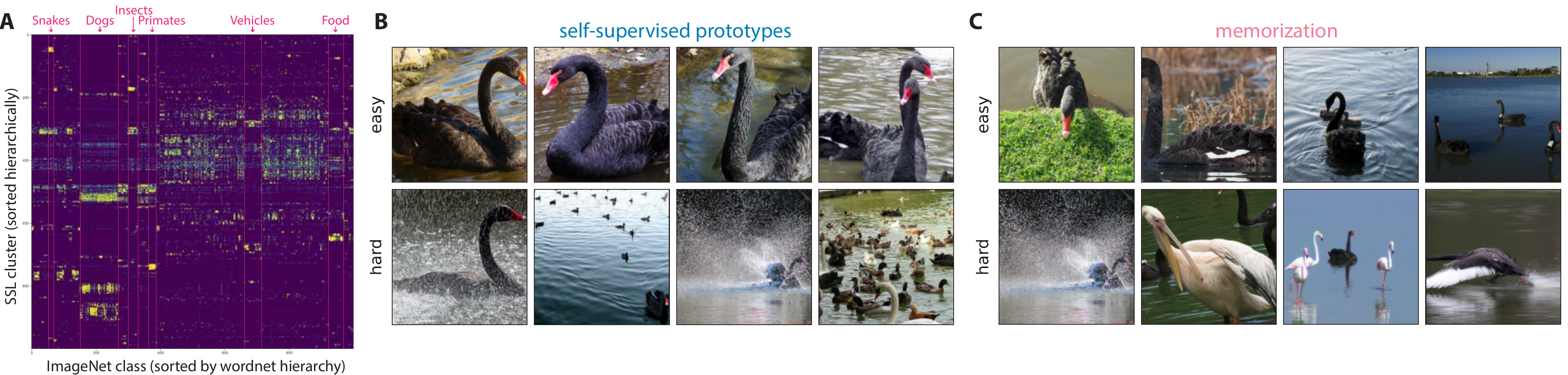}
\centering
\caption{\textbf{A:}~Heat map where each row denotes the probability that images in a given cluster come from each ImageNet class. \textbf{B:}~The four easiest and hardest images under our self-supervised pruning metric and the best previously published supervised metric (memorization, shown in \textbf{C}) for ImageNet class 100 (\texttt{black swan}).}
\label{fig:qualitative_prototypical_images}
\end{figure*}

To assess whether the clusters found by our metric align with ImageNet classes, we compared their overlaps in Fig.~\ref{fig:qualitative_prototypical_images}A. Interestingly, we found alignment for some but not all classes. For example, class categories such as snakes were largely aligned to a small number of unsupervised clusters, while other classes were dispersed across many such clusters. If class information is available, we can enforce alignment between clusters and classes by simply computing a single prototype for each class (by averaging the embeddings of all examples of this class). While originally intended to be an additional baseline metric (called supervised prototypes, light blue in Fig~\ref{fig:ImageNet-1K_panel}C), this metric remarkably outperforms other supervised metrics and largely matches the performance of memorization, which is prohibitively expensive to compute. Moreover, the performance of the best self-supervised and supervised metrics are similar, demonstrating the promise of self-supervised pruning.

One important choice for the self-supervised prototype metric is the number of clusters $k$. We found, reassuringly, our results were robust to this choice: $k$ can deviate one order of magnitude more or less than the true number of classes (i.e.\ $1000$ for ImageNet) without affecting performance (App.~\ref{app:clusters}). 

To better understand example difficulty under various metrics, we visualize extremal images for our self-supervised prototype metric and the memorization metric for one class (Fig~\ref{fig:qualitative_prototypical_images}B,C). Qualitatively, easy examples correspond to highly similar, redundant images, while hard examples look like idiosyncratic outliers. 
See App.~\ref{app:extreme_images}, Figs.~\ref{fig:extreme_images_class_100_I},\ref{fig:extreme_images_class_100_II},\ref{fig:extreme_images_class_200_I},\ref{fig:extreme_images_class_200_II},\ref{fig:extreme_images_class_300_I},\ref{fig:extreme_images_class_300_II},\ref{fig:extreme_images_class_500_I},\ref{fig:extreme_images_class_500_II} for more classes and metrics.  %

\section{Discussion}
\paragraph{Summary.}

We have shown, both in theory and practice, how to break beyond slow power law scaling of error versus dataset size to faster exponential scaling, through data pruning. Additionally we have developed a simple self-supervised pruning metric that enables us to discard 20\% of ImageNet without sacrificing performance, on par with the best and most compute intensive supervised metric.  

\paragraph{Limitations.}
\label{discussion:limitations}
The most notable limitation is that achieving exponential scaling requires a high quality data pruning metric. Since most metrics developed for smaller datasets scale poorly to ImageNet, our results emphasize the importance of future work in identifying high quality, scalable metrics. Our self-supervised metric provides a strong initial baseline. Moreover, a key advantage of data pruning is reduced computational cost due to training on a smaller dataset for the same number of epochs as the full dataset (see App. \ref{app:compute_scaling}). However, we found that performance often increased when training on the pruned dataset for the same number of  {\it iterations} as on the full dataset, resulting in the same training time, but additional training epochs. However, this performance gain saturated {\it before} training time on the pruned dataset approached that on the whole dataset (App.~\ref{app:interaction_duration}) thereby still yielding a computational efficiency gain. Overall this tradeoff between accuracy and training time on pruned data is important to consider in evaluating potential gains due to data pruning. Finally, we found that class-balancing was essential to maintain performance on data subsets (App.~\ref{app:class_imbalance}). Future work will be required to identify ways to effectively select the appropriate amount of class-balancing.

\paragraph{Ethical considerations.} A potential negative societal impact could be that data-pruning leads to unfair outcomes for certain groups.  We have done a preliminary analysis of how data-pruning affects performance on individual ImageNet classes (App.~\ref{app:fairness}), finding no substantial differential effects across classes. However proper fairness tests specific to deployment settings should always be conducted on every model, whether trained on pruned data or not. Additionally, we analyzed the impact of pruning on OOD performance (App.~\ref{app:OOD_analysis}).
\paragraph{Outlook: Towards foundation datasets.} We believe the most promising future direction is the further development of scalable, unsupervised data pruning metrics. Indeed our theory predicts that the application of pruning metrics on larger scale datasets should yield larger gains by allowing more aggressive pruning. This makes data pruning especially exciting for use on the massive unlabeled datasets used to train large foundation models (e.g.\ 400M image-text pairs for CLIP \cite{radford2021learning}, 3.5B Instagram images \cite{Mahajan2018-qf}, 
650M images for the DALLE-2 encoder \cite{ramesh2022hierarchical}, 780B tokens for PALM \cite{chowdhery2022palm}). If highly pruned versions of these datasets can be used to train a large number of different models, one can conceive of such carefully chosen data subsets as {\it foundation datasets} in which the initial computational cost of data pruning can be amortized across efficiency gains in training many downstream models, just at the initial computational cost of training foundation models is amortized across the efficiency gains of fine-tuning across many downstream tasks. Together, our results demonstrate the promise and potential of data pruning for large-scale training and pretraining. 

\subsubsection*{Acknowledgments}
We thank Priya Goyal, Berfin Simsek, Pascal Vincent valuable discussions, Qing Jin for insights about the optimal pruning distribution, Isaac Seessel for VISSL support as well as Kashyap Chitta and José Álvarez for kindly providing their ensemble active learning score.

\bibliography{biblio}

\begin{thebibliography}{10}

\bibitem{Hestness2017-yq}
Joel Hestness, Sharan Narang, Newsha Ardalani, Gregory Diamos, Heewoo Jun,
  Hassan Kianinejad, Md~Patwary, Mostofa Ali, Yang Yang, and Yanqi Zhou.
\newblock Deep learning scaling is predictable, empirically.
\newblock {\em arXiv preprint arXiv:1712.00409}, 2017.

\bibitem{Kaplan2020-ti}
Jared Kaplan, Sam McCandlish, Tom Henighan, Tom~B Brown, Benjamin Chess, Rewon
  Child, Scott Gray, Alec Radford, Jeffrey Wu, and Dario Amodei.
\newblock Scaling laws for neural language models.
\newblock {\em arXiv preprint arXiv:2001.08361}, 2020.

\bibitem{Henighan2020-jf}
Tom Henighan, Jared Kaplan, Mor Katz, Mark Chen, Christopher Hesse, Jacob
  Jackson, Heewoo Jun, Tom~B Brown, Prafulla Dhariwal, Scott Gray, et~al.
\newblock Scaling laws for autoregressive generative modeling.
\newblock {\em arXiv preprint arXiv:2010.14701}, 2020.

\bibitem{rosenfeld2020a}
Jonathan~S. Rosenfeld, Amir Rosenfeld, Yonatan Belinkov, and Nir Shavit.
\newblock A constructive prediction of the generalization error across scales.
\newblock {\em International Conference on Learning Representations}, 2020.

\bibitem{Gordon2021-az}
Mitchell~A Gordon, Kevin Duh, and Jared Kaplan.
\newblock Data and parameter scaling laws for neural machine translation.
\newblock In {\em Proceedings of the 2021 Conference on Empirical Methods in
  Natural Language Processing}, pages 5915--5922, Online and Punta Cana,
  Dominican Republic, November 2021. Association for Computational Linguistics.

\bibitem{Hernandez2021-ix}
Danny Hernandez, Jared Kaplan, Tom Henighan, and Sam McCandlish.
\newblock Scaling laws for transfer.
\newblock {\em arXiv preprint arXiv:2102.01293}, 2021.

\bibitem{Zhai2021-dl}
Xiaohua Zhai, Alexander Kolesnikov, Neil Houlsby, and Lucas Beyer.
\newblock Scaling vision transformers.
\newblock {\em arXiv preprint arXiv:2106.04560}, 2021.

\bibitem{Hoffmann2022-gw}
Jordan Hoffmann, Sebastian Borgeaud, Arthur Mensch, Elena Buchatskaya, Trevor
  Cai, Eliza Rutherford, Diego de~Las Casas, Lisa~Anne Hendricks, Johannes
  Welbl, Aidan Clark, et~al.
\newblock Training compute-optimal large language models.
\newblock {\em arXiv preprint arXiv:2203.15556}, 2022.

\bibitem{Toneva2019-hj}
Mariya Toneva, Alessandro Sordoni, Remi~Tachet des Combes, Adam Trischler,
  Yoshua Bengio, and Geoffrey~J Gordon.
\newblock An empirical study of example forgetting during deep neural network
  learning.
\newblock In {\em {ICLR}}, 2019.

\bibitem{Paul2021-ci}
Mansheej Paul, Surya Ganguli, and Gintare~Karolina Dziugaite.
\newblock Deep learning on a data diet: Finding important examples early in
  training.
\newblock {\em Adv. Neural Inf. Process. Syst.}, 34, December 2021.

\bibitem{Chitta2021-se}
Kashyap Chitta, Jos{\'e}~M {\'A}lvarez, Elmar Haussmann, and Cl{\'e}ment
  Farabet.
\newblock Training data subset search with ensemble active learning.
\newblock {\em IEEE Trans. Intell. Transp. Syst.}, pages 1--12, 2021.

\bibitem{Bommasani2021-mu}
Rishi Bommasani, Drew~A Hudson, Ehsan Adeli, Russ Altman, Simran Arora, Sydney
  von Arx, Michael~S Bernstein, Jeannette Bohg, Antoine Bosselut, Emma
  Brunskill, et~al.
\newblock On the opportunities and risks of foundation models.
\newblock {\em arXiv preprint arXiv:2108.07258}, 2021.

\bibitem{Feldman2020-yv}
Vitaly Feldman and Chiyuan Zhang.
\newblock What neural networks memorize and why: Discovering the long tail via
  influence estimation.
\newblock {\em Adv. Neural Inf. Process. Syst.}, 33:2881--2891, 2020.

\bibitem{harutyunyan2021estimating}
Hrayr Harutyunyan, Alessandro Achille, Giovanni Paolini, Orchid Majumder,
  Avinash Ravichandran, Rahul Bhotika, and Stefano Soatto.
\newblock Estimating informativeness of samples with smooth unique information.
\newblock In {\em International Conference on Learning Representations}, 2021.

\bibitem{Settles2009-mo}
Burr Settles.
\newblock Active learning literature survey.
\newblock {\em Technical Report}, 2009.

\bibitem{Bordes2005-nb}
Antoine Bordes, Seyda Ertekin, Jason Weston, L{\'e}on Botton, and Nello
  Cristianini.
\newblock Fast kernel classifiers with online and active learning.
\newblock {\em J. Mach. Learn. Res.}, 6(9), 2005.

\bibitem{Emam2021-wa}
Zeyad Ali~Sami Emam, Hong-Min Chu, Ping-Yeh Chiang, Wojciech Czaja, Richard
  Leapman, Micah Goldblum, and Tom Goldstein.
\newblock Active learning at the {ImageNet} scale.
\newblock {\em arXiv preprint arXiv:2111.12880}, 2021.

\bibitem{Sener2017-on}
Ozan Sener and Silvio Savarese.
\newblock Active learning for convolutional neural networks: A core-set
  approach.
\newblock {\em arXiv preprint arXiv:1708.00489}, 2017.

\bibitem{Karamcheti2021-bs}
Siddharth Karamcheti, Ranjay Krishna, Li~Fei-Fei, and Christopher~D Manning.
\newblock Mind your outliers! {Investigating} the negative impact of outliers
  on active learning for visual question answering.
\newblock {\em arXiv preprint arXiv:2107.02331}, 2021.

\bibitem{Mirzasoleiman2020-fy}
Baharan Mirzasoleiman, Jeff Bilmes, and Jure Leskovec.
\newblock Coresets for data-efficient training of machine learning models.
\newblock In Hal~Daum{\'e} Iii and Aarti Singh, editors, {\em Proceedings of
  the 37th International Conference on Machine Learning}, volume 119 of {\em
  Proceedings of Machine Learning Research}, pages 6950--6960. PMLR, 2020.

\bibitem{Birodkar2019-on}
V~Birodkar, H~Mobahi, and S~Bengio.
\newblock Semantic redundancies in {Image-Classification} datasets: The 10\%
  you don't need.
\newblock {\em arXiv preprint arXiv:1901.11409}, 2019.

\bibitem{meding2022trivial}
Kristof Meding, Luca M.~Schulze Buschoff, Robert Geirhos, and Felix~A.
  Wichmann.
\newblock Trivial or impossible---dichotomous data difficulty masks model
  differences (on {ImageNet} and beyond).
\newblock In {\em International Conference on Learning Representations}, 2022.

\bibitem{JMLR:v23:20-1111}
Utkarsh Sharma and Jared Kaplan.
\newblock Scaling laws from the data manifold dimension.
\newblock {\em Journal of Machine Learning Research}, 23(9):1--34, 2022.

\bibitem{DBLP:journals/corr/abs-2102-06701}
Yasaman Bahri, Ethan Dyer, Jared Kaplan, Jaehoon Lee, and Utkarsh Sharma.
\newblock Explaining neural scaling laws.
\newblock {\em CoRR}, abs/2102.06701, 2021.

\bibitem{DBLP:phd/us/Rosenfeld21}
Jonathan~S. Rosenfeld.
\newblock {\em Scaling Laws for Deep Learning}.
\newblock PhD thesis, Massachusetts Institute of Technology, {USA}, 2021.

\bibitem{Engel2001-sp}
A~Engel and C~V den Broeck.
\newblock {\em Statistical Mechanics of Learning}.
\newblock Cambridge Univ. Press, 2001.

\bibitem{Advani2013-en}
Madhu Advani, Subhaneil Lahiri, and Surya Ganguli.
\newblock Statistical mechanics of complex neural systems and high dimensional
  data.
\newblock {\em J. Stat. Mech: Theory Exp.}, 2013(03):P03014, 2013.

\bibitem{Bahri2020-mi}
Yasaman Bahri, Jonathan Kadmon, Jeffrey Pennington, Sam~S Schoenholz, Jascha
  Sohl-Dickstein, and Surya Ganguli.
\newblock Statistical mechanics of deep learning.
\newblock {\em Annual Review of Condensed Matter Physics}, March 2020.

\bibitem{Zdeborova2016-vk}
Lenka Zdeborov{\'a} and Florent Krzakala.
\newblock Statistical physics of inference: thresholds and algorithms.
\newblock {\em Adv. Phys.}, 65(5):453--552, September 2016.

\bibitem{Gardner1988-wr}
E~Gardner.
\newblock The space of interactions in neural network models.
\newblock {\em J. of Physics A}, 21:257--270, 1988.

\bibitem{Seung1992-ob}
H~S Seung, H~Sompolinsky, and N~Tishby.
\newblock Statistical mechanics of learning from examples.
\newblock {\em Phys. Rev. A}, 45(8):6056, 1992.

\bibitem{Freund1992-uv}
Yoav Freund, H~Sebastian Seung, Eli Shamir, and Naftali Tishby.
\newblock Information, prediction, and query by committee.
\newblock {\em Adv. Neural Inf. Process. Syst.}, 5, 1992.

\bibitem{Zhou2019-ha}
Hai-Jun Zhou.
\newblock Active online learning in the binary perceptron problem.
\newblock {\em Commun. Theor. Phys.}, 71(2):243, February 2019.

\bibitem{Cui2021-kp}
Hugo Cui, Luca Saglietti, and Lenka Zdeborov{\`a}.
\newblock Large deviations in the perceptron model and consequences for active
  learning, 2021.

\bibitem{Mezard1987-pc}
M~Mezard, G~Parisi, and M~A Virasoro.
\newblock {\em Spin glass theory and beyond}.
\newblock World scientific Singapore, 1987.

\bibitem{caron_swav}
Mathilde Caron, Ishan Misra, Julien Mairal, Priya Goyal, Piotr Bojanowski, and
  Armand Joulin.
\newblock Unsupervised learning of visual features by contrasting cluster
  assignments.
\newblock In H.~Larochelle, M.~Ranzato, R.~Hadsell, M.F. Balcan, and H.~Lin,
  editors, {\em Advances in Neural Information Processing Systems}, volume~33,
  pages 9912--9924. Curran Associates, Inc., 2020.

\bibitem{radford2021learning}
Alec Radford, Jong~Wook Kim, Chris Hallacy, Aditya Ramesh, Gabriel Goh,
  Sandhini Agarwal, Girish Sastry, Amanda Askell, Pamela Mishkin, Jack Clark,
  et~al.
\newblock Learning transferable visual models from natural language
  supervision.
\newblock In {\em International Conference on Machine Learning}, pages
  8748--8763. PMLR, 2021.

\bibitem{Mahajan2018-qf}
Dhruv Mahajan, Ross Girshick, Vignesh Ramanathan, Kaiming He, Manohar Paluri,
  Yixuan Li, Ashwin Bharambe, and Laurens Van Der~Maaten.
\newblock Exploring the limits of weakly supervised pretraining.
\newblock In {\em Proceedings of the European conference on computer vision
  ({ECCV})}, pages 181--196, 2018.

\bibitem{ramesh2022hierarchical}
Aditya Ramesh, Prafulla Dhariwal, Alex Nichol, Casey Chu, and Mark Chen.
\newblock Hierarchical text-conditional image generation with clip latents.
\newblock {\em arXiv preprint arXiv:2204.06125}, 2022.

\bibitem{chowdhery2022palm}
Aakanksha Chowdhery, Sharan Narang, Jacob Devlin, Maarten Bosma, Gaurav Mishra,
  Adam Roberts, Paul Barham, Hyung~Won Chung, Charles Sutton, Sebastian
  Gehrmann, et~al.
\newblock Palm: Scaling language modeling with pathways.
\newblock {\em arXiv preprint arXiv:2204.02311}, 2022.

\bibitem{goyal2021vissl}
Priya Goyal, Quentin Duval, Jeremy Reizenstein, Matthew Leavitt, Min Xu,
  Benjamin Lefaudeux, Mannat Singh, Vinicius Reis, Mathilde Caron, Piotr
  Bojanowski, Armand Joulin, and Ishan Misra.
\newblock {VISSL}.
\newblock \url{https://github.com/facebookresearch/vissl}, 2021.

\bibitem{russakovsky2015imagenet}
Olga Russakovsky, Jia Deng, Hao Su, Jonathan Krause, Sanjeev Satheesh, Sean Ma,
  Zhiheng Huang, Andrej Karpathy, Aditya Khosla, Michael Bernstein, et~al.
\newblock {ImageNet} large scale visual recognition challenge.
\newblock {\em {International Journal of Computer Vision}}, 115(3):211--252,
  2015.

\bibitem{yang2020towards}
Kaiyu Yang, Klint Qinami, Li~Fei-Fei, Jia Deng, and Olga Russakovsky.
\newblock Towards fairer datasets: Filtering and balancing the distribution of
  the people subtree in the {ImageNet} hierarchy.
\newblock In {\em Proceedings of the 2020 Conference on Fairness,
  Accountability, and Transparency}, pages 547--558, 2020.

\bibitem{asano2021pass}
Yuki~M Asano, Christian Rupprecht, Andrew Zisserman, and Andrea Vedaldi.
\newblock {PASS}: An {ImageNet} replacement for self-supervised pretraining
  without humans.
\newblock {\em arXiv preprint arXiv:2109.13228}, 2021.

\bibitem{rw2019timm}
Ross Wightman.
\newblock Pytorch image models.
\newblock \url{https://github.com/rwightman/pytorch-image-models}, 2019.

\bibitem{johnson2019survey}
Justin~M Johnson and Taghi~M Khoshgoftaar.
\newblock Survey on deep learning with class imbalance.
\newblock {\em Journal of Big Data}, 6(1):1--54, 2019.

\bibitem{geirhos2021partial}
Robert Geirhos, Kantharaju Narayanappa, Benjamin Mitzkus, Tizian Thieringer,
  Matthias Bethge, Felix~A Wichmann, and Wieland Brendel.
\newblock Partial success in closing the gap between human and machine vision.
\newblock {\em Advances in Neural Information Processing Systems},
  34:23885--23899, 2021.

\bibitem{wichmann2017methods}
Felix~A Wichmann, David~HJ Janssen, Robert Geirhos, Guillermo Aguilar, Heiko~H
  Sch{\"u}tt, Marianne Maertens, and Matthias Bethge.
\newblock Methods and measurements to compare men against machines.
\newblock {\em Electronic Imaging, Human Vision and Electronic Imaging},
  2017(14):36--45, 2017.

\bibitem{geirhos2018generalisation}
Robert Geirhos, Carlos~RM Temme, Jonas Rauber, Heiko~H Sch{\"u}tt, Matthias
  Bethge, and Felix~A Wichmann.
\newblock Generalisation in humans and deep neural networks.
\newblock In {\em {Advances in Neural Information Processing Systems}}, 2018.

\bibitem{geirhos2019imagenettrained}
Robert Geirhos, Patricia Rubisch, Claudio Michaelis, Matthias Bethge, Felix~A.
  Wichmann, and Wieland Brendel.
\newblock {ImageNet}-trained {CNN}s are biased towards texture; increasing
  shape bias improves accuracy and robustness.
\newblock In {\em {International Conference on Learning Representations}},
  2019.

\bibitem{wang2019learning}
Haohan Wang, Songwei Ge, Zachary Lipton, and Eric~P Xing.
\newblock Learning robust global representations by penalizing local predictive
  power.
\newblock {\em Advances in Neural Information Processing Systems}, 32, 2019.

\bibitem{geirhos2020beyond}
Robert Geirhos, Kristof Meding, and Felix~A Wichmann.
\newblock Beyond accuracy: quantifying trial-by-trial behaviour of {CNNs} and
  humans by measuring error consistency.
\newblock {\em {Advances in Neural Information Processing Systems}}, 33, 2020.

\bibitem{he2015delving}
Kaiming He, Xiangyu Zhang, Shaoqing Ren, and Jian Sun.
\newblock Delving deep into rectifiers: Surpassing human-level performance on
  {ImageNet} classification.
\newblock In {\em Proceedings of the IEEE International Conference on Computer
  Vision}, pages 1026--1034, 2015.

\bibitem{miller2021accuracy}
John~P Miller, Rohan Taori, Aditi Raghunathan, Shiori Sagawa, Pang~Wei Koh,
  Vaishaal Shankar, Percy Liang, Yair Carmon, and Ludwig Schmidt.
\newblock Accuracy on the line: on the strong correlation between
  out-of-distribution and in-distribution generalization.
\newblock In {\em International Conference on Machine Learning}, pages
  7721--7735. PMLR, 2021.

\end{thebibliography}
\bibliographystyle{unsrt}

\newpage

\appendix

\addcontentsline{toc}{section}{Appendix} %
\part{Appendix} %
\parttoc %

\section{A theory of data-pruning for the perceptron: detailed derivations}

\label{app:theory}

All code required to reproduce the theory figures and numerical simulations throughout this paper can be run in the Colab notebook at 
\url{https://colab.research.google.com/drive/1in35C6jh7y_ynwuWLBmGOWAgmUgpl8dF?usp=sharing}.

\subsection{Problem setup}

In this section we introduce a theory for data pruning in the teacher-student
perceptron setting, using the tools of statistical mechanics. We study
the problem of classifying a dataset of $P$ examples $\{\textbf{x}^\mu,y^{\mu}\}_{\mu=1,\ldots,P}$,
where $\textbf{x}^\mu\sim\mathcal{N}(0,I_{N})$ are i.i.d.~zero mean unit
variance random Gaussian inputs, and $y^{\mu}=\text{sign}(\textbf{T}\cdot x)$
are labels generated by a teacher perceptron $\textbf{T}\in\mathbb{R}^{N}$, which
we will assume is randomly drawn from a uniform distribution on the
sphere $\textbf{T}\sim\text{Unif}(\mathbb{S}^{N-1}(\sqrt{N}))$. We work in the high-dimensional
statistics limit where $N,P\to\infty$ but the ratio $\alpha_{\text{tot}}=P/N$
remains $O(1)$. The generalization error of a perceptron trained
on such an isotropic dataset is a classical problem (see e.g.\ \cite{Engel2001-sp}). However, we are interested in the setting where the training
dataset is not isotropic, but instead has inherited some structure
due to data pruning.

In particular, consider pruning the training dataset by keeping only
the examples with the smallest margin $|z^{\mu}|=|\textbf{J}_{\text{probe}}\cdot \textbf{x}^\mu|$
along a probe student $\textbf{J}_{\text{probe}}$. The pruned dataset will
follow some distribution $p(z)$ along the direction of $\textbf{J}_{\text{probe}}$,
and remain isotropic in the nullspace of $\textbf{J}_{\text{probe}}$.
In what follows we will derive a general theory for an arbitrary data
distribution $p(z)$, and specialize to the case of small-margin pruning
only at the very end (in which case $p(z)$ will take the form of
a truncated Gaussian). We will also make no assumptions on the form
of the probe student $\textbf{J}_{\text{probe}}$ or the learning rule used
to train it; only that $\textbf{J}_{\text{probe}}$ has developed some overlap
with the teacher, quantified by the angle $\theta=\cos^{-1}\big(\frac{\textbf{J}_{\text{probe}}\cdot \textbf{T}}{\|\textbf{J}_{\text{probe}}\|_2\|\textbf{T}\|_2}\big)$ (Fig.~\ref{fig:teacher_student_overlap}\textbf{A}).

After the dataset has been pruned, we consider training a new student
$J$ from scratch on the pruned dataset. A typical training algorithm
(used in support vector machines and the solution to which SGD converges
on separable data) is to find the solution $J$ which classifies the
training data with the maximal margin $\kappa=\min_{\mu}\textbf{J}\cdot(y^{\mu}\textbf{x}^\mu)$.
Our goal is to compute the generalization error $\varepsilon_{g}$
of this student, which is simply governed by the overlap between the
student and the teacher, $\varepsilon_{g}=\cos^{-1}(R)/\pi$, where $R=\textbf{J}\cdot \textbf{T}/\|\textbf{J}\|_2 \|\textbf{T}\|_2$.

\subsection{Main result and overview}

Our main result is a set of self-consistent equations which can be solved to obtain the generalization error $\varepsilon(\alpha,p,\theta)$ for any $\alpha$ and any data distribution $p(z)$ along a probe student at any angle $\theta$ relative to the teacher. These equations take the form,

\noindent %
\noindent\fbox{\begin{minipage}[t]{\linewidth}%
\noindent 
\begin{align}
\frac{R-\rho\cos\theta}{\sin^{2}\theta}&=\frac{\alpha}{\pi\Lambda}\bigg<\int_{-\infty}^{\kappa}dt\ \exp\left(-\frac{\Delta(t,z)}{2\Lambda^{2}}\right)(\kappa-t)\bigg>_{z} \\
1-\frac{\rho^{2}+R^{2}-2\rho R\cos\theta}{\sin^{2}\theta}&=2\alpha\bigg<\int_{-\infty}^{\kappa}dt\frac{e^{-\frac{(t-\rho z)^{2}}{2(1-\rho^{2})}}}{\sqrt{2\pi}\sqrt{1-\rho^{2}}}H\bigg(\frac{\Gamma(t,z)}{\sqrt{1-\rho^{2}}\Lambda}\bigg)(\kappa-t)^{2}\bigg>_{z}  \\
\frac{\rho-R\cos\theta}{\sin^{2}\theta}&=2\alpha\bigg<\int_{-\infty}^{\kappa}dt\frac{e^{-\frac{(t-\rho z)^{2}}{2(1-\rho^{2})}}}{\sqrt{2\pi}\sqrt{1-\rho^{2}}}H\bigg(\frac{\Gamma(t,z)}{\sqrt{1-\rho^{2}}\Lambda}\bigg)\bigg(\frac{z-\rho t}{1-\rho^{2}}\bigg)(\kappa-t) \nonumber \\
& \quad \quad \quad +\frac{1}{2\pi\Lambda}\exp\left(-\frac{\Delta(t,z)}{2\Lambda^{2}}\right)\bigg(\frac{\rho R-\cos\theta}{1-\rho^{2}}\bigg)(\kappa-t)\bigg>_{z}
\end{align}

Where,
\begin{align}
\Lambda&=\sqrt{\sin^{2}\theta-R^{2}-\rho^{2}+2\rho R\cos\theta}, \\
\Gamma(t,z)&=z(\rho R-\cos\theta)-t(R-\rho\cos\theta), \\
\Delta(t,z)&=z^{2}\left(\rho^{2}+\cos^{2}\theta-2\rho R\cos\theta\right)+2tz(R\cos\theta-\rho)+t^{2}\sin^{2}\theta.
\end{align}
\end{minipage}}

Where $\langle \cdot \rangle_z$ represents an average over the pruned data distribution $p(z)$ along the probe student. For any $\alpha, p(z), \theta$, these equations can be solved for the order parameters $R,\rho,\kappa$, from which the generalization error can be easily read off as $\varepsilon_{g}=\cos^{-1}(R)/\pi$. This calculation results in the solid theory curves in Figs \ref{fig:theory},\ref{fig:teacher_student_overlap},\ref{fig:scaling_practice}, which show an excellent match to numerical simulations. In the following section we will walk through the derivation of these equations using replica theory. In Section \ref{app:information_gain} we will derive an expression for the information gained per training example, and show that with Pareto optimal data pruning this information gain can be made to converge to a finite rate, resulting in at least exponential decay in test error. In Section \ref{app:imperfect_teacher_probe_overlap}, we will show that super-exponential scaling eventually breaks down when the probe student does not match the teacher perfectly, resulting in power law scaling at at a minimum pruning fraction $f_\text{min}(\theta).$

\subsection{Replica calculation of the generalization error}

To obtain Eqs. 1,2,3, we follow the approach
of Elizabeth Gardner and compute the volume $\Omega(\textbf{x}^\mu,\textbf{T},\kappa)$
of solutions $J$ which perfectly classify the training data up to
a margin $\kappa$ (known as the Gardner volume) \cite{Gardner1988-wr,Engel2001-sp}.
As $\kappa$ grows, the volume of solutions shrinks until it reaches
a unique solution at a critical $\kappa$, the max-margin solution. The Gardner volume $\Omega$ takes the form,

\begin{equation}
\Omega(\textbf{x}^{\mu},\textbf{T},\kappa)=\int d\mu(\textbf{J})\ \prod_{\mu}\Theta\bigg(\frac{\textbf{T}\cdot\textbf{x}^{\mu}}{\sqrt{N}} \bigg(\frac{\textbf{J}\cdot\textbf{x}^{\mu}}{\sqrt{N}}-\kappa\bigg) \bigg)
\end{equation}

Because the student's decision boundary is invariant to an overall
scaling of $\textbf{J}$, we enforce normalization of $\textbf{J}$ via the measure
$d\mu(\textbf{J})$,

\begin{equation}
d\mu(\textbf{J})=\frac{1}{(2\pi e)^{N/2}}\delta(\|\textbf{J}\|^{2}-N)
\end{equation}

In the thermodynamic limit $N,P\to\infty$ the typical value of the
entropy $S(\kappa)=\langle\langle\log\Omega(\textbf{x}^\mu,\textbf{T},\kappa)\rangle\rangle$
is dominated by particular values of $R,\kappa$, where the double
angle brackets $\langle\langle\cdot\rangle\rangle$ denote a quenched
average over disorder introduced by random realizations of the training
examples $\textbf{x}^{\mu}$ and the teacher $\textbf{T}$. However,
computing this quenched average is intractable since the integral
over $\textbf{J}$ cannot be performed analytically for every individual
realization of the examples. We rely on the replica trick from statistical
physics,

\begin{equation}
\ln(x)=\lim_{n\to0}\frac{x^{n}-1}{n}
\end{equation}

Which allows us to evaluate $S(\kappa)$ in terms of easier-to-compute
powers of $\Omega$,

\begin{equation}
S(\kappa)=\langle\langle\ln\Omega(\textbf{x}^\mu,\textbf{T},\kappa)\rangle\rangle=\frac{\langle\langle\Omega^{n}(\textbf{x}^\mu,\textbf{T},\kappa)\rangle\rangle-1}{n}
\end{equation}

This reduces our problem to computing powers of $\Omega$, which for
integer $n$ can be written in terms of $\alpha=1,\ldots,n$ replicated
copies of the original system,

\begin{equation}
\Omega^{(n)}\equiv\langle\langle\Omega^{n}(\textbf{x}^{\mu},\textbf{T},\kappa)\rangle\rangle=\bigg<\bigg<\int\prod_{\alpha=1}^{n}d\mu(\textbf{J}^{\alpha})\prod_{\alpha,\mu}\Theta\bigg(\frac{\textbf{T}\cdot\textbf{x}^{\mu}}{\sqrt{N}} \bigg(\frac{\textbf{J}\cdot\textbf{x}^{\mu}}{\sqrt{N}}-\kappa\bigg) \bigg)\bigg>\bigg>
\end{equation}

We begin by introducing auxiliary variables,

\begin{equation}
\lambda_{\mu}^{\alpha}=\frac{\textbf{J}^{\alpha}\cdot\textbf{x}^{\mu}}{\sqrt{N}},\quad u_{\mu}=\frac{\textbf{T}\cdot\textbf{x}^{\mu}}{\sqrt{N}}
\end{equation}

by $\delta-$functions, to pull the dependence on $\textbf{J}$ and
$\textbf{T}$ outside of the Heaviside function,

\begin{multline}
\Omega^{(n)}=\int\prod_{\alpha=1}^{n}d\mu(\textbf{J}^\alpha)\int\prod_{\alpha,\mu}d\lambda_{\mu}^{\alpha}\int\prod_{\mu}du_{\mu}\prod_{\alpha,\mu}\Theta\big(u_{\mu}(\lambda_{\mu}^{\alpha}-\kappa)\big) \\
\times\bigg<\bigg<\delta\bigg(\lambda_{\mu}^{\alpha}-\frac{1}{\sqrt{N}}\textbf{J}^{\alpha}\cdot\textbf{x}^{\mu}\bigg)\delta\bigg(u_{\mu}-\frac{1}{\sqrt{N}}\textbf{T}\cdot\textbf{x}^{\mu}\bigg)\bigg>\bigg>
\end{multline}

Using the integral representation of the $\delta$-functions, 

\begin{align}
\Omega^{(n)}&=\int\prod_{\alpha=1}^{n}d\mu(\textbf{J}^{\alpha})\int\prod_{\alpha,\mu}\frac{d\lambda_{\mu}^{\alpha}d\hat{\lambda}_{\mu}^{\alpha}}{2\pi}\int\prod_{\mu}\frac{du_{\mu}d\hat{u}_{\mu}}{2\pi} \\
&\times\prod_{\alpha,\mu}\Theta\big(u_{\mu}(\lambda_{\mu}^{\alpha}-\kappa)\big)\exp\bigg(i\sum_{\mu,\alpha}\lambda_{\mu}^{\alpha}\hat{\lambda}_{\mu}^{\alpha}+i\sum_{\mu}u_{\mu}\hat{u}_{\mu}\bigg) \\
&\times\langle\langle\exp\bigg(-\frac{i}{\sqrt{N}}\sum_{\mu,\alpha}\hat{\lambda}_{\mu}^{\alpha}\textbf{J}^{\alpha}\cdot\textbf{x}^{\mu}-\frac{i}{\sqrt{N}}\sum_{\mu}\hat{u}_{\mu}\textbf{T}\cdot\textbf{x}^{\mu}\bigg)\rangle\rangle
\end{align}

The data obeys some distribution $p(z)$ along the direction
of $\textbf{J}_{\text{probe}}$ and is isotropic in the nullspace
of $\textbf{J}_{\text{probe}}$. Hence we can decompose a training
example $\textbf{x}^{\mu}$ as follows, $\textbf{x}^{\mu}=\textbf{J}_{\text{probe}}z^{\mu}+(I-\textbf{J}_{\text{probe}}\textbf{J}_{\text{probe}}^{T})\textbf{s}^{\mu}$,
where $z^{\mu}\sim p(z)$ and $\textbf{s}^{\mu}\sim\mathcal{N}(0,I_{N})$,

\begin{align}
\Omega^{(n)}&=\int\prod_{\alpha=1}^{n}d\mu(\textbf{J}^{\alpha})\int\prod_{\alpha,\mu}\frac{d\lambda_{\mu}^{\alpha}d\hat{\lambda}_{\mu}^{\alpha}}{2\pi}\int\prod_{\mu}\frac{du_{\mu}d\hat{u}_{\mu}}{2\pi} \\
&\times\prod_{\alpha,\mu}\Theta\big(u_{\mu}(\lambda_{\mu}^{\alpha}-\kappa)\big)\exp\bigg(i\sum_{\mu,\alpha}\lambda_{\mu}^{\alpha}\hat{\lambda}_{\mu}^{\alpha}+i\sum_{\mu}u_{\mu}\hat{u}_{\mu}\bigg) \\
&\times\bigg<\bigg<\exp\bigg(-\frac{i}{\sqrt{N}}\sum_{\mu,\alpha}\hat{\lambda}_{\mu}^{\alpha}(\textbf{J}^{\alpha}\cdot\textbf{J}_{\text{probe}}z^{\mu}+\textbf{J}_{\perp}^{\alpha}\cdot\textbf{s}^{\mu})-\frac{i}{\sqrt{N}}\sum_{\mu}\hat{u}_{\mu}(\textbf{T}\cdot\textbf{J}_{\text{probe}}z^{\mu}+\textbf{T}_{\perp}\cdot\textbf{s}^{\mu}\bigg)\bigg>\bigg>
\end{align}

Where $\textbf{J}_{\perp}=(1-\textbf{J}_{\text{probe}}\textbf{J}_{\text{probe}}^{T})\textbf{J}$
and $\textbf{T}_{\perp}=(1-\textbf{J}_{\text{probe}}\textbf{J}_{\text{probe}}^{T})\textbf{T}$.
Now we can average over the patterns $\textbf{s}^{\mu}\sim\mathcal{N}(0,I_{N})$,

\begin{equation}
\bigg<\exp\bigg(-\frac{i}{\sqrt{N}}\sum_{\mu,\alpha}\hat{\lambda}_{\mu}^{\alpha}\textbf{J}_{\perp}^{\alpha}\cdot\textbf{s}^{\mu}-\frac{i}{\sqrt{N}}\sum_{\mu}\hat{u}_{\mu}\textbf{T}_{\perp}\cdot\textbf{s}^{\mu}\bigg)\bigg>_{s^{\mu}}=\exp\bigg(-\frac{1}{2N}\|\sum_{\mu,\alpha}\hat{\lambda}_{\mu}^{\alpha}\textbf{J}_{\perp}^{\alpha}+\hat{u}_{\mu}\textbf{T}_{\perp}\|^{2}\bigg)
\end{equation}

\begin{equation}
=\exp\bigg(-\frac{1}{2N}\sum_{\mu}\bigg(\sum_{\alpha\beta}\hat{\lambda}_{\mu}^{\alpha}\hat{\lambda}_{\mu}^{\beta}\textbf{J}_{\perp}^{\alpha}\cdot\textbf{J}_{\perp}^{\beta}+2\sum_{\alpha}\hat{\lambda}_{\mu}^{\alpha}\hat{u}_{\mu}\textbf{J}_{\perp}^{\alpha}\cdot\textbf{T}_{\perp}+\hat{u}_{\mu}^{2}\|\textbf{T}_{\perp}\|^{2}\bigg)\bigg).
\end{equation}

Inserting this back into our expression for the Gardner volume,

\begin{align}
\begin{split}
\Omega^{(n)}&=\int\prod_{\alpha=1}^{n}d\mu(\textbf{J}^{\alpha})\int\prod_{\alpha,\mu}\frac{d\lambda_{\mu}^{\alpha}d\hat{\lambda}_{\mu}^{\alpha}}{2\pi}\int\prod_{\mu}\frac{du_{\mu}d\hat{u}_{\mu}}{2\pi} \\
&\times\prod_{\alpha,\mu}\Theta\big(u_{\mu}(\lambda_{\mu}^{\alpha}-\kappa)\big)\exp\bigg(i\sum_{\mu,\alpha}\lambda_{\mu}^{\alpha}\hat{\lambda}_{\mu}^{\alpha}+i\sum_{\mu}u_{\mu}\hat{u}_{\mu}\bigg) \\
&\times\bigg<\bigg<\exp\bigg[-\frac{1}{2N}\sum_{\mu}\bigg(\sum_{\alpha\beta}\hat{\lambda}_{\mu}^{\alpha}\hat{\lambda}_{\mu}^{\beta}\textbf{J}_{\perp}^{\alpha}\cdot\textbf{J}_{\perp}^{\beta}+2\sum_{\alpha}\hat{\lambda}_{\mu}^{\alpha}\hat{u}_{\mu}\textbf{J}_{\perp}^{\alpha}\cdot\textbf{T}_{\perp}+\hat{u}_{\mu}^{2}\|\textbf{T}_{\perp}\|^{2}\bigg) \\
& -i\sum_{\mu}\bigg(\sum_{\alpha}\hat{\lambda}_{\mu}^{\alpha}\textbf{J}^{\alpha}\cdot\textbf{J}_{\text{probe}}+\hat{u}_{\mu}\textbf{T}\cdot\textbf{J}_{\text{probe}}\bigg)z^{\mu}\bigg]\bigg>\bigg>_{T,z^{\mu}}
\end{split}
\end{align}

As is typical in replica calculations of this type, we now introduce
order parameters,

\begin{equation}
q^{\alpha\beta}=\frac{\textbf{J}^{\alpha}\cdot\textbf{J}^{\beta}}{N},\quad R^{\alpha}=\frac{\textbf{T}\cdot\textbf{J}^{\alpha}}{N}
\end{equation}

which will allow us to decouple the $\textbf{J}$- from the $\lambda$-$\mu$-$z$-
integrals. $q^{\alpha\beta}$ represents the overlaps between replicated
students, and $R^{\alpha}$ the overlap between each replicated student
and the teacher. However, because our problem involves the additional
role of the probe student, we must introduce an additional order parameter,

\begin{equation}
\rho^{\alpha}=\frac{\textbf{J}^{\alpha}\cdot\textbf{J}_{\text{probe}}}{N}
\end{equation}
which represents the overlap between each replicated student and the
probe student. Notice that,

\begin{equation}
\textbf{J}_{\perp}^{\alpha}\cdot\textbf{J}_{\perp}^{\beta}=\textbf{J}^{\alpha}\cdot\textbf{J}^{\beta}-\textbf{J}_{\parallel}^{\alpha}\cdot\textbf{J}_{\parallel}^{\beta}=N(q^{\alpha\beta}-\rho^{\alpha}\rho^{\beta})
\end{equation}

\begin{equation}
\textbf{J}_{\perp}^{\alpha}\cdot\textbf{T}_{\perp}=\textbf{J}^{\alpha}\cdot\textbf{T}-\textbf{J}_{\parallel}^{\alpha}\cdot\textbf{T}_{\parallel}=N(R^{\alpha}-\rho^{\alpha}\cos\theta)
\end{equation}

With this new set of order parameters in hand, we can decouple the
$\textbf{J}$ from the $\lambda-u-z-$integrals.

\begin{align}
\begin{split}
\Omega^{(n)}&=\int\prod_{\alpha<\beta}dq^{\alpha\beta}\int\prod_{\alpha}dR^{\alpha}\int\prod_{\alpha}d\rho^{\alpha} \\
&\times\int\prod_{\alpha=1}^{n}d\mu(\textbf{J}^{\alpha})\bigg<\prod_{\alpha}\delta(\textbf{T}\cdot\textbf{J}^{\alpha}-NR^{\alpha})\bigg>_{\textbf{T}}
\prod_{\alpha<\beta}\delta(\textbf{J}^{\alpha}\cdot\textbf{J}^{\beta}-Nq^{\alpha\beta})
\prod_{\alpha}\delta(\textbf{J}^{\alpha}\cdot\textbf{J}_{\text{probe}}-N\rho^{\alpha}) \\
&\times\int\prod_{\alpha,\mu}\frac{d\lambda_{\mu}^{\alpha}d\hat{\lambda}_{\mu}^{\alpha}}{2\pi}\int\prod_{\mu}\frac{du_{\mu}d\hat{u}_{\mu}}{2\pi}\prod_{\alpha,\mu}\Theta\big(u_{\mu}(\lambda_{\mu}^{\alpha}-\kappa)\big)\prod_{\mu}\exp\bigg(i\sum_{\alpha}\lambda_{\mu}^{\alpha}\hat{\lambda}_{\mu}^{\alpha}+iu_{\mu}\hat{u}_{\mu}\bigg) \\
&\times\bigg<\exp\bigg[-\frac{1}{2}\sum_{\alpha\beta}\hat{\lambda}_{\mu}^{\alpha}\hat{\lambda}_{\mu}^{\beta}(q^{\alpha\beta}-\rho^{\alpha}\rho^{\beta})-\sum_{\alpha}\hat{\lambda}_{\mu}^{\alpha}\hat{u}_{\mu}(R^{\alpha}-\rho^{\alpha}\cos\theta)-\frac{1}{2}\hat{u}_{\mu}^{2}\sin^{2}\theta \\
& -i\bigg(\sum_{\alpha}\hat{\lambda}_{\mu}^{\alpha}\rho^{\alpha}+\hat{u}_{\mu}\cos\theta\bigg)z^{\mu}\bigg]\bigg>_{z^{\mu}}
\end{split}
\end{align}

We can now perform the gaussian integral over $\hat{u}_{\mu}$,

\begin{align}
\begin{split}
\Omega^{(n)}&=\int\prod_{\alpha<\beta}dq^{\alpha\beta}\int\prod_{\alpha}dR^{\alpha}\int\prod_{\alpha}d\rho^{\alpha} \\
&\times\int\prod_{\alpha=1}^{n}d\mu(\textbf{J}^{\alpha})\bigg<\prod_{\alpha}\delta(\textbf{T}\cdot\textbf{J}^{\alpha}-NR^{\alpha})\bigg>_{\textbf{T}}
\prod_{\alpha<\beta}\delta(\textbf{J}^{\alpha}\cdot\textbf{J}^{\beta}-Nq^{\alpha\beta})
\prod_{\alpha}\delta(\textbf{J}^{\alpha}\cdot\textbf{J}_{\text{probe}}-N\rho^{\alpha}) \\
&\times\int\prod_{\alpha,\mu}\frac{d\lambda_{\mu}^{\alpha}d\hat{\lambda}_{\mu}^{\alpha}}{2\pi}\int\prod_{\mu}\frac{du_{\mu}d\hat{u}_{\mu}}{2\pi}\prod_{\alpha,\mu}\Theta\big(u_{\mu}(\lambda_{\mu}^{\alpha}-\kappa)\big)\prod_{\mu}\exp\bigg(i\sum_{\alpha}\lambda_{\mu}^{\alpha}\hat{\lambda}_{\mu}^{\alpha}\bigg) \\
&\times\bigg<\exp\bigg[-\frac{1}{2}\sum_{\alpha\beta}\hat{\lambda}_{\mu}^{\alpha}\hat{\lambda}_{\mu}^{\beta}(q^{\alpha\beta}-\rho^{\alpha}\rho^{\beta})-i\sum_{\alpha}\hat{\lambda}_{\mu}^{\alpha}\rho^{\alpha}z^{\mu} \\
& \quad \quad \quad +\frac{1}{2\sin^{2}\theta}\bigg(i(u_{\mu}-z^{\mu}\cos\theta)-\sum_{\alpha}\hat{\lambda}_{\mu}^{\alpha}(R^{\alpha}-\rho^{\alpha}\cos\theta)\bigg)^{2}\bigg]\bigg>_{z^{\mu}}
\end{split}
\end{align}

Expanding,

\begin{align}
\begin{split}
\Omega^{(n)}&=\int\prod_{\alpha<\beta}dq^{\alpha\beta}\int\prod_{\alpha}dR^{\alpha}\int\prod_{\alpha}d\rho^{\alpha} \\
&\times\int\prod_{\alpha=1}^{n}d\mu(\textbf{J}^{\alpha})\bigg<\prod_{\alpha}\delta(\textbf{T}\cdot\textbf{J}^{\alpha}-NR^{\alpha})\bigg>_{\textbf{T}}
\prod_{\alpha<\beta}\delta(\textbf{J}^{\alpha}\cdot\textbf{J}^{\beta}-Nq^{\alpha\beta})
\prod_{\alpha}\delta(\textbf{J}^{\alpha}\cdot\textbf{J}_{\text{probe}}-N\rho^{\alpha}) \\
&\times\int\prod_{\alpha,\mu}\frac{d\lambda_{\mu}^{\alpha}d\hat{\lambda}_{\mu}^{\alpha}}{2\pi}\int\prod_{\mu}\frac{du_{\mu}d\hat{u}_{\mu}}{2\pi}\prod_{\alpha,\mu}\Theta\big(u_{\mu}(\lambda_{\mu}^{\alpha}-\kappa)\big)\prod_{\mu}\exp\bigg(i\sum_{\alpha}\lambda_{\mu}^{\alpha}\hat{\lambda}_{\mu}^{\alpha}-\frac{1}{2}\frac{(u_{\mu}-z^{\mu}\cos\theta)^{2}}{\sin^{2}\theta}\bigg) \\
&\times\bigg<\exp\bigg[-\frac{1}{2}\sum_{\alpha\beta}\hat{\lambda}_{\mu}^{\alpha}\hat{\lambda}_{\mu}^{\beta}(q^{\alpha\beta}-\rho^{\alpha}\rho^{\beta})-i\sum_{\alpha}\hat{\lambda}_{\mu}^{\alpha}\rho^{\alpha}z^{\mu} \\
&-\frac{i}{\sin^{2}\theta}\big(u_{\mu}-z^{\mu}\cos\theta)\sum_{\alpha}\hat{\lambda}_{\mu}^{\alpha}(R^{\alpha}-\rho^{\alpha}\cos\theta)+\frac{1}{2\sin^{2}\theta}\sum_{\alpha\beta}\hat{\lambda}_{\mu}^{\alpha}\hat{\lambda}_{\mu}^{\beta}(R^{\alpha}-\rho^{\alpha}\cos\theta)(R^{\beta}-\rho^{\beta}\cos\theta)\bigg]\bigg>_{z^{\mu}}
\end{split}
\end{align}

Simplifying,

\begin{align}
\begin{split}
\Omega^{(n)}&=\int\prod_{\alpha<\beta}dq^{\alpha\beta}\int\prod_{\alpha}dR^{\alpha}\int\prod_{\alpha}d\rho^{\alpha} \\
&\times\int\prod_{\alpha=1}^{n}d\mu(\textbf{J}^{\alpha})\bigg<\prod_{\alpha}\delta(\textbf{T}\cdot\textbf{J}^{\alpha}-NR^{\alpha})\bigg>_{\textbf{T}}
\prod_{\alpha<\beta}\delta(\textbf{J}^{\alpha}\cdot\textbf{J}^{\beta}-Nq^{\alpha\beta})
\prod_{\alpha}\delta(\textbf{J}^{\alpha}\cdot\textbf{J}_{\text{probe}}-N\rho^{\alpha}) \\
& \times\int\prod_{\alpha,\mu}\frac{d\lambda_{\mu}^{\alpha}d\hat{\lambda}_{\mu}^{\alpha}}{2\pi}\int\prod_{\mu}\frac{du_{\mu}d\hat{u}_{\mu}}{2\pi}\prod_{\alpha,\mu}\Theta\big(u_{\mu}(\lambda_{\mu}^{\alpha}-\kappa)\big)\exp\bigg(i\sum_{\mu,\alpha}\lambda_{\mu}^{\alpha}\hat{\lambda}_{\mu}^{\alpha}-\frac{1}{2}\frac{(u_{\mu}-z^{\mu}\cos\theta)^{2}}{\sin^{2}\theta}\bigg) \\
& \times\bigg<\exp\bigg[-\frac{1}{2}\sum_{\mu}\sum_{\alpha\beta}\hat{\lambda}_{\mu}^{\alpha}\hat{\lambda}_{\mu}^{\beta}\bigg(q^{\alpha\beta}-\rho^{\alpha}\rho^{\beta}-\frac{(R^{\alpha}-\rho^{\alpha}\cos\theta)(R^{\beta}-\rho^{\beta}\cos\theta)}{\sin^{2}\theta}\bigg)-i\sum_{\mu}\sum_{\alpha}\hat{\lambda}_{\mu}^{\alpha}\rho^{\alpha}z^{\mu} \\
& -\frac{i}{\sin^{2}\theta}\big(u_{\mu}-z^{\mu}\cos\theta)\sum_{\alpha}\hat{\lambda}_{\mu}^{\alpha}(R^{\alpha}-\rho^{\alpha}\cos\theta)\bigg]\bigg>_{z^{\mu}}
\end{split}
\end{align}

Now we introduce
integral representations for the remaining delta functions, including
the measure $d\mu(\textbf{J}^{\alpha})$, for which we introduce the
parameter $\hat{k}^{\alpha}$,

\begin{align}
\begin{split}
\Omega^{(n)}&=\int\prod_{\alpha}\frac{d\hat{k}^{\alpha}}{4\pi}\int\prod_{\alpha<\beta}\frac{dq^{\alpha\beta}d\hat{q}^{\alpha\beta}}{2\pi/N}\int\prod_{\alpha}\frac{dR^{\alpha}d\hat{R}^{\alpha}}{2\pi/N}\int\prod_{\alpha}\frac{d\rho^{\alpha}d\hat{\rho}^{\alpha}}{2\pi/N} \\
& \times\exp\bigg(i\frac{N}{2}\sum_{\alpha}\hat{k}^{\alpha}+iN\sum_{\alpha<\beta}q^{\alpha\beta}\hat{q}^{\alpha\beta}+iN\sum_{\alpha}R^{\alpha}\hat{R}^{\alpha}+iN\sum_{\alpha}\rho^{\alpha}\hat{\rho}^{\alpha}\bigg) \\
& \times\int\prod_{i,\alpha}\frac{dJ_{i}^{\alpha}}{\sqrt{2\pi e}}\exp\bigg(-\frac{i}{2}\sum_{\alpha}\hat{k}^{\alpha}\|\textbf{J}^{\alpha}\|^{2}-i\sum_{\alpha<\beta}\hat{q}^{\alpha\beta}\textbf{J}^{\alpha}\cdot\textbf{J}^{\beta}-i\sum_{\alpha}\hat{R}_{\alpha}\textbf{J}^{\alpha}\cdot\textbf{T}-i\sum_{\alpha}\hat{\rho}_{\alpha}\textbf{J}^{\alpha}\cdot\textbf{J}_{\text{probe}}\bigg) \\
& \times\int\prod_{\alpha,\mu}\frac{d\lambda_{\mu}^{\alpha}d\hat{\lambda}_{\mu}^{\alpha}}{2\pi}\int\prod_{\mu}\frac{du_{\mu}d\hat{u}_{\mu}}{2\pi}\prod_{\alpha,\mu}\Theta\big(u_{\mu}(\lambda_{\mu}^{\alpha}-\kappa)\big)\exp\bigg(i\sum_{\mu,\alpha}\lambda_{\mu}^{\alpha}\hat{\lambda}_{\mu}^{\alpha}-\frac{1}{2}\frac{(u_{\mu}-z^{\mu}\cos\theta)^{2}}{\sin^{2}\theta}\bigg) \\
& \times\bigg<\exp\bigg[-\frac{1}{2}\sum_{\mu}\sum_{\alpha\beta}\hat{\lambda}_{\mu}^{\alpha}\hat{\lambda}_{\mu}^{\beta}\bigg(q^{\alpha\beta}-\rho^{\alpha}\rho^{\beta}-\frac{(R^{\alpha}-\rho^{\alpha}\cos\theta)(R^{\beta}-\rho^{\beta}\cos\theta)}{\sin^{2}\theta}\bigg)-i\sum_{\mu}\sum_{\alpha}\hat{\lambda}_{\mu}^{\alpha}\rho^{\alpha}z^{\mu} \\
& -\frac{i}{\sin^{2}\theta}\big(u_{\mu}-z^{\mu}\cos\theta)\sum_{\alpha}\hat{\lambda}_{\mu}^{\alpha}(R^{\alpha}-\rho^{\alpha}\cos\theta)\bigg]\bigg>_{z^{\mu}}
\end{split}
\end{align}

Notice that the $u_{\mu}-\lambda_{\mu}^{\alpha}-\hat{\lambda}_{\mu}^{\alpha}-z_{\mu}$-integrals
factorize in $\mu$, and can be written as a single integral to the
power of $P=\alpha N$.

\begin{align}
\begin{split}
\Omega^{(n)} &= k\int\prod_{\alpha}\frac{d\hat{k}^{\alpha}}{4\pi}\int\prod_{\alpha<\beta}\frac{dq^{\alpha\beta}d\hat{q}^{\alpha\beta}}{2\pi/N}\int\prod_{\alpha}\frac{dR^{\alpha}d\hat{R}^{\alpha}}{2\pi/N}\int\prod_{\alpha}\frac{d\rho^{\alpha}d\hat{\rho}^{\alpha}}{2\pi/N} \\
& \times\exp\bigg(N\bigg[\frac{i}{2}\sum_{\alpha}\hat{k}^{\alpha}+i\sum_{\alpha<\beta}q^{\alpha\beta}\hat{q}^{\alpha\beta}+i\sum_{\alpha}R^{\alpha}\hat{R}^{\alpha}+i\sum_{\alpha}\rho^{\alpha}\hat{\rho}^{\alpha} \\
& +G_{S}(\hat{k}^{\alpha},\hat{q}^{\alpha\beta},\hat{R}^{\alpha},\hat{\rho}^{\alpha})+\alpha G_{E}(q^{\alpha\beta},R^{\alpha},\rho^{\alpha})\bigg]\bigg)
\end{split}
\end{align}

Where we have written the Gardner volume in terms of an $\textit{entropic}$
part $G_{S}$, which measures how many spherical couplings satisfy
the constraints,

\begin{equation}
G_{S}=\frac{1}{N}\log\int\prod_{\alpha}\frac{d\textbf{J}^\alpha}{\sqrt{2\pi e}}\exp\bigg(-\frac{i}{2}\sum_{\alpha}\hat{k}^{\alpha}\|\textbf{J}^\alpha\|^{2}-i\sum_{\alpha<\beta}\hat{q}^{\alpha\beta}\textbf{J}^{\alpha}\cdot\textbf{J}^{\beta}-i\sum_{\alpha}\hat{R}^{\alpha}\textbf{J}^{\alpha}\cdot\textbf{T}-i\sum_{\alpha}\hat{\rho}^{\alpha}\textbf{J}^{\alpha}\cdot\textbf{J}_{\text{probe}}\bigg)
\end{equation}

And an $\textit{energetic}$ part $G_{E}$, 

\begin{multline}
G_{E}=\log\int\frac{du}{\sqrt{2\pi}}\int\prod_{\alpha}\frac{d\lambda^{\alpha}d\hat{\lambda}^{\alpha}}{2\pi}\prod_{\alpha}\Theta\big(u(\lambda^{\alpha}-\kappa)\big)\exp\bigg(i\sum_{\alpha}\lambda^{\alpha}\hat{\lambda}^{\alpha}-\frac{1}{2}\frac{(u_{\mu}-z^{\mu}\cos\theta)^{2}}{\sin^{2}\theta}\bigg) \\
\times\bigg<\exp\bigg[-\frac{1}{2}\sum_{\alpha\beta}\hat{\lambda}^{\alpha}\hat{\lambda}^{\beta}\bigg(q^{\alpha\beta}-\rho^{\alpha}\rho^{\beta}-\frac{(R^{\alpha}-\rho^{\alpha}\cos\theta)(R^{\beta}-\rho^{\beta}\cos\theta)}{\sin^{2}\theta}\bigg)-i\sum_{\alpha}\hat{\lambda}^{\alpha}\rho^{\alpha}z \\
-\frac{i}{\sin^{2}\theta}\big(u-z\cos\theta)\sum_{\alpha}\hat{\lambda}^{\alpha}(R^{\alpha}-\rho^{\alpha}\cos\theta)\bigg]\bigg>_{z}
\end{multline}

We first evaluate the entropic part, $G_S$, by introducing the $n\times n$ matrices $A,B$,

\begin{align}
A_{\alpha\beta}&=i\hat{k}^{\alpha}\delta_{\alpha\beta}+i\hat{q}^{\alpha\beta}(1-\delta_{\alpha\beta}) \\
B_{\alpha\beta}&=\delta_{\alpha\beta}+q^{\alpha\beta}(1-\delta_{\alpha\beta})
\end{align}

Inserting this our expression for $G_{S}$ becomes

\begin{equation}
G_{S}=\frac{1}{N}\log\int\prod_{\alpha}\frac{d\textbf{J}^\alpha}{\sqrt{2\pi e}}\exp\bigg(-\frac{1}{2}\sum_{\alpha,\beta}{\textbf{J}^\alpha}^T A_{\alpha\beta}\textbf{J}^{\beta}-i\sum_{\alpha}\textbf{J}^\alpha\cdot(\textbf{T}\hat{R}^{\alpha}+\textbf{J}_{\text{probe}}\hat{\rho}^{\alpha})\bigg)
\end{equation}

Integrating over $\textbf{J}^\alpha$,

\begin{equation}
G_{S}=-\frac{n}{2}-\frac{1}{2}\log(\det A)-\frac{1}{2N}\sum_{\alpha,\beta}(\textbf{T}\hat{R}^{\alpha}+\textbf{J}_{\text{probe}}\hat{\rho}^{\alpha})^T A_{\alpha\beta}^{-1}(\textbf{T}\hat{R}^{\beta}+\textbf{J}_{\text{probe}}\hat{\rho}^{\beta})
\end{equation}

Now we can include the remaining terms in the expression for $\Omega^{(n)}$
outside of $G_{E}$ and $G_{S}$ by noting that
\begin{align}
    tr(AB)&=\sum_{\alpha\beta}A_{\alpha\beta}B_{\beta\alpha} \\
    &=\sum_{\alpha\beta}(i\hat{k}^{\alpha}\delta_{\alpha\beta}+i\hat{q}^{\alpha\beta}(1-\delta_{\alpha\beta}))(\delta_{\alpha\beta}+q^{\alpha\beta}(1-\delta_{\alpha\beta})) \\
    &=\sum_{\alpha}i\hat{k}^{\alpha}+2\sum_{\alpha<\beta}iq^{\alpha\beta}\hat{q}^{\alpha\beta}
\end{align}

Additionally, we can use $\log\det A=tr(\log A)$. Thus all terms
in the exponent except $G_{E}$ can be written as

\begin{equation}
-\frac{n}{2}-\frac{1}{2}tr(\log A)-\frac{1}{2N}\sum_{\alpha,\beta}(\textbf{T}\hat{R}^{\alpha}+\textbf{J}_{\text{probe}}\hat{\rho}^{\alpha})^T A_{\alpha\beta}^{-1}(\textbf{T}\hat{R}^{\beta}+\textbf{J}_{\text{probe}}\hat{\rho}^{\beta})+\frac{1}{2}tr(AB)+i\sum_{\alpha}R^{\alpha}\hat{R}^{\alpha}+i\sum_{\alpha}\rho^{\alpha}\hat{\rho}^{\alpha}\textbf{J}_{\text{probe}}
\end{equation}

Now we extremize wrt $\hat{R}^{\alpha}$ and the elements of $A$
by setting the derivatives wrt $\hat{R}^{\gamma}$, $\hat{\rho}^{\gamma}$
and $A^{\gamma\delta}$ equal to zero:

\begin{align}
0&=-\sum_{\alpha}A_{\alpha\gamma}^{-1}\textbf{T}\cdot(\textbf{T}\hat{R}^{\alpha}+\textbf{J}_{\text{probe}}\hat{\rho}^{\alpha})+iR^{\gamma}=-\sum_{\alpha}A_{\alpha\gamma}^{-1}(\hat{R}^{\alpha}+\hat{\rho}^{\alpha}\cos\theta)+iR^{\gamma} \\
0 &=-\sum_{\alpha}A_{\alpha\gamma}^{-1}\textbf{J}_{\text{probe}}\cdot(\textbf{T}\hat{R}^{\alpha}+\textbf{J}_{\text{probe}}\hat{\rho}^{\alpha})+i\rho^{\gamma}=-\sum_{\alpha}A_{\alpha\gamma}^{-1}(\hat{R}^{\alpha}\cos\theta+\hat{\rho}^{\alpha})+i\rho^{\gamma} \\
0&=-\frac{1}{2}A_{\gamma\delta}^{-1}+\frac{1}{2}\sum_{\alpha,\beta}(\textbf{T}\hat{R}^{\alpha}+\textbf{J}_{\text{probe}}\hat{\rho}^{\alpha})^{T}A_{\alpha\gamma}^{-1}A_{\beta\delta}^{-1}(\textbf{T}\hat{R}^{\beta}+\textbf{J}_{\text{probe}}\hat{\rho}^{\beta})+\frac{1}{2}B_{\gamma\delta}
\end{align}

Solving these gives

\begin{equation}
\hat{R}^{\alpha}=i\sum_{\beta}A_{\alpha\beta}\frac{R^{\beta}-\rho^{\beta}\cos\theta}{\sin^{2}\theta}
\end{equation}

\begin{equation}
\hat{\rho}^{\alpha}=i\sum_{\beta}A_{\alpha\beta}\frac{\rho^{\beta}-R^{\beta}\cos\theta}{\sin^{2}\theta}
\end{equation}

and

\begin{equation}
A_{\gamma\delta}^{-1}=B_{\gamma\delta}-\frac{R^{\gamma}R^{\delta}-R^{\gamma}\rho^{\delta}\cos\theta-R^{\delta}\rho^{\gamma}\cos\theta+\rho^{\gamma}\rho^{\delta}}{\sin^{2}\theta}\equiv C_{\gamma\delta}
\end{equation}

and now we are left with

\begin{equation}
\Omega^{(n)}\sim\exp\bigg(N\text{extr}_{q^{\alpha\beta},R^{\alpha},\rho^{\alpha}}\bigg[\frac{1}{2}tr(\log C)+\alpha G_{E}(q^{\alpha\beta},R^{\alpha})\bigg]\bigg)
\end{equation}

\subsubsection{Replica symmetry ansatz}

In order to extremize wrt $q^{\alpha\beta}, R^{\alpha},\rho^\alpha$, we
take the replica symmetry ansatz \cite{Engel2001-sp},

\begin{equation}
q^{\alpha\beta}=q,\quad R^{\alpha}=R,\quad\rho^{\alpha}=\rho
\end{equation}

Then $C$ takes the form

\begin{equation}
C_{\alpha\beta}=\delta_{\alpha\beta}-\frac{R^{2}-2R\rho\cos\theta+\rho^{2}}{\sin^{2}\theta}+q(1-\delta_{\alpha\beta})
\end{equation}
A matrix with $E$ on the diagonal and $F$ elsewhere, $C_{\alpha\beta}=E\delta_{\alpha\beta}+F(1-\delta_{\alpha\beta})$,
has $n-1$ degenerate eigenvalues $E-F$ and one eigenvalue $E+(n-1)F$.
Hence in our case $C$ has $n-1$ degenerate eigenvalues

\begin{equation}
\bigg(1-\frac{R^{2}-2R\rho\cos\theta+\rho^{2}}{\sin^2\theta}\bigg)-\bigg(q-\frac{R^{2}-2R\rho\cos\theta+\rho^{2}}{\sin^2\theta}\bigg)=1-q
\end{equation}

and one other eigenvalue, 
\begin{equation}
\bigg(1-\frac{R^{2}-2R\rho\cos\theta+\rho^{2}}{\sin^2\theta}\bigg)+(n-1)\bigg(q -\frac{R^{2}-2R\rho\cos\theta+\rho^{2}}{\sin^2\theta}\bigg)
=1-q+n\bigg(q-\frac{R^{2}-2R\rho\cos\theta+\rho^{2}}{\sin^2\theta}\bigg)
\end{equation}

Therefore,

\begin{align}
\begin{split}
tr(\log C)&=(n-1)\log\bigg(1-q\bigg)+\log\bigg[1-q+n\bigg(q-\frac{R^{2}-2R\rho\cos\theta+\rho^{2}}{\sin^2\theta}\bigg)\bigg] \\
&=n\log\bigg(1-q\bigg)+\log\bigg[1+n\bigg(\frac{q\sin^2\theta-(R^{2}-2R\rho\cos\theta+\rho^{2})}{(1-q)\sin^2\theta}\bigg)\bigg]
\end{split}
\end{align}

We next evaluate the energetic part, $G_{E}$,

\begin{align}
\begin{split}
G_{E}&=\log\int\frac{du}{\sqrt{2\pi}}\int\prod_{\alpha}\frac{d\lambda^{\alpha}d\hat{\lambda}^{\alpha}}{2\pi}\prod_{\alpha}\Theta\big(u(\lambda^{\alpha}-\kappa)\big)\exp\bigg(i\sum_{\alpha}\lambda^{\alpha}\hat{\lambda}^{\alpha}-\frac{1}{2}\frac{(u-z\cos\theta)^{2}}{\sin^{2}\theta}\bigg) \\
&\times\bigg<\exp\bigg[-\frac{1}{2}\sum_{\alpha}(\hat{\lambda}^{\alpha})^{2}\bigg(1-\rho^{2}-\frac{(R-\rho\cos\theta)^{2}}{\sin^{2}\theta}\bigg)-\frac{1}{2}\sum_{\alpha\neq\beta}\hat{\lambda}^{\alpha}\hat{\lambda}^{\beta}\bigg(q-\rho^{2}-\frac{(R-\rho\cos\theta)^{2}}{\sin^{2}\theta}\bigg) \\
&-i\sum_{\alpha}\hat{\lambda}^{\alpha}\rho^{\alpha}z-\frac{i}{\sin^{2}\theta}\big(u-z\cos\theta)\sum_{\alpha}\hat{\lambda}^{\alpha}(R^{\alpha}-\rho^{\alpha}\cos\theta)\bigg]\bigg>_{z}
\end{split}
\end{align}

First note that we can rewrite the terms

\begin{align}
\begin{split}
&-\frac{1}{2}\sum_{\alpha}\bigg(1-\rho^{2}-\frac{(R-\rho\cos\theta)}{\sin^{2}\theta}\bigg)(\hat{\lambda}^{\alpha})^{2}-\frac{1}{2}\sum_{\alpha\neq\beta}\hat{\lambda}^{\alpha}\hat{\lambda}^{\beta}\bigg(q-\rho^{2}-\frac{(R-\rho\cos\theta)^{2}}{\sin^{2}\theta}\bigg) \\
& \quad =-\frac{1}{2}\sum_{\alpha}\bigg(1-\rho^{2}-\frac{(R-\rho\cos\theta)^{2}}{\sin^{2}\theta}\bigg)(\hat{\lambda}^{\alpha})^{2}-\frac{1}{2}\bigg(q-\rho^{2}-\frac{(R-\rho\cos\theta)^{2}}{\sin^{2}\theta}\bigg)\bigg[(\sum_{\alpha}\hat{\lambda}^{\alpha})^{2}-\sum_{\alpha}(\hat{\lambda}^{\alpha})^{2}\bigg] \\
& \quad =-\frac{1}{2}\sum_{\alpha}(1-q)(\hat{\lambda}^{\alpha})^{2}-\frac{1}{2}\bigg(q-\rho^{2}-\frac{(R-\rho\cos\theta)^{2}}{\sin^{2}\theta}\bigg)(\sum_{\alpha}\hat{\lambda}^{\alpha})^{2}
\end{split}
\end{align}

To simplify the last term we apply the Hubbard-Stratonovich transformation,
$e^{b^{2}/2}=\int Dte^{bt}$, introducing auxiliary field $t$,

\begin{multline}
=\log\int\frac{du}{\sqrt{2\pi}}\int\prod_{\alpha}\frac{d\lambda^{\alpha}d\hat{\lambda}^{\alpha}}{2\pi}\prod_{\alpha}\Theta\big(u(\lambda^{\alpha}-\kappa)\big)\int Dt\bigg<\exp\bigg[-\frac{1-q}{2}\sum_{\alpha}(\hat{\lambda}^{\alpha})^{2} \\
+i\sum_{\alpha}\hat{\lambda}^{\alpha}\bigg(\lambda^{\alpha}-\rho^{\alpha}z-\frac{\big(u-z\cos\theta)(R^{\alpha}-\rho^{\alpha}\cos\theta)}{\sin^{2}\theta}-\sqrt{q-\rho^{2}-\frac{(R-\rho\cos\theta)^{2}}{\sin^{2}\theta}}t\bigg)-\frac{1}{2}\frac{(u-z\cos\theta)^{2}}{\sin^{2}\theta}\bigg]\bigg>_{z}
\end{multline}

Using the $\Theta$-function to restrict the bounds of integration,

\begin{multline}
=\log2\int_{0}^{\infty}\frac{du}{\sqrt{2\pi}}\int_{\kappa}^{\infty}\prod_{\alpha}\frac{d\lambda^{\alpha}}{\sqrt{2\pi}}\int\prod_{\alpha}\frac{d\hat{\lambda}^{\alpha}}{\sqrt{2\pi}}\int Dt\bigg<\exp\bigg[-\frac{1-q}{2}\sum_{\alpha}(\hat{\lambda}^{\alpha})^{2} \\
+i\sum_{\alpha}\hat{\lambda}^{\alpha}\bigg(\lambda^{\alpha}-\rho^{\alpha}z-\frac{\big(u-z\cos\theta)(R^{\alpha}-\rho^{\alpha}\cos\theta)}{\sin^{2}\theta}-\sqrt{q-\rho^{2}-\frac{(R-\rho\cos\theta)^{2}}{\sin^{2}\theta}}t\bigg)-\frac{1}{2}\frac{(u-z\cos\theta)^{2}}{\sin^{2}\theta}\bigg]\bigg>_{z}
\end{multline}

Now we can perform the gaussian integrals over $\hat{\lambda}^{\alpha}$,

\begin{multline}
=\log2\int_{0}^{\infty}\frac{du}{\sqrt{2\pi}}\int_{\kappa}^{\infty}\prod_{\alpha}\frac{d\lambda^{\alpha}}{\sqrt{2\pi}}\int Dt \bigg<\exp\bigg[-\frac{1}{2(1-q)}\bigg(\lambda^{\alpha}-\rho^{\alpha}z \\ -\frac{\big(u-z\cos\theta)(R^{\alpha}-\rho^{\alpha}\cos\theta)}{\sin^{2}\theta}-\sqrt{q-\rho^{2}-\frac{(R-\rho\cos\theta)^{2}}{\sin^{2}\theta}}t\bigg)^{2}-\frac{1}{2}\frac{(u-z\cos\theta)^{2}}{\sin^{2}\theta}\bigg]\bigg>_{z}
\end{multline}

And $\lambda^{\alpha}$,

\begin{multline}
=\log2\int_{0}^{\infty}\frac{du}{\sqrt{2\pi}}\int Dt\bigg<H^{n}\bigg[-\frac{1}{\sqrt{1-q}}\bigg(\kappa-\rho^{\alpha}z+\frac{\big(u-z\cos\theta)(R^{\alpha}-\rho^{\alpha}\cos\theta)}{\sin^{2}\theta}+\sqrt{q-\rho^{2}-\frac{(R-\rho\cos\theta)^{2}}{\sin^{2}\theta}}t\bigg)^{2}\bigg] \\
\times\exp\bigg[-\frac{1}{2}\frac{(u-z\cos\theta)^{2}}{\sin^{2}\theta}\bigg]\bigg>_{z}
\end{multline}

Shifting the integration variable $t\to(\rho^{\alpha}z+\frac{(u-z\cos\theta)(R^{\alpha}-\rho^{\alpha}\cos\theta)}{\sin^{2}\theta}+\sqrt{q-\rho^{2}-\frac{(R-\rho\cos\theta)^{2}}{\sin^{2}\theta}}t)/\sqrt{q}$,
we can finally perform the gaussian integral over $u$,

\begin{multline}
=\log2\int\frac{dt}{\sqrt{2\pi}}\ \bigg<\exp\bigg(-\frac{(\sqrt{q}t-z\rho)^{2}}{2(q-\rho)^{2}}\bigg)\sqrt{\frac{q}{q-\rho^{2}}}H^{n}\bigg(-\sqrt{\frac{q}{1-q}}t\bigg) \\
\times H\bigg(\frac{1}{\sqrt{q-\rho^{2}}}\frac{\kappa-(qR_{0}z+z\rho(R-2\rho\cos\theta)-\sqrt{q}t(R-\rho\cos\theta))}{\sqrt{q\sin^{2}\theta+2R\rho\cos\theta-R^{2}-\rho}}\bigg)\bigg>_{z}
\end{multline}

We can simplify this further by taking $t\to(\sqrt{q}t-z\rho)/\sqrt{q-\rho^{2}}$,

\begin{equation}
=\log2\int Dt\ \bigg<H^{n}\bigg(\frac{\kappa-(z\rho+\sqrt{q-\rho^{2}})t}{\sqrt{1-q}}\bigg)  H\bigg(\frac{t\left(\sqrt{q-\rho^{2}}+\rho z\right)(R-\rho\cos\theta)+qz\cos\theta-\rho Rz}{\sqrt{-\left(q-\rho^{2}\right)\left(\rho^{2}-q\sin^{2}\theta+R^{2}-2\rho R\cos\theta\right)}}\bigg)\bigg>_{z}
\end{equation}

\subsection{Quenched entropy}

Putting everything together, and using the replica identity, $\langle\log\text{\ensuremath{\Omega\rangle}=\ensuremath{\lim_{n\to0}}\ensuremath{(\langle\Omega^{n}\rangle-1)}}/n,$
we obtain an expression for the quenched entropy of the teacher-student perceptron under data pruning:

\fbox{\begin{minipage}[t]{\linewidth}%
\begin{multline}
\label{eq:full_entropy_expression}
\frac{1}{N}\langle\log\Omega\rangle=\text{extr}_{q,R,\rho}\bigg[\frac{1}{2}\log\bigg(1-q\bigg)+\frac{1}{2}\bigg(\frac{q-(R^{2}-2R\rho\cos\theta+\rho^{2})/\sin^2\theta}{1-q}\bigg) \\
+2\alpha\bigg<\int Dt\log H\bigg(\frac{\kappa-(z\rho+\sqrt{q-\rho^{2}})t}{\sqrt{1-q}}\bigg) \\
\times H\bigg(\frac{t\left(\sqrt{q-\rho^{2}}+\rho z\right)(R-\rho\cos\theta)+z(q\cos\theta-\rho R)}{\sqrt{\left(q-\rho^{2}\right)\left(R^{2}+\rho^{2}-q\sin^{2}\theta-2\rho R\cos\theta\right)}}\bigg)\bigg>_{z}\bigg]
\end{multline}
\end{minipage}}

We will now unpack this equation and use it to make predictions in
several specific settings.

\subsection{Perfect teacher-probe overlap}

We will begin by considering the case where the probe student has
learned to perfectly match the teacher, $J_{\text{probe}}=T$, which
we can obtain by the limit $\theta\to0$, $\rho\to R$. In this limit
the second $H$-function in Eq. \ref{eq:full_entropy_expression} becomes increasingly sharp, approaching a
step function:

\begin{equation}
H\bigg(\frac{t\left(\sqrt{q-\rho^{2}}+\rho z\right)(R-\rho\cos\theta)+z(q\cos\theta-\rho R)}{\sqrt{\left(q-\rho^{2}\right)\left(R^{2}+\rho^{2}-q\sin^{2}\theta-2\rho R\cos\theta\right)}}\bigg)\to\Theta(z)
\end{equation}

Hence we are left with,

\noindent %
\fbox{\begin{minipage}[t]{\linewidth}%
\begin{multline}
\label{eq:entropy_perfect_teacher_student_overlap}
\frac{1}{N}\langle\langle\ln\Omega(\textbf{x}^\mu,T,\kappa)\rangle\rangle=\text{extr}_{q,R}\bigg[\frac{1}{2}\log\bigg(1-q\bigg)+\frac{1}{2}\bigg(\frac{q-R^{2}}{1-q}\bigg) \\
+2\alpha\int Dt\int dzp(z)\Theta(z)\log H\bigg(-\frac{\sqrt{q-R^{2}}t+Rz-\kappa}{\sqrt{1-q}}\bigg)\bigg]
\end{multline}
\end{minipage}}

\subsubsection{Saddle point equations}

We can now obtain a set of self-consistent saddle point equations
by setting set to zero the derivatives with respect to $R$ and $q$
of the right side of Eq. \ref{eq:entropy_perfect_teacher_student_overlap}. As $\kappa$ approaches its critical
value, the space of solutions shrinks to a unique solution, and hence
the overlap between students $q$ approaches one. In the limit $q\to1$,
after some partial integration, we find,

\noindent %
\noindent\fbox{\begin{minipage}[t]{\linewidth}%
\begin{equation}
\label{eq:perfect_saddle_point1}
R=2\alpha\int_{-\infty}^{\kappa}\frac{dt}{\sqrt{2\pi}\sqrt{1-R^{2}}}\int_{0}^{\infty}dzp(z)\exp\bigg(-\frac{(t-Rz)^{2}}{2(1-R^{2})}\bigg)\bigg(\frac{z-Rt}{1-R^{2}}\bigg)(\kappa-t) 
\end{equation}

\begin{equation}
\label{eq:perfect_saddle_point2}
1-R^{2}=2\alpha\int_{-\infty}^{\kappa}\frac{dt}{\sqrt{2\pi}\sqrt{1-R^{2}}}\int_{0}^{\infty}dzp(z)\exp\bigg(-\frac{(t-Rz)^{2}}{2(1-R^{2})}\bigg)\big(\kappa-t\big)^{2}
\end{equation}
\end{minipage}}

These saddle point equations can be solved numerically to find $R$
and $\kappa$ as a function of $\alpha$ for a student perceptron
trained on a dataset with an arbitrary distribution along the teacher
direction $p(z)$.
We can specialize to the case of data pruning by setting $p(z)$ to the
distribution found after pruning an initially Gaussian-distributed
dataset so that only a fraction $f$ of those examples with the smallest
margin along the teacher are kept, $p(z)=\frac{e^{-z^{2}/2}}{\sqrt{2\pi}f}\Theta(\gamma-|z|)$,
where the threshold $\gamma=H^{-1}\big(\frac{1-f}{2}\big)$.

\noindent\fbox{\begin{minipage}[t]{\linewidth}%
\begin{equation}
R=\frac{2\alpha}{f\sqrt{2\pi}\sqrt{1-R^{2}}}\int_{-\infty}^{\kappa}Dt\ \exp\bigg(-\frac{R^{2}t^{2}}{2(1-R^{2})}\bigg)\bigg[1-\exp\bigg(-\frac{\gamma(\gamma-2Rt)}{2(1-R^{2})}\bigg)\bigg](\kappa-t) 
\end{equation}

\begin{equation}
1-R^{2}=\frac{2\alpha}{f}\int_{-\infty}^{\kappa}Dt\ \bigg[H\bigg(-\frac{Rt}{\sqrt{1-R^{2}}}\bigg)-H\bigg(-\frac{Rt-\gamma}{\sqrt{1-R^{2}}}\bigg)\bigg](\kappa-t)^{2}
\end{equation}
\end{minipage}}

Solving these saddle point equations numerically for $R$ and $\kappa$
yields an excellent fit to numerical simulations, as can be seen in
Fig.~\ref{fig:theory}\textbf{A}. It is also easy to verify that in
the limit of no data pruning ($f\to1,\gamma\to\infty$) we recover
the saddle point equations for the classical teacher-student perceptron (Eqs. 4.4 and 4.5 in \cite{Engel2001-sp}),

\begin{equation}
R=\frac{2\alpha}{\sqrt{2\pi}\sqrt{1-R^{2}}}\int Dt\exp\bigg(-\frac{R^{2}t^{2}}{2(1-R^{2})}\bigg)(\kappa-t)
\end{equation}

\begin{equation}
1-R^{2}=2\alpha\int Dt\ H\bigg(-\frac{Rt}{\sqrt{1-R^{2}}}\bigg)\big(\kappa-t\big)^{2}
\end{equation}

\subsection{Information gain per example}
\label{app:information_gain}

Why does data pruning allow for super-exponential performance with
dataset size $\alpha$? We can define the amount of information gained
from each new example, $I(\alpha),$as the fraction by which the space
of solutions which perfectly classify the data is reduced when a new
training example is added, $I(\alpha)=\Omega(\frac{P+1}{N})/\Omega(\frac{P}{N})$.
Or, equivalently, the rate at which the entropy is reduced, $I(\alpha)=-\frac{d}{d\alpha}S(\alpha)$.
Of coure, the volume of solutions shrinks to zero at the max-margin
solution; so to study the volume of solutions which perfectly classify
the data we simply set the margin to zero $\kappa=0$. In \cite{Freund1992-uv} the information gain for a perceptron in the classical
teacher-student setting is shown to take the form,

\begin{equation}
\label{eq:info_gain_no_pruning}
I(\alpha)=-2\int Dt\ H\bigg(\sqrt{\frac{R}{1-R}}t\bigg)\log H\bigg(\sqrt{\frac{R}{1-R}}t\bigg)
\end{equation}

Which goes to zero in the limit of large $\alpha$ as $I(\alpha)\sim1/\alpha$.
Data pruning can increase the information gained per example by pruning
away the uninformative examples. To show this, we generalize the calculation
of the information gain to pruned datasets, using the expression for
the entropy we obtained in the previous section (Eq. \ref{eq:entropy_perfect_teacher_student_overlap}). 

\begin{equation}
\label{eq:gibbs_entropy}
S(\alpha)=\frac{1}{N}\langle\log\Omega\rangle=\text{extr}_{q,R}\bigg[\frac{1}{2}\log\bigg(1-R\bigg)+\frac{1}{2}R+2\alpha\int Dt\int_0^\infty dz\ p(z)\log H\bigg(-\sqrt{R}t-\frac{R}{\sqrt{1-R}}z\bigg)\bigg]
\end{equation}

Hence the information gain $I(\alpha)=-\frac{d}{d\alpha}S(\alpha)$
is given by

\begin{equation}
I(\alpha)=2\alpha\int Dt\int dzp(z)\Theta(z)\log H\bigg(-\sqrt{R}t-\frac{R}{\sqrt{1-R}}z\bigg)
\end{equation}

Changing variables to $t\to-(\sqrt{R}t+\frac{R}{\sqrt{1-R}}z)/\sqrt{q}$,

\begin{equation}
I(\alpha)=2\alpha\int Dt\int_{0}^{\infty}dzp(z)\frac{1}{\sqrt{1-R}}\frac{1}{\sqrt{2\pi}}\exp\bigg(-\frac{z^{2}+2\sqrt{R}tz}{2(1-R)}\bigg)\log H\bigg(\sqrt{\frac{R}{1-R}}t\bigg)
\end{equation}

Now, assuming that we prune to a fraction $f$, so that $p(z)=\Theta\big(|z|-\gamma\big)\frac{\exp(-z^{2}/2)}{\sqrt{2\pi}f}$,
where $\gamma=H^{-1}\bigg(\frac{1-f}{2}\bigg)$

\begin{equation}
\label{eq:info_gain_prune_window}
I(\alpha)=\frac{2\alpha}{f}\int Dt\ \bigg[H\bigg(\sqrt{\frac{R}{1-R}}t\bigg)-H\bigg(\frac{\gamma+\sqrt{R}t}{\sqrt{1-R}}\bigg)\bigg]\log H\bigg(\sqrt{\frac{R}{1-R}}t\bigg)\bigg]
\end{equation}

$I(\alpha)$ is plotted for varying values of $f$ in Fig.~\ref{fig:theory}\textbf{F}. Notice that for $f\to1$, $\gamma\to\infty$ and we recover
Eq. \ref{eq:info_gain_no_pruning}. To obtain the optimal pruning fraction $f_\text{opt}$ for any $\alpha$, we first need an equation for $R$, which can be obtained by taking the saddle point of Eq. \ref{eq:gibbs_entropy}. Next we optimize $I(\alpha)$ by setting the derivative of Eq. \ref{eq:info_gain_prune_window} with respect to $f$ equal to zero. This gives us a pair of equations which can be solved numerically to obtain $f_\text{opt}$ for any $\alpha$.

Finally, Eq. \ref{eq:info_gain_prune_window} reveals that as we prune more aggressively the information gain per example approaches a finite rate. As $f\to0$, $\gamma\to0$, and we obtain,

\begin{equation}
\label{eq:info_gain_infinite_pruning}
I(\alpha)=-\int Dt\ \log H(\sqrt{R}t)
\end{equation}

Which allows us to produce to trace the Pareto frontier in Fig.~\ref{fig:theory}\textbf{F}. For $R\to1$, Eq. \ref{eq:info_gain_infinite_pruning} gives
the asymptotic information gain $I(\infty)=1$ nat/example.

\subsection{Imperfect teacher-probe overlap}
\label{app:imperfect_teacher_probe_overlap}

In realistic settings we expect the probe student to have only partial
information about the target function. What happens if the probe student
does not perfectly match the teacher? To understand this carefully,
we need to compute the full set of saddle point equations over $R,q,$
and $\rho$, which we will do in the following section. But to first
get an idea for what goes wrong, we include in this section a simple
sketch which reveals the limiting behavior.

Consider the case where the angle between the probe student and teacher
is $\theta$. Rotate coordinates so that the first canonical basis
vector aligns with the student $J=(1,0,\ldots,0)$, and the teacher
lies in the span of the first two canonical basis vectors, $T=(\cos\theta,\sin\theta,0,\ldots,0)$.
Consider the margin along the teacher of a new training example $x$
drawn from the pruned distribution.

\begin{equation}
\mathbb{E}|T\cdot x|^{2}=\mathbb{E}[x_{0}^{2}\cos^{2}\theta+x_{1}^{2}\sin^{2}\theta]
\end{equation}

As the fraction of examples kept goes to zero, $\mathbb{E}x_{0}^{2}\to0$,
and the average margin of a new example converges to a fixed value,

\begin{equation}
\mathbb{E}|T\cdot x|^{2}=\sin^{2}\theta
\end{equation}

Hence the data ultimately stops concentrating around the teacher's
decision boundary, and the information gained from each new example
goes to zero. Therefore we expect the generalization error to converge
to a power law, where the constant prefactor is roughly that of pruning
with a prune fraction $f_{\text{min}}$ which yields an average margin
of $1-R^{2}$. This ``minimum'' pruning fraction lower bounds the generalization error envelope (see Fig.~\ref{fig:teacher_student_overlap}), and satisfies the following
equation,

\begin{equation}
\label{eq:fmin_self_consistent}
\int_{-\gamma_\text{min}}^{\gamma_\text{min}}dxp(x)x^{2}=\frac{1}{2}-\frac{e^{-\gamma_\text{min}^{2}/2}\gamma_\text{min}}{\sqrt{2\pi}(1-2H(\gamma_\text{min}))}=1-R^{2}
\end{equation}

where $\gamma_\text{min}=H^{-1}\bigg(\frac{1-f_\text{min}}{2}\bigg)$. Eq. \ref{eq:fmin_self_consistent} can
be solved numerically, and we use it to produce the lower-bounding power laws shown in red in Fig.~\ref{fig:teacher_student_overlap}\textbf{C,D}. The minimum achievable pruning fraction $f_{\text{min}}(\theta)$ approaches zero as the angle between the probe student and the teacher shrinks, and we can obtain its scaling by taking $R\to1$, in which case we find,

\begin{equation}
f_{\text{min}}(\theta) \sim\theta
\end{equation}

\subsection{Optimal pruning policy}

\begin{figure}[h!]
    \centering
    \includegraphics[width=0.8\linewidth]{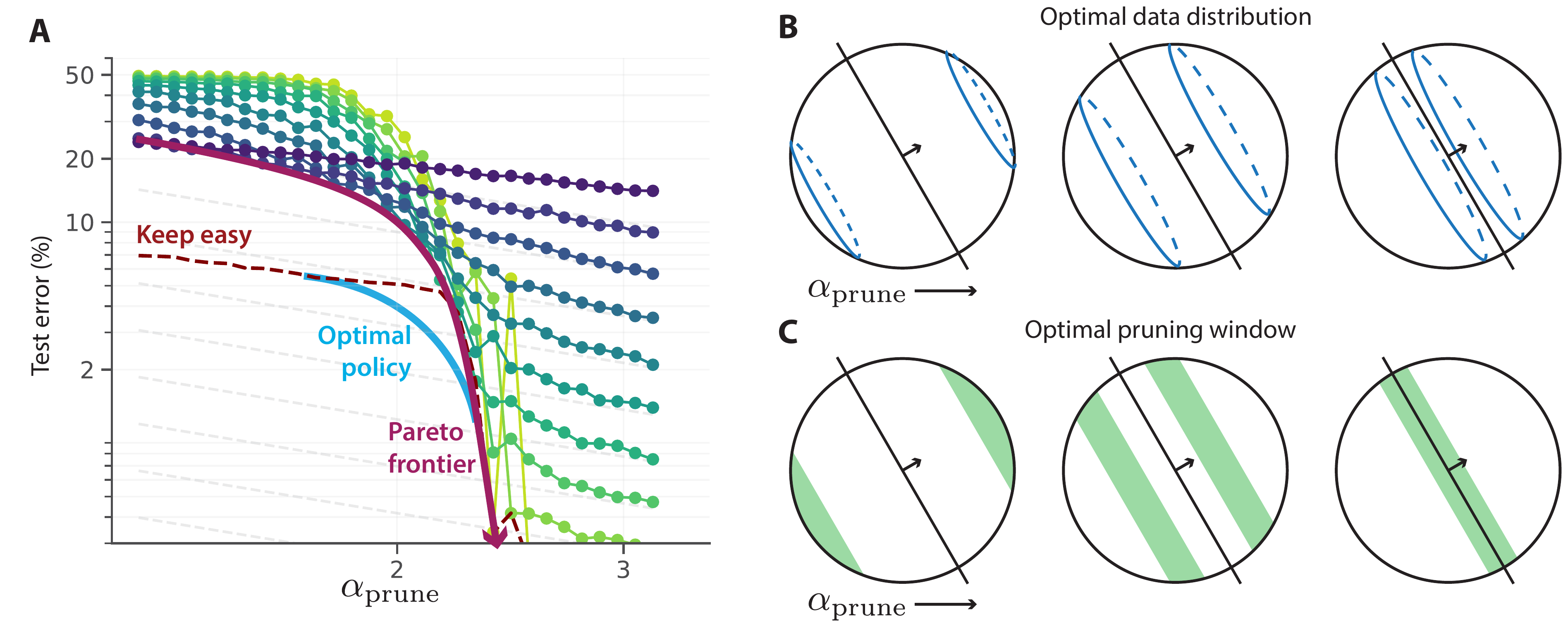}
    \caption{Optimal pruning policy as a function of $\alpha_\text{prune}$. \textbf{A}, The Pareto frontier in Fig. \ref{app:theory}\textbf{A} can be lowered in the small $\alpha_\text{prune}$ regime if one adaptively switches pruning policies from a ``keep easy'' to a ``keep hard'' policy. The dashed purple line indicates the ``keep easy'' frontier (computed using numerical simulations). The optimal pruning window (derived below) interpolates between the two policies, achieving the lowest possible Pareto frontier (cartooned in blue). \textbf{B}, The optimal data distribution along the teacher $p(z|\alpha_\text{prune},f)$ is a delta function, which selects easy examples for small $\alpha_\text{prune}$, intermediate examples for intermediate $\alpha_\text{prune}$, and hard examples for large $\alpha_\text{prune}$. \textbf{C}, The optimal pruning window similarly selects easy examples for small $\alpha_\text{prune}$, intermediate examples for intermediate $\alpha_\text{prune}$, and hard examples for large $\alpha_\text{prune}$ }
    \label{fig:optimal_pruning_cartoon}
\end{figure}

The saddle point equations Eq. \ref{eq:perfect_saddle_point1},\ref{eq:perfect_saddle_point2} reveal that the optimal pruning policy varies as a function of $\alpha_\text{prune}$. For $\alpha_\text{prune}$ large the best policy is to retain only the ``hardest" (smallest-margin) examples. But when $\alpha_\text{prune}$ is small, keeping the ``hardest" examples performs worse than chance, suggesting that the best policy in the $\alpha_\text{prune}$ small regime is to keep the easiest examples. Indeed by switching between the ``keep easy'' and ``keep hard'' strategies as $\alpha_\text{prune}$ grows, one can achieve a lower Pareto frontier than the one shown in Fig. \ref{fig:theory}A in the small $\alpha_\text{prune}$ regime (Fig. \ref{fig:optimal_pruning_cartoon}C).

These observations beg the question: what is the best policy in the intermediate $\alpha_\text{prune}$ regime? Is there a globally optimal pruning policy which interpolates between the ``keep easy'' and ``keep hard'' strategies and achieves the lowest possible Pareto frontier (blue curve in Fig. \ref{fig:optimal_pruning_cartoon}A)?

In this section we investigate this question. Using  the calculus of variations, we first derive the optimal data distribution $p(z|\alpha_\text{prune},f)$ along the teacher for a given $\alpha_\text{prune}$, $f$. We begin by framing the problem using the method of Lagrange multipliers. Seeking to optimize $R$ under the constraints imposed by the saddle point equations Eqs. \ref{eq:perfect_saddle_point1},\ref{eq:perfect_saddle_point2}, we define the Lagrangian,

\begin{equation}
    \mathcal L = R + \mu \bigg( R- 2\alpha \int_{0}^{\infty}dz \ p(z) \varphi(z;R,k) \bigg) + \lambda \bigg(1-R^2 -  2\alpha \int_{0}^{\infty}dz \ p(z) \psi(z;R,k)\bigg).
\end{equation}

Where,

\begin{align}
    \varphi(z;R,k) 
    &= \int_{-\infty}^{\kappa}\frac{dt}{\sqrt{2\pi}\sqrt{1-R^{2}}}\exp\bigg(-\frac{(t-Rz)^{2}}{2(1-R^{2})}\bigg)\bigg(\frac{z-Rt}{1-R^{2}}\bigg)(\kappa-t), \\
    \intertext{and,}
    \psi(z;R,k) &= \int_{-\infty}^{\kappa}\frac{dt}{\sqrt{2\pi}\sqrt{1-R^{2}}}\exp\bigg(-\frac{(t-Rz)^{2}}{2(1-R^{2})}\bigg)\big(\kappa-t\big)^{2}
\end{align}

Taking a variational derivative $\frac{\delta \mathcal L}{\delta p}$ with respect to the data distribution $p$, we obtain an equation for $z$, indicating that the optimal distribution is a delta function at $z=z^*$. To find the optimal location of the delta function $z^*$, we take derivatives with respect to the remaining variables $R,k,\mu,\lambda$ and solve the resulting set of equations numerically. The qualitative behavior is shown in Fig. \ref{fig:optimal_pruning_cartoon}A. As $\alpha_\text{prune}$ grows, the location of the delta function shifts from infinity to zero, confirming that the optimal strategy for small $\alpha_\text{prune}$ is to keep the "easy" (large-margin) examples, and for large $\alpha_\text{prune}$ to keep the "hard" (small-margin) examples.

Interestingly, this calculation also reveals that if the location of the delta function is chosen optimally, the student can perfectly recover the teacher ($R=1$, zero generalization error) for any $\alpha_\text{prune}$. This observation, while interesting, is of no practical consequence because it relies on an infinitely large training set from which examples can be precisely selected to perfectly recover the teacher. Therefore, to derive the optimal pruning policy for a more realistic scenario, we assume a gaussian distribution of data along the teacher direction and model pruning as keeping only those examples which fall inside a window $a<z<b$. The saddle point equations, Eqs. \ref{eq:perfect_saddle_point1},\ref{eq:perfect_saddle_point2}, then take the form,

\begin{equation}
    R = \frac{2\alpha}{f\sqrt{\pi/2}\sqrt{1-R^2}} \int_{-\infty}^\kappa Dt \left[\exp\left(-\frac{(a-Rt)^2}{2(1-R^2)}\right) - \exp\left(-\frac{(b-Rt)^2}{2(1-R^2)} \right)\right](\kappa-t)
\end{equation}

\begin{equation}
    1-R^2 = \frac{4\alpha}{f} \int_{-\infty}^\kappa Dt \left[H\left(\frac{a-R t}{\sqrt{1- R^2}}\right)-H\left(\frac{b-R t}{\sqrt{1- R^2}}\right)\right](\kappa-t)^2
\end{equation}

Where $a$ must satisfy $a=H^{-1}(f/2+H(b))$. For each $f,\alpha$, we find the optimal location of this window using the method of Lagrange multipliers. Defining the Lagrangian as before,

\begin{equation}
    \mathcal L = R + \mu \bigg( R- 2\alpha \int_{0}^{\infty}dz \ p(z) \varphi(z;R,k) \bigg) + \lambda \bigg(1-R^2 -  2\alpha \int_{0}^{\infty}dz \ p(z) \psi(z;R,k)\bigg),
\end{equation}

Where now,

\begin{align}
    \phi(b;R,k) &= \left[\exp\left(-\frac{(a-Rt)^2}{2(1-R^2)}\right) - \exp\left(-\frac{(b-Rt)^2}{2(1-R^2)} \right)\right](\kappa-t) \\
    \psi(b;R,k) &= \left[H\left(\frac{a-R t}{\sqrt{1- R^2}}\right)-H\left(\frac{b-R t}{\sqrt{1- R^2}}\right)\right](\kappa-t)^2
\end{align}

 To find the optimal location of the pruning window, we take derivatives with respect to the remaining variables $b,R,k,\mu,\lambda$ and solve the resulting set of equations numerically. Consistent with the results for the optimal distribution, the location of the optimal window shifts from around infinity to around zero as $\alpha_\text{prune}$ grows (Fig. \ref{fig:optimal_pruning_cartoon}\textbf{C}).

\subsection{Exact saddle point equations}

To obtain exact expressions for the generalization error for all $\theta$,
we can extremize Eq. \ref{eq:full_entropy_expression} wrt $R,q,\rho$.

\paragraph{Derivative wrt $R$}

\begin{equation}
\frac{R-\rho\cos\theta}{(1-q)\sin^{2}\theta}=\int dt\int dz\ p(z)\frac{\sqrt{2}\alpha}{\pi}\bigg(\frac{t\sqrt{q-\rho^{2}}}{\sqrt{2}\sqrt{\left(\rho^{2}-q\right)}\Lambda}-\frac{(R-\rho\cos\theta)\Gamma(t,z)}{\sqrt{2}\sqrt{q-\rho^{2}}\Lambda^{3}}\bigg)
\end{equation}

\begin{equation}
\times\log\left(H\left(\frac{\kappa-t\sqrt{q-\rho^{2}}-\rho z}{\sqrt{1-q}}\right)\right)\exp\left(-\frac{\Gamma(t,z)^{2}}{2\left(q-\rho^{2}\right)\Lambda^{2}}-\frac{t^{2}}{2}\right)
\end{equation}

Where we have defined, \begin{equation}\Lambda=\sqrt{q\sin^{2}\theta-R^{2}-\rho^{2}+2R\rho\cos\theta},
\end{equation}
\begin{equation}\Gamma(t,z)=(R-\rho\cos\theta)\left(t\sqrt{q-\rho^{2}}+\rho z\right)+qz\cos\theta-\rho Rz.
\end{equation}
Integrating the right-hand side by parts,

\begin{equation}
\frac{R-\rho\cos\theta}{(1-q)\sin^{2}\theta}=-\int dt\int dz\ p(z)\frac{\alpha}{\pi\Lambda}\frac{\sqrt{q-\rho^{2}}e^{-\frac{\left(\kappa-t\sqrt{q-\rho^{2}}-\rho z\right)^{2}}{2(1-q)}}}{\sqrt{2\pi}\sqrt{1-q}\text{H}\left(\frac{\kappa-t\sqrt{q-\rho^{2}}-\rho z}{\sqrt{1-q}}\right)}\exp\left(-\frac{\Delta(t,z)}{2\Lambda^{2}}\right)
\end{equation}

where 

\begin{equation}
\Delta(t,z)=2tz\sqrt{q-\rho^{2}}(R-\rho\cos\theta)\cos\theta+qt^{2}\sin^{2}\theta+qz^{2}\cos^{2}\theta-\rho^{2}t^{2}\sin^{2}\theta-\rho^{2}z^{2}\cos^{2}\theta
\end{equation}

Changing variables to $t\to t\sqrt{q-\rho^{2}}+\rho z$ and taking
the limit $q\to1$,

\noindent 
\begin{equation}
\frac{R-\rho\cos\theta}{\sin^{2}\theta}=\bigg<\int_{-\infty}^{\kappa}dt\frac{\alpha}{\pi\Lambda}\exp\left(-\frac{\Delta(t,z)}{2\Lambda^{2}}\right)(\kappa-t)\bigg>_{z}
\end{equation}

Where with this change of variables,

\begin{equation}
\Lambda=\sqrt{q\sin^{2}\theta-R^{2}-\rho^{2}+2\rho R\cos\theta}
\end{equation}

\begin{equation}
\Delta(t,z)=z^{2}\left(\rho^{2}+\cos^{2}\theta-2\rho R\cos\theta\right)+2tz(R\cos\theta-\rho)+t^{2}\sin^{2}\theta
\end{equation}

\paragraph{Derivative wrt $q$,}

\begin{equation}
\frac{q-(\rho^{2}+R^{2}-2\rho R\cos\theta)/\sin^2\theta}{2(1-q)^{2}}=\int dt\int dz\ p(z)\frac{2\alpha}{\pi}\left(\frac{\kappa-t\sqrt{q-\rho^{2}}-\rho z}{(1-q)^{3/2}}-\frac{t}{\sqrt{1-q}\sqrt{q-\rho^{2}}}\right)
\end{equation}

\begin{equation}
\times\exp\left(-\frac{\left(\kappa-t\sqrt{q-\rho^{2}}-\rho z\right)^{2}}{2(1-q)}-\frac{t^{2}}{2}\right)
\end{equation}

\begin{equation}
\times H\left(-\frac{(R-\rho\cos\theta)\left(t\sqrt{q-\rho^{2}}+\rho z\right)+qz\cos\theta-\rho Rz}{\sqrt{\left(\rho^{2}-q\right)\left(\rho^{2}-q\sin^{2}\theta+R^{2}-2\rho R\cos\theta\right)}}\right)H\left(\frac{\kappa-t\sqrt{q-\rho^{2}}-\rho z}{\sqrt{1-q}}\right)
\end{equation}

\begin{equation}
-\frac{\sqrt{2}\alpha}{\pi}\bigg(\frac{\frac{t(R-\rho\cos\theta)}{2\sqrt{q-\rho^{2}}}+z\cos}{\sqrt{2}\sqrt{\left(q-\rho^{2}\right)}\Lambda}-\frac{\left(\left(q-\rho^{2}\right)\sin^{2}\theta+\Lambda^{2}\right)\Gamma(t,z)}{2\sqrt{2}\left(\rho^{2}-q\right)^{3/2}\Lambda^{3}}\bigg)
\end{equation}

\begin{equation}
\times\log\left(H\left(\frac{\kappa-t\sqrt{q-\rho^{2}}-\rho z}{\sqrt{1-q}}\right)\right)\exp\left(-\frac{\left((R-\rho\cos\theta)\left(t\sqrt{q-\rho^{2}}+\rho z\right)+qz\cos\theta-\rho Rz\right)^{2}}{2\left(q-\rho^{2}\right)\Lambda^{2}}-\frac{t^{2}}{2}\right)
\end{equation}

Where $\Gamma(t,z)=(R-\rho\cos\theta)\left(t\sqrt{q-\rho^{2}}+\rho z\right)+qz\cos\theta-\rho Rz$.
After integating by parts,

\begin{equation}
\frac{q-(\rho^{2}+R^{2}-2\rho R\cos\theta)/\sin^{2}\theta}{2(1-q)^{2}}=\int dt\int dz\ p(z)\frac{\alpha\exp\left(-\frac{2t\sqrt{q-\rho^{2}}(\rho z-\kappa)-\left(\rho^{2}-1\right)t^{2}+(\kappa-\rho z)^{2}}{2(1-q)}\right)}{4\pi\text{H}\left(\frac{\kappa-t\sqrt{q-\rho^{2}}-\rho z}{\sqrt{1-q}}\right)^{2}}
\end{equation}

\begin{equation}
\times\sqrt{\frac{2}{\pi}}e^{-\frac{\left(\kappa-t\sqrt{q-\rho^{2}}-\rho z\right)^{2}}{2(1-q)}}\text{H}\left(-\frac{\Gamma(t,z)}{\sqrt{\left(q-\rho^{2}\right)\Lambda}}\right)
\end{equation}

Changing variables to $t\to t\sqrt{q-\rho^{2}}+\rho z$ and taking
the limit $q\to1$,

\noindent 
\begin{equation}
1-\frac{\rho^{2}+R^{2}-2\rho R\cos\theta}{\sin^{2}\theta}=2\alpha\bigg<\int_{-\infty}^{\kappa}\frac{dt\ e^{-\frac{(t-\rho z)^{2}}{2(1-\rho^{2})}}}{\sqrt{2\pi}\sqrt{1-\rho^{2}}}H\bigg(\frac{\Gamma(t,z)}{\sqrt{1-\rho^{2}}\Lambda}\bigg)(\kappa-t)^{2}\bigg>_{z}
\end{equation}

Where now $\Gamma(t,z)=z(\rho R-\cos\theta)-t(R-\rho\cos\theta).$

\paragraph{Derivative wrt $\rho$,}

\begin{equation}
\frac{\rho-R\cos\theta}{(1-q)\sin^{2}\theta}=\frac{\alpha}{2\pi}\frac{\left(\frac{\rho t}{\sqrt{q-\rho^{2}}}-z\right)\exp\left(-\frac{\left(\kappa-t\sqrt{q-\rho^{2}}-\rho z\right)^{2}}{2(1-q)}-\frac{t^{2}}{2}\right)H\left(-\frac{\Gamma(t,z)}{\sqrt{q-\rho^{2}}\Lambda}\right)}{\sqrt{1-q}\text{H}\left(\frac{\kappa-t\sqrt{q-\rho^{2}}-\rho z}{\sqrt{1-q}}\right)}
\end{equation}

\begin{equation}
+\frac{\sqrt{2}\alpha}{\pi}\bigg(\frac{(R-\rho\cos\theta)\left(z-\frac{\rho t}{\sqrt{q-\rho^{2}}}\right)-\left(t\sqrt{q-\rho^{2}}+\rho z\right)\cos\theta-Rz}{\sqrt{2}\sqrt{q-\rho^{2}}\Lambda}
\end{equation}

\begin{equation}
-\frac{\left(-2\rho\Lambda^{2}-\left(q-\rho^{2}\right)(2\rho-2R\cos\theta)\right)\Gamma(t,z)}{2\sqrt{2}\left(q-\rho^{2}\right)^{3/2}\Lambda^{3}}\bigg)
\end{equation}

\begin{equation}
\times\log\left(\frac{1}{2}\text{erfc}\left(\frac{\kappa-t\sqrt{q-\rho^{2}}-\rho z}{\sqrt{2}\sqrt{1-q}}\right)\right)\exp\left(-\frac{\Gamma(t,z)^{2}}{2\left(q-\rho^{2}\right)\Lambda^{2}}-\frac{t^{2}}{2}\right)
\end{equation}

Integrating the second term by parts,

\begin{equation}
\frac{\rho-R\cos\theta}{(1-q)\sin^{2}\theta}=\frac{\alpha}{\pi}\frac{\left(\frac{\rho t}{\sqrt{q-\rho^{2}}}-z\right)\exp\left(-\frac{\left(\kappa-t\sqrt{q-\rho^{2}}-\rho z\right)^{2}}{2(1-q)}-\frac{t^{2}}{2}\right)H\left(-\frac{\Gamma(t,z)}{\sqrt{q-\rho^{2}}\Lambda}\right)}{\sqrt{1-q}\text{H}\left(\frac{\kappa-t\sqrt{q-\rho^{2}}-\rho z}{\sqrt{1-q}}\right)}
\end{equation}

\begin{equation}
-\frac{\alpha}{\pi\Delta}\exp\left(-\frac{\Delta(t,z)}{2\Lambda^{2}}\right)\left(\frac{\rho R-q\cos\theta}{q-\rho^{2}}\right)
\end{equation}

Changing variables to $t\to t\sqrt{q-\rho^{2}}+\rho z$ and taking
the limit $q\to1$,

\begin{equation}
\frac{\rho-R\cos\theta}{\sin^{2}\theta}=2\alpha\bigg<\int_{-\infty}^{\kappa}dt\frac{e^{-\frac{(t-\rho z)^{2}}{2(1-\rho^{2})}}}{\sqrt{2\pi}\sqrt{1-\rho^{2}}}H\bigg(\frac{\Gamma(t,z)}{\sqrt{1-\rho^{2}}\Lambda}\bigg)\bigg(\frac{z-\rho t}{1-\rho^{2}}\bigg)(\kappa-t)
\end{equation}

\begin{equation}
+\frac{1}{2\pi\Lambda}\exp\left(-\frac{\Delta(t,z)}{2\Lambda^{2}}\right)\bigg(\frac{\rho R-\cos\theta}{1-\rho^{2}}\bigg)(\kappa-t)\bigg>_{z}
\end{equation}

So together we have three saddle point equations:

\noindent %
\noindent\fbox{\begin{minipage}[t]{1\columnwidth - 2\fboxsep - 2\fboxrule}%
\noindent 
\begin{align}
\frac{R-\rho\cos\theta}{\sin^{2}\theta}&=\frac{\alpha}{\pi\Lambda}\bigg<\int_{-\infty}^{\kappa}dt\ \exp\left(-\frac{\Delta(t,z)}{2\Lambda^{2}}\right)(\kappa-t)\bigg>_{z} \\
1-\frac{\rho^{2}+R^{2}-2\rho R\cos\theta}{\sin^{2}\theta}&=2\alpha\bigg<\int_{-\infty}^{\kappa}dt\frac{e^{-\frac{(t-\rho z)^{2}}{2(1-\rho^{2})}}}{\sqrt{2\pi}\sqrt{1-\rho^{2}}}H\bigg(\frac{\Gamma(t,z)}{\sqrt{1-\rho^{2}}\Lambda}\bigg)(\kappa-t)^{2}\bigg>_{z} \\
\frac{\rho-R\cos\theta}{\sin^{2}\theta}&=2\alpha\bigg<\int_{-\infty}^{\kappa}dt\frac{e^{-\frac{(t-\rho z)^{2}}{2(1-\rho^{2})}}}{\sqrt{2\pi}\sqrt{1-\rho^{2}}}H\bigg(\frac{\Gamma(t,z)}{\sqrt{1-\rho^{2}}\Lambda}\bigg)\bigg(\frac{z-\rho t}{1-\rho^{2}}\bigg)(\kappa-t) \\
& \quad \quad \quad +\frac{1}{2\pi\Lambda}\exp\left(-\frac{\Delta(t,z)}{2\Lambda^{2}}\right)\bigg(\frac{\rho R-\cos\theta}{1-\rho^{2}}\bigg)(\kappa-t)\bigg>_{z}
\end{align}

Where 
\begin{align}
\Lambda&=\sqrt{\sin^{2}\theta-R^{2}-\rho^{2}+2\rho R\cos\theta}, \\
\Gamma(t,z)&=z(\rho R-\cos\theta)-t(R-\rho\cos\theta), \\
\Delta(t,z)&=z^{2}\left(\rho^{2}+\cos^{2}\theta-2\rho R\cos\theta\right)+2tz(R\cos\theta-\rho)+t^{2}\sin^{2}\theta.
\end{align}
\end{minipage}}

Solving these equations numerically yields an excellent fit to numerical
simulations on structured data (Fig.~\ref{fig:teacher_student_overlap}\textbf{BCD}).

\section{Model training method details \& dataset information}
\label{app:training_details}

\paragraph{Perceptron in the teacher-student setting}

All code to reproduce these simulations can be found at: \url{https://colab.research.google.com/drive/1in35C6jh7y_ynwuWLBmGOWAgmUgpl8dF?usp=sharing}. Perceptrons were trained on a synthetic dataset of $P$ examples $\{\textbf{x}^\mu,y^{\mu}\}_{\mu=1,\ldots,P}$,
where $\textbf{x}^\mu\sim\mathcal{N}(0,I_{N})$ are i.i.d.~zero mean unit variance random Gaussian inputs, and $y^{\mu}=\text{sign}(\textbf{T}\cdot x)$
are labels generated by a teacher perceptron $\textbf{T}\in\mathbb{R}^{N}$, which
was randomly drawn from a uniform distribution on the sphere $\textbf{T}\sim\text{Unif}(\mathbb{S}^{N-1}(\sqrt{N}))$. For all of our experiments we fixed $N=200$ and set $P=\alpha N$ where $\alpha$ varied between $10^{0.1}$ and $10^{0.5}$. Each synthetic dataset was pruned to keep a fraction $f$ of the smallest-margin examples, where $f$ varied between $0.1$ and $1$ in Fig. \ref{fig:theory} and between $0.2$ and $1$ in Figs. \ref{fig:teacher_student_overlap},\ref{fig:scaling_practice} to match the real-world experiments in Fig. \ref{fig:scaling_practice}. Perceptrons were optimized to find the max-margin separating solution using a standard quadratic programming (QP) algorithm from the CVXPY library (for analysis of the computational complexity of this algorithm see Fig. \ref{fig:compute_scaling}). Results were averaged over $100$ independent draws of the teacher and training examples.

\paragraph{ImageNet.} ImageNet model training was performed using a standard ResNet-50 through the VISSL library \cite{goyal2021vissl} (stable version \texttt{v0.1.6}), which provides default configuration files for supervised ResNet-50 training (accessible \href{https://github.com/facebookresearch/vissl/blob/v0.1.6/configs/config/pretrain/supervised/supervised_8gpu_resnet.yaml}{here}; released under the MIT license). Each model was trained on a single node of 8 NVIDIA V100 32GB graphics cards with BATCHSIZE\_PER\_REPLICA = 256, using the Stochastic Gradient Descent (SGD) optimizer with a base learning rate = 0.1, nesterov momentum = 0.9, and weight decay = 0.001. For our scaling experiments (Fig.~\ref{fig:scaling_practice} and Fig.~\ref{fig:SSL_metric_scaling}), we trained one model per fraction of data kept (0.1-1.0) for each dataset size. In total, these plot required training 97 models on (potentially a subset of) ImageNet. All the models were trained with matched number of iterations, corresponding to 105 epochs on the full ImageNet dataset. The learning rate was decayed by a factor of 10 after the number of iterations corresponding to 30, 60, 90, and 100 epochs on the full ImageNet dataset.

For our main ImageNet experiments (Fig.~\ref{fig:ImageNet-1K_panel}) we trained one model per fraction of data kept (1.0, 0.9, 0.8, 0.7, 0.6) $\times$ metric (11 metrics in total). In the plot itself, since any variation in the ``fraction of data kept = 1.0'' setting is due to random variation across runs not due to potential metric differences, we averaged model performances to obtain a single datapoint here (while also keeping track of the variation across models, which is plotted as $\pm 2$ standard deviations). In total, this plot required training 55 models on (potentially a subset of) ImageNet. For Fig.~\ref{fig:ImageNet-1K_panel}C, in order to reduce noise from random variation, we additionally trained five models per datapoint and metric, and plot the averaged performance in addition to error bars showing one standard deviation of the mean. Numerical results from Figure~\ref{fig:ImageNet-1K_panel}BC are available from Table~\ref{tab:imagenet_pruning_top5}. ImageNet \cite{russakovsky2015imagenet} is released under the \href{https://www.image-net.org/download.php}{ImageNet terms of access}. It is important to note that ImageNet images are often biased \cite{yang2020towards,asano2021pass}. The SWaV model used to compute our prototypicality metrics was obtained via \texttt{torch.hub.load(`facebookresearch/swav:main', `resnet50')}, which is the original model provided by \cite{caron_swav}; we then used the \texttt{avgpool} layer's activations.

\begin{table}[ht]
\centering
\begin{tabular}{@{}lcccc@{}}
\toprule
& \multicolumn{4}{c}{Fraction of data kept}\\
metric                     & 0.9    & 0.8    & 0.7    & 0.6    \\ \midrule
random                     & 89.816 & 89.882 & 89.108 & 88.206 \\
memorization               & 90.209 & 90.664 & 89.837 & 89.032 \\
supervised prototypes      & 90.076 & 90.528 & 89.358 & 88.925 \\
self-supervised prototypes & 90.245 & 90.466 & 89.46  & 88.547 \\
DDD                        & 89.428 & 89.060 & 88.682 & 88.484 \\
EL2N (1 model)             & 90.534 & 90.310 & 89.818 & 88.476 \\
EL2N (20 models)           & 90.348 & 89.698 & 90.352 & 88.674 \\
active learning            & 90.354 & 90.314 & 89.334 & 89.360 \\
forgetting                 & 90.352 & 90.120 & 89.486 & 89.112 \\
influence max              & 90.590 & 89.804 & 88.684 & 88.062 \\
influence sum-abs          & 89.558 & 90.224 & 88.618 & 88.082 \\ \bottomrule
\end{tabular}
\caption{Benchmark table of pruning results (top-5 ImageNet validation accuracy) corresponding to Figure~\ref{fig:ImageNet-1K_panel}BC. Values for ResNet-50 trained with VISSL (baseline unpruned top-5 accuracy: 90.848).}
\label{tab:imagenet_pruning_top5}
\end{table}

\paragraph{CIFAR-10 and SVHN.} CIFAR-10 and SVHN model training was performed using a standard ResNet-18 through the PyTorch library. Each model was trained on a single NVIDIA TITAN Xp 12GB graphics card with batch size = 128, using the Stochastic Gradient Descent (SGD) optimizer with learning rate = 0.1, nesterov momentum = 0.9, and weight decay = 0.0005. Probe models were trained for 20 epochs each for CIFAR-10 and 40 epochs each for SVHN. Pruning scores were then computed using the EL2Ns metric \cite{Paul2021-ci}, averaged across 10 independent initializations of the probe models. To evaluate data pruning performance, fresh models were trained from scratch on each pruned dataset for 200 epochs, with the learning rate decayed by a factor of 5 after 60, 120 and 160 epochs.

\paragraph{Data pruning for transfer learning.} To assess the effect of pruning downstream finetuning data on transfer learning performance, vision transformers (ViTs) pre-trained on ImageNet21k were fine-tuned on different pruned subsets of CIFAR-10. Pre-trained models were obtained from the timm model library \cite{rw2019timm}. Each model was trained on a single NVIDIA TITAN Xp 12GB graphics card with batch size = 128, using the Adam optimizer with learning rate = 1e-5 and no weight decay. Probe models were trained for 2 epochs each. Pruning scores were then computed using the EL2Ns metric \cite{Paul2021-ci}, averaged across 10 independent random seeds. To evaluate data pruning performance, pre-trained models were fine-tuned on each pruned dataset for 10 epochs.

To assess the effect of pruning upstream pretraining data on transfer learning performance, each of the ResNet-50s pre-trained on pruned subsets of ImageNet1k in Fig.~\ref{fig:scaling_practice}\textbf{D} was fine-tuned on all of CIFAR-10. Each model was trained on a single NVIDIA TITAN Xp 12GB graphics card with batch size = 128, using the RMSProp optimizer with learning rate = 1e-4 and no weight decay. Probe models were trained for 2 epochs each. Pruning scores were then computed using the EL2Ns metric \cite{Paul2021-ci}, averaged across 10 independent random seeds. To evaluate data pruning performance, pre-trained models were fine-tuned on each pruned dataset for 10 epochs.

\section{Breaking compute scaling laws via data pruning}
\label{app:compute_scaling}

\begin{figure}[h!]
    \centering
    \includegraphics[width=1\linewidth]{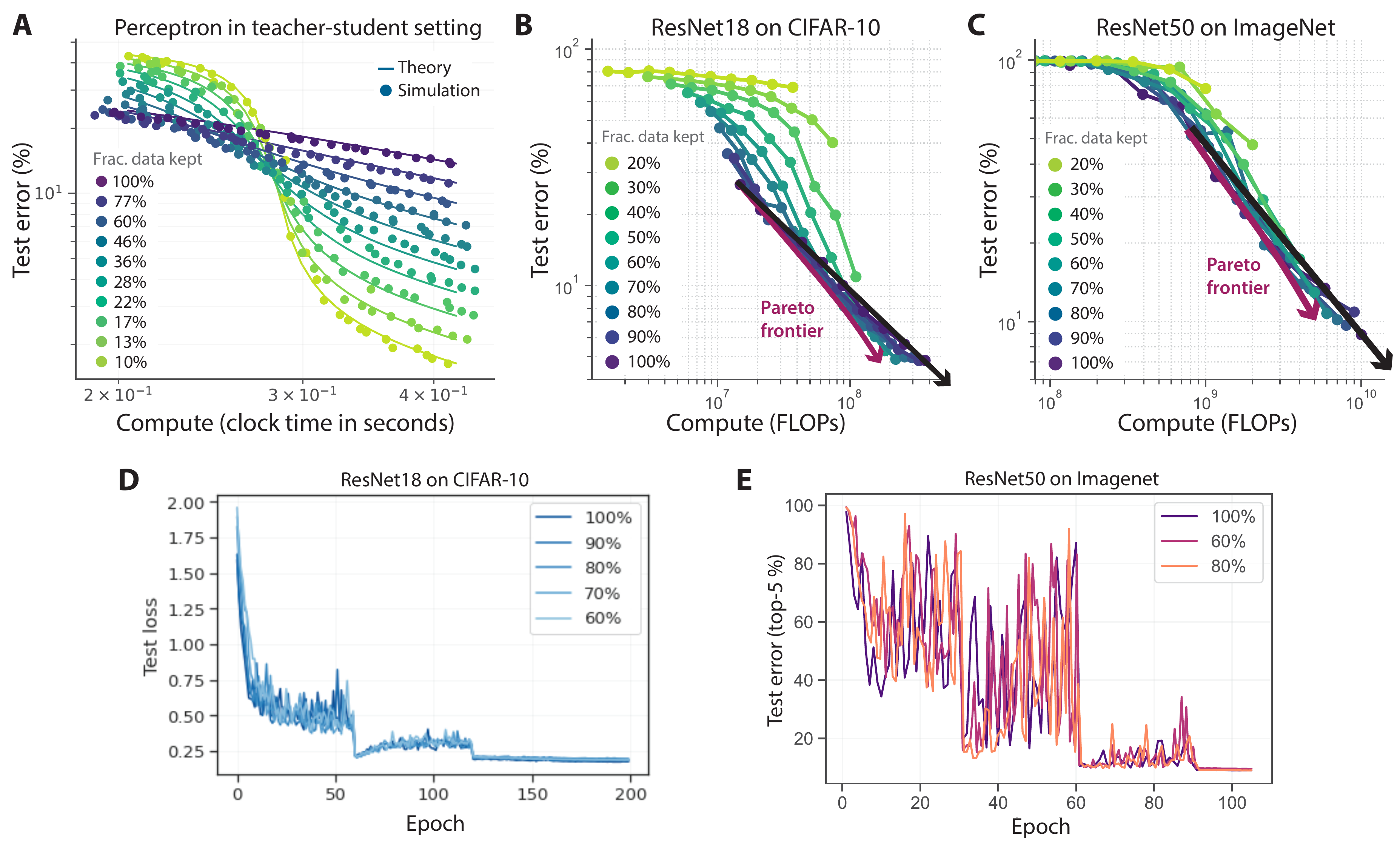}
    \caption{Breaking compute scaling laws via data pruning. \textbf{A,B,C}, We repeat the experiments in Figs. \ref{fig:theory}\textbf{A}, \ref{fig:scaling_practice}\textbf{C,D}, replacing the x-axis with compute, measured as clock time to convergence for the perceptron (\textbf{A}), and FLOPs in a fixed-epoch training setting for the ResNets (\textbf{B,C}). Theoretical curves in \textbf{A} are overlaid by linearly regressing clock time to convergence $T$ from $\alpha_\text{prune}$, with $T=0.96\alpha_\text{prune} + 0.80$. Perceptrons in \textbf{A} are trained on a CPU on a google Colab. \textbf{E,D},  CIFAR-10 and ImageNet learning curves for fixed epochs.}
    \label{fig:compute_scaling}
\end{figure}

Do the savings in training dataset size we have identified translate to savings in compute, and can data pruning be used to beat widely observed compute scaling laws \cite{Kaplan2020-ti,Gordon2021-az,Zhai2021-dl}? Here we show for the perceptron that data pruning can afford exponential savings in compute, and we provide preliminary evidence that the same is true for ResNets trained on CIFAR-10 and ImageNet. We repeat the perceptron learning experiments in Fig. \ref{fig:theory}A, keeping track of the computational complexity of each experiment, measured by the time to convergence of the quadratic programming algorithm used to find a max-margin solution (see \ref{app:training_details} for details). Across all experiments, the convergence time $T$ was linearly proportional to $\alpha_\text{prune}$ with $T=0.96\alpha_\text{prune} + 0.80$, allowing us to replace the x-axis of \ref{fig:theory}A with compute to produce Fig. \ref{fig:compute_scaling}A, which reveals that data pruning can be used to break compute scaling laws for the perceptron.

Motivated by this, we next investigate whether the convergence time of neural networks trained on pruned datasets depends largely on the number of examples and not their difficulty, potentially allowing for exponential compute savings. We investigate the learning curves of a ResNet18 trained on CIFAR-10 and a ResNet50 on ImageNet for several different pruning fractions (Fig. \ref{fig:compute_scaling}B). While previous works have fixed the number of iterations \cite{Paul2021-ci}, here we fix the number of \textit{epochs}, so that the model trained on 60\% of the full dataset is trained for only 60\% the iterations of the model trained on the full dataset, using only 60\% the compute. Nevertheless, we find that the learning curves are strikingly similar across pruning fractions, and appear to converge equally quickly. These results suggest that data pruning could lead to large compute savings in practical settings, and in ongoing experiments we are working to make the analogs of Fig. \ref{fig:compute_scaling}A for ResNets on CIFAR-10 and ImageNet to quantify this benefit.

\section{Additional scaling experiments}
\label{app:additional_scaling_experiments}

\begin{figure}[h!]
    \centering
    \includegraphics[width=\linewidth]{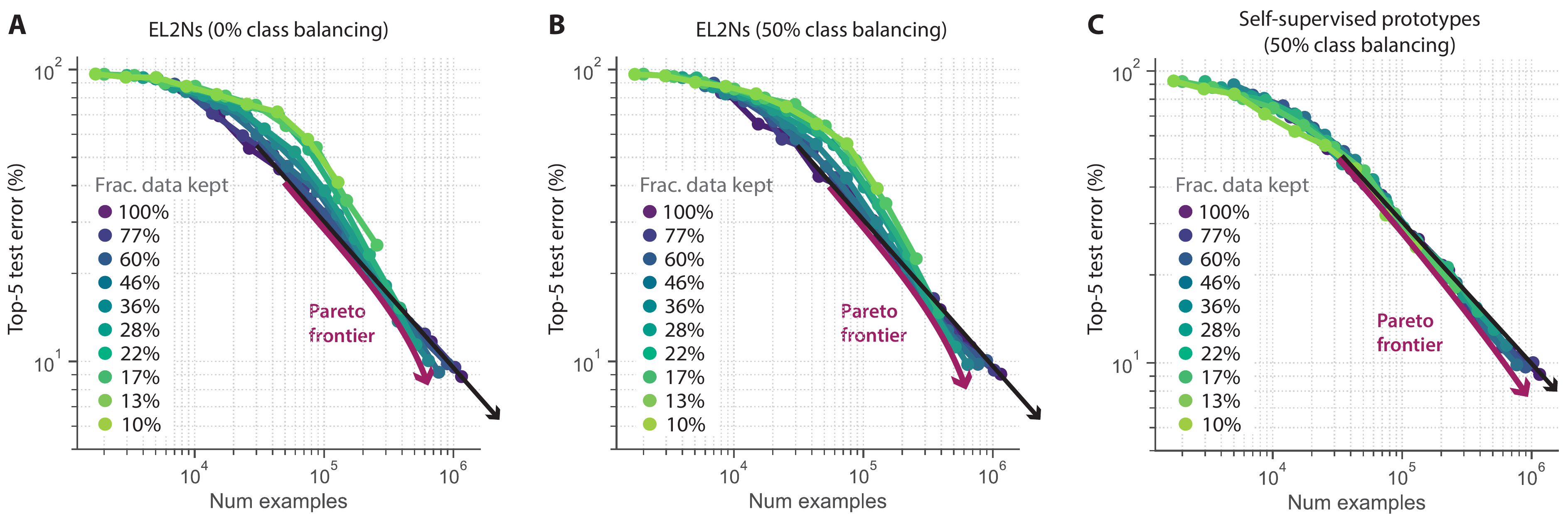}
    \caption{Additional scaling experiments. We reproduce the scaling results on ImageNet in Fig.~\ref{fig:scaling_practice}\textbf{D} using three additional metrics: two supervised and one self-supervised. Each shows some signatures of breaking power law scaling, although the effect is less dramatic than for the best metric, memorization (Fig.~\ref{fig:scaling_practice}\textbf{D}). (\textbf{A}) EL2Ns with a class balancing fraction of 0\% (see App.~Section \ref{app:class_imbalance} for details), (\textbf{B}) EL2Ns with a class balancing fraction of 50\%, and (\textbf{C}) Self-supervised prototypes with a class balancing fraction of 50\%.}
    \label{fig:SSL_metric_scaling}
\end{figure}

\begin{figure}[h!]
    \centering
    \includegraphics[width=0.8\linewidth]{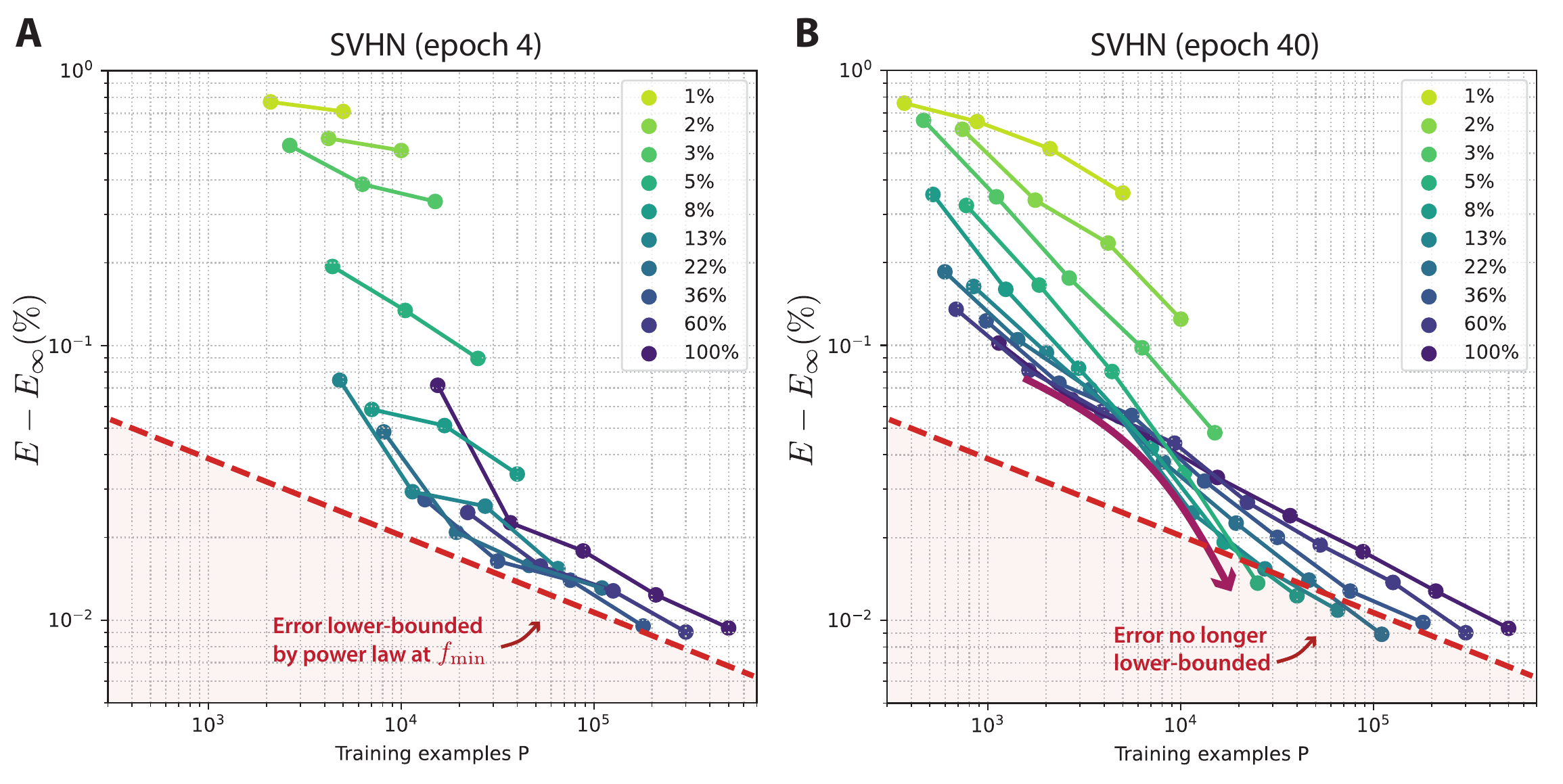}
    \caption{Consistent with a prediction from the perceptron theory (Fig.~\ref{fig:teacher_student_overlap}), when SVHN is pruned using a weak metric (a probe trained for only 4 epochs), the learning curve envelope is lower-bounded by a power law at some $f_\text{min}$ (\textbf{A}). However, with a stronger pruning metric (a probe trained for 40 epochs), the learning curve can break through this power law to achieve lower generalization error (\textbf{B}). }
    \label{fig:svhn_teacher_student_overlap}
\end{figure}

In Fig. \ref{fig:SSL_metric_scaling} we perform additional scaling experiments using the EL2Ns and self-supervised prototypes metrics. In Fig.~\ref{fig:svhn_teacher_student_overlap} we give a practical example of a cross over from exponential to power-law scaling when the probe student has limited information about the teacher (here a model trained for only a small number of epochs on SVHN) . 

\begin{figure}[h!]
    \centering
    \includegraphics[width=0.8\linewidth]{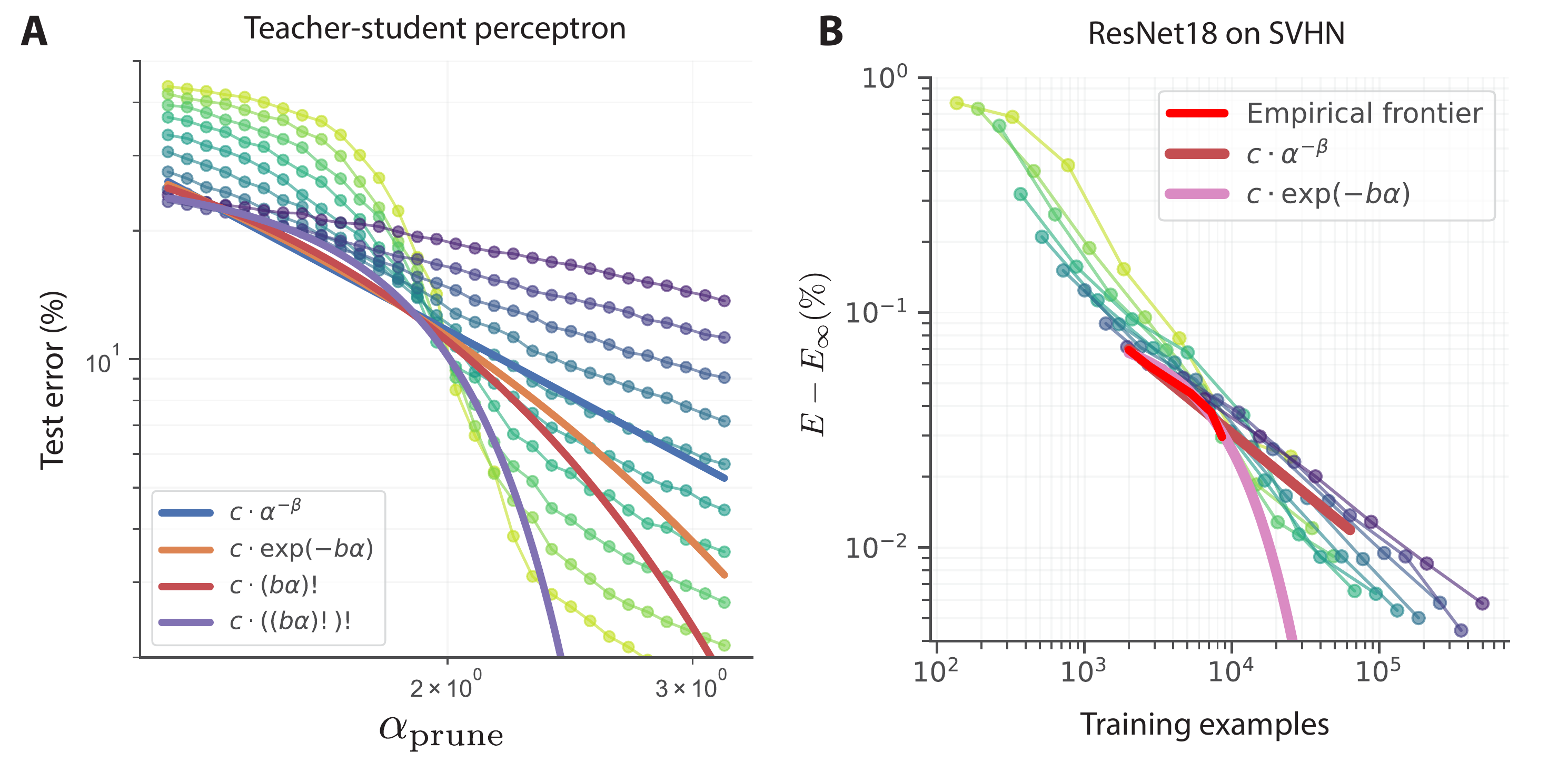}
    \caption{Scaling of the Pareto frontier. (\textbf{A}) Different functional forms fitted to the Pareto frontier of the teacher-student perceptron using least squares: power-law (blue), exponential (orange), factorial (red), and iterated factorial (purple). (\textbf{B}), We estimate the empirical Pareto frontier (red) for ResNet18 on SVHN at higher resolution by training an additional 300 models in the region around the frontier. We then fit two functional forms to the empirical frontier using least squares: power-law (dark red) and exponential (pink). The exponential shows a much better fit to the empirical Pareto frontier, indicating that the scaling is at least exponential.}
    \label{fig:scaling_experiments}
\end{figure}

\section{Extremal images according to different metrics}
\label{app:extreme_images}
In Fig.~\ref{fig:qualitative_prototypical_images}, we showed extremal images for two metrics (self-supervised prototypes, memorization) and a single class. In order to gain a better understanding of how extremal images (i.e.\ images that are easiest or hardest to learn according to different metrics) look like for all metrics and more classes, we here provide additional figures. In order to avoid cherry-picking classes while at the same time making sure that we are visualizing images for very different classes, we here show extreme images for classes 100, ..., 500 while leaving out classes 0 and 400 (which would have been part of the visualization) since those classes almost exclusively consist of images containing people (0: tench, 400: academic gown). The extremal images are shown in Figures~\ref{fig:extreme_images_class_100_I},\ref{fig:extreme_images_class_100_II},\ref{fig:extreme_images_class_200_I},\ref{fig:extreme_images_class_200_II},\ref{fig:extreme_images_class_300_I},\ref{fig:extreme_images_class_300_II},\ref{fig:extreme_images_class_500_I},\ref{fig:extreme_images_class_500_II}.%

\newcommand{\ExtremeImages}[6]{
    \begin{figure}[#5]
        \foreach \metricName in #3{
    	    \begin{subfigure}{1.0\linewidth}
    		    	\centering
    			    \includegraphics[width=\linewidth]{figures/empirical/extreme-images/ImageNet-1K_extreme-images_class-index-#1_\metricName_n-imgs-#6.pdf}
        	\end{subfigure}
    	}\hfill
    	\caption{Extreme images according to different metrics for ImageNet class #1 (\texttt{#2}). For each metric, the top row shows images that are ranked as ``easy'' (most pruneable) according to the metric, and the bottom row shows images that are ranked as ``hard'' (least pruneable).}
    	\label{fig:extreme_images_class_#1_#4}
    \end{figure}
}

\newcommand*\MetricListI{random-100,influence-max,influence-sum-abs,SSL-k-means-SWAV-25-percent,SSL-prototypes,EL2N}
\newcommand*\MetricListII{DDD,EL2N-epochs-20-seeds-20,memorization,active-learning,forgetting}

\ExtremeImages{100}{black swan}{\MetricListI}{I}{h!}{8}
\ExtremeImages{100}{black swan}{\MetricListII}{II}{h!}{8}

\ExtremeImages{200}{Tibetan terrier}{\MetricListI}{I}{h!}{8}
\ExtremeImages{200}{Tibetan terrier}{\MetricListII}{II}{h!}{8}

\ExtremeImages{300}{tiger beetle}{\MetricListI}{I}{h!}{8}
\ExtremeImages{300}{tiger beetle}{\MetricListII}{II}{h!}{8}

\ExtremeImages{500}{cliff dwelling}{\MetricListI}{I}{h!}{8}
\ExtremeImages{500}{cliff dwelling}{\MetricListII}{II}{h!}{8}

\clearpage

\section{Impact of number of clusters $k$ on self-supervised prototypes}
\label{app:clusters}
Our self-supervised prototype metric is based on $k$-means clustering, which has a single hyperparameter $k$. By default and throughout the main paper, we set $k=1000$, corresponding to the number of classes in ImageNet. Here, we investigate other settings of $k$ to understand how this hyperparameter impacts performance. As can be seen in Table~\ref{tab:cluster_impact}, $k$ does indeed have an impact on performance, and very small values for $k$ (e.g.\ $k<10$) as well as very large values for $k$ (e.g.\ $k=50,000$) both lead to performance impairments. At the same time, performance is relatively high across very different in-between settings for $k$. In order to assess these results, it may be important to keep in mind that $\pm 0.54\%$ corresponds to plus/minus 2 standard deviations of performance when simply training the same model multiple times (with different random initialization). Overall, these results suggest that if the number of clusters $k$ deviates at most by one order of magnitude from the number of classes in the dataset (for ImageNet-1K), the exact choice of $k$ does not matter much.

\begin{table}[h!]
\setlength{\tabcolsep}{3.1pt}{
\begin{tabular}{lrrrrrrrrrrrrr}
\toprule
$k$ &  1 &  5 & 10 & 50 & 100 & 200 & 400 & 600 & 800 & 1K & 5K & 10K & 50K\\
\midrule
$acc$   & 88.35 & 89.09 & 88.76 & 89.04 &  89.48 &  90.27 &  89.85 &  89.96 &  90.44 &   90.56 &   90.33 &    90.57 &    88.67\\
\bottomrule
\end{tabular}}\\
\caption{Performance (top-5 accuracy, denoted as $acc$) when pruning away 20\% of ImageNet according to our self-supervised prototype metric as a function of the number of prototypes $k$ (hyperparameter used for self-supervised clustering indicating the number of clusters). Results based on training a ResNet-50 architecture with VISSL without class balancing.}
\label{tab:cluster_impact}
\end{table}

\section{Impact of ensemble prototypes}
The self-supervised prototypes metric is based on $k$-means clustering in the embedding space of a self-supervised (=SSL) model. Since even otherwise identical models trained with different random seeds can end up with somewhat different embedding spaces, we here investigated how the performance of our self-supervised prototypes metric would change when averaging the scores derived from five models, instead of just using a single model's score. The results, shown in Table~\ref{tab:prototype_ensembles}, indicate that ensembling the self-supervised prototype scores neither improves nor hurts performance. This is both good and bad news: Bad news since naturally any improvement in metric development leads to better data efficiency; on the other hand this is also good news since ensembles increase the computational cost of deriving the metric---and this suggests that ensembling is not necessary to achieve the performance we achieved (unlike in other methods such as ensemble active learning).

\begin{table}[h!]
\centering
\begin{tabular}{lrrr}
\toprule
Fraction of data kept &  0.9 & 0.8 & 0.7\\
\midrule
score from single model   & 90.03 & 90.10 & 89.38\\
score from ensemble model & 89.93 & 90.40 & 89.38\\
\bottomrule
\end{tabular}\\
\caption{Performance (top-5 acuracy) when pruning away 10\%, 20\% or 30\% of ImageNet based on the self-supervised prototype metric derived from either a single SSL model, or an ensemble of SSL models. Supervised model training on the pruned dataset was performed with a ResNet-50 architecture trained using VISSL without class balancing. Overall, we do not observe any performance difference between the two settings: the differences are still well within a single standard deviation of training multiple models with identical settings. In order to reduce the influence of random variations, each data point in this table shows the average of four independent runs.}
\label{tab:prototype_ensembles}
\end{table}

\section{Relationship between pruning and class (im-)balance}
\label{app:class_imbalance}
\paragraph{Motivation.} It is well-known that strong class imbalance in a dataset is a challenge that needs to be addressed. In order to understand the effect of pruning on class (im-)balance, we quantified this relationship. For context, if pruning according to a certain metric preferentially leads to discarding most (or even all) images from certain classes, it is likely that the performance on those classes will drop as a result if this imbalance is not addressed.

\paragraph{Class imbalance metric: pruning amplifies class imbalance.} Since a standard measure of class (im-)balance---dividing the number of images for the majority class by the number of images for the minority class---is highly sensitive to outliers and discards information about the 998 non-extreme ImageNet classes, we instead calculated a class balance score $ b\in [0\%, 100\%]$ as the average class imbalance across any two pairs of classes by computing the expectation over taking two random classes, and then computing how many images the minority class has in proportion to the majority class. For instance, a class balance score of 90\% means that on average, when selecting two random classes from the dataset, the smaller of those two classes contains 90\% of the number of images of the larger class (higher=better; 100\% would be perfectly balanced).

\begin{figure}[h!]
	\begin{subfigure}{.33\linewidth}
			\centering
			\includegraphics[width=\linewidth]{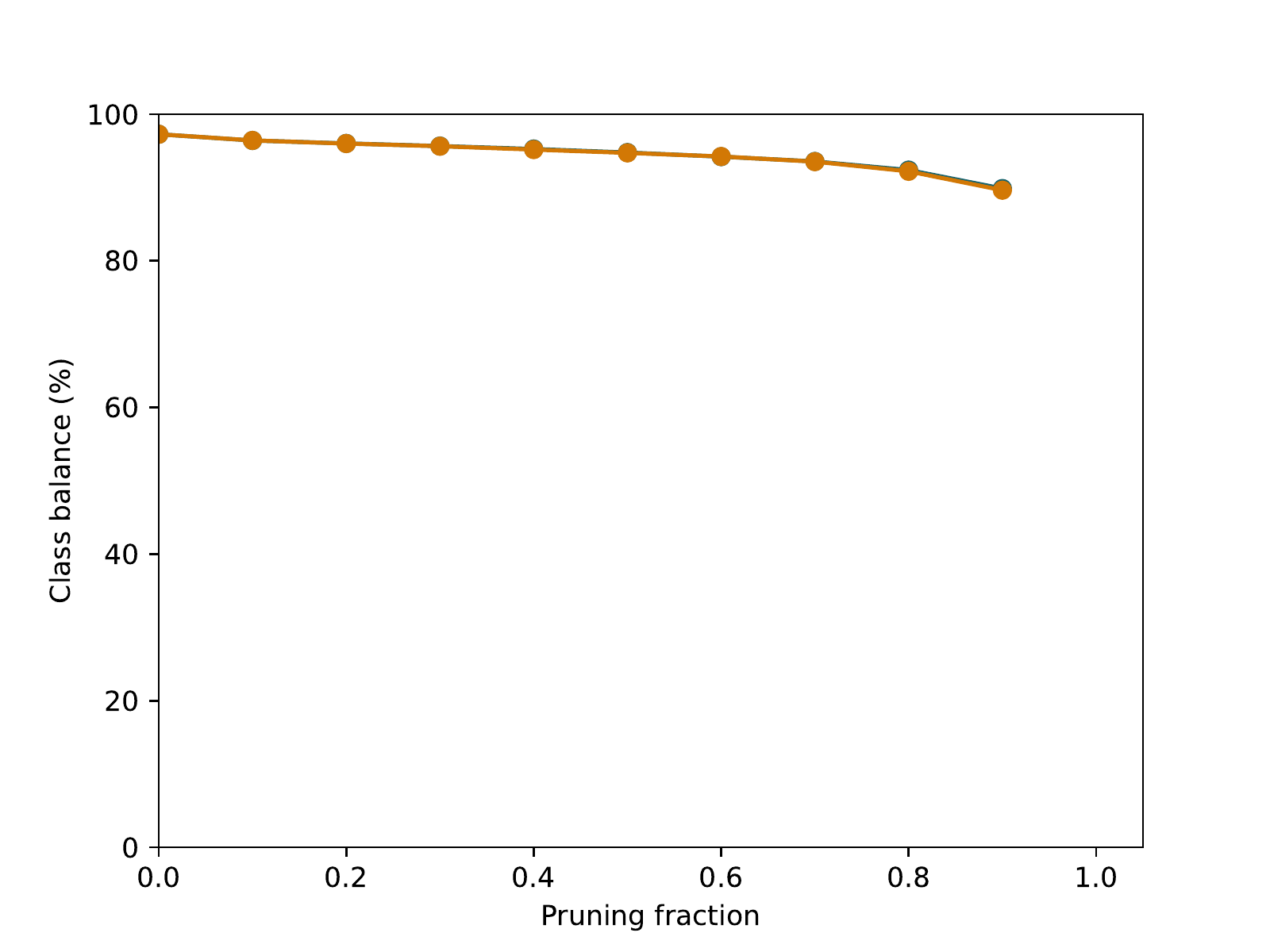}
			\caption{random}
	\end{subfigure}
	\begin{subfigure}{.33\linewidth}
			\centering
			\includegraphics[width=\linewidth]{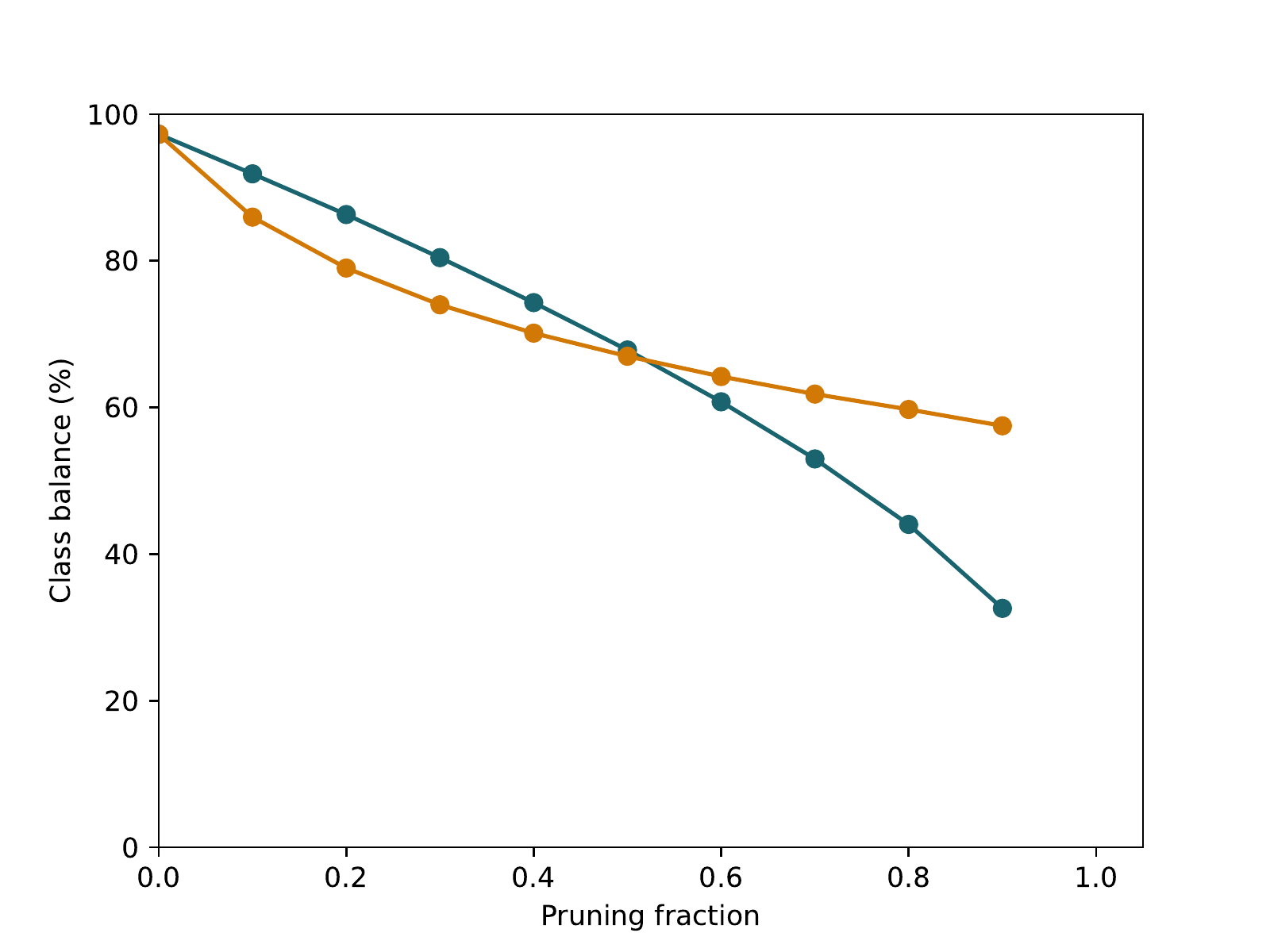}
			\caption{influence max}
	\end{subfigure}
	\begin{subfigure}{.33\linewidth}
			\centering
			\includegraphics[width=\linewidth]{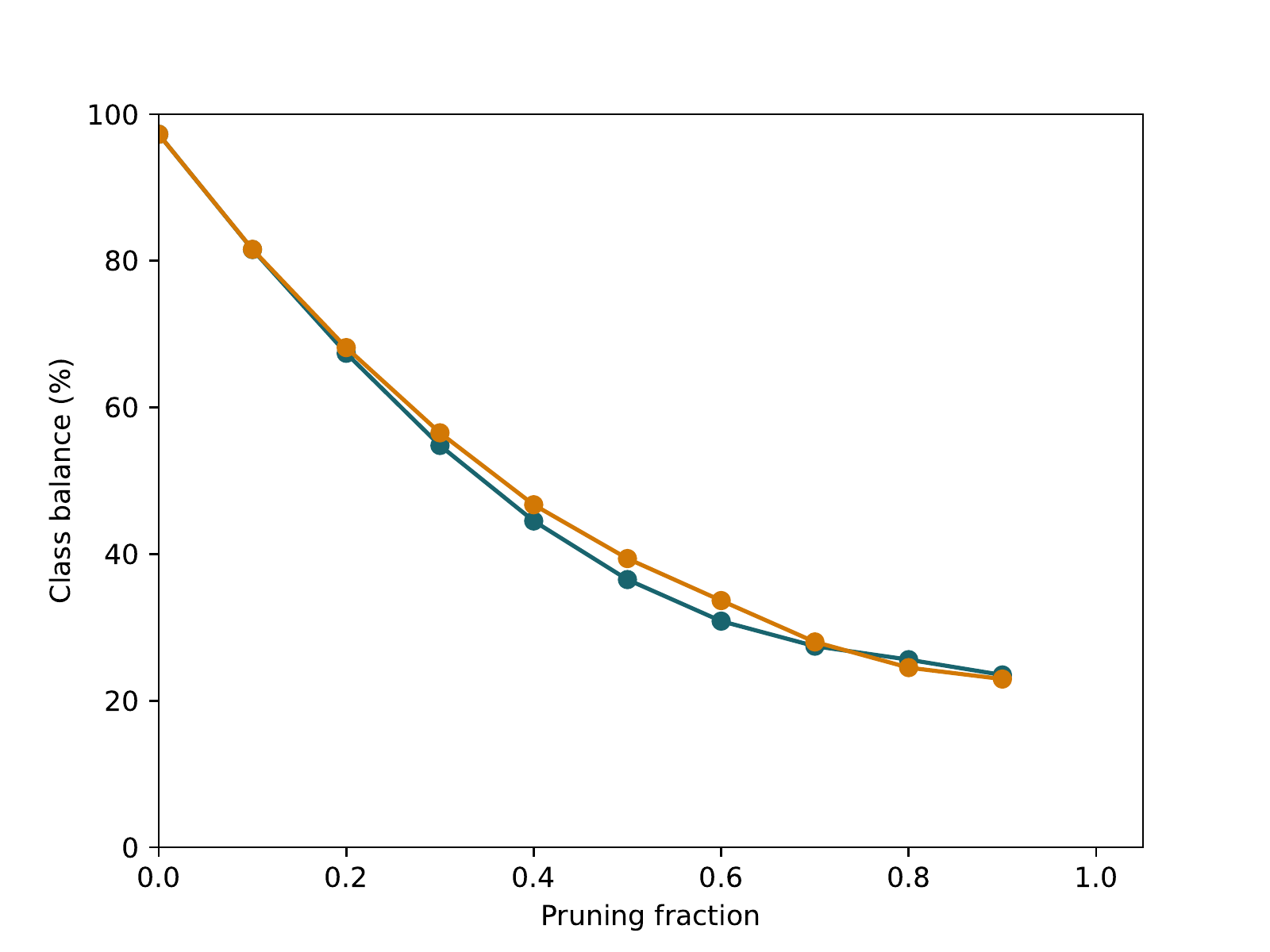}
			\caption{influence sum-abs}
	\end{subfigure}
	\begin{subfigure}{.33\linewidth}
			\centering
			\includegraphics[width=\linewidth]{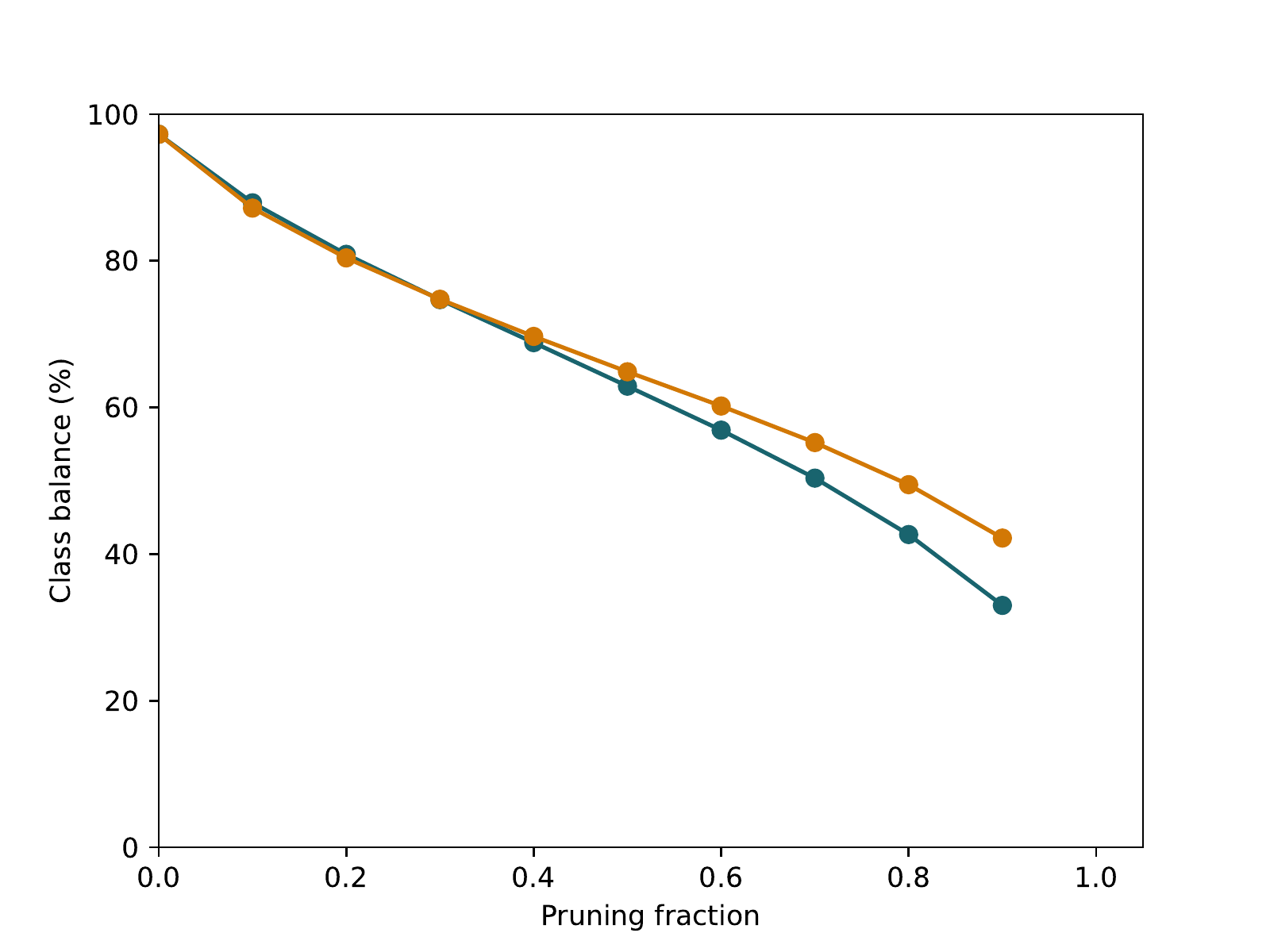}
			\caption{self-supervised prototypes}
	\end{subfigure}\hfill
	\begin{subfigure}{.33\linewidth}
			\centering
			\includegraphics[width=\linewidth]{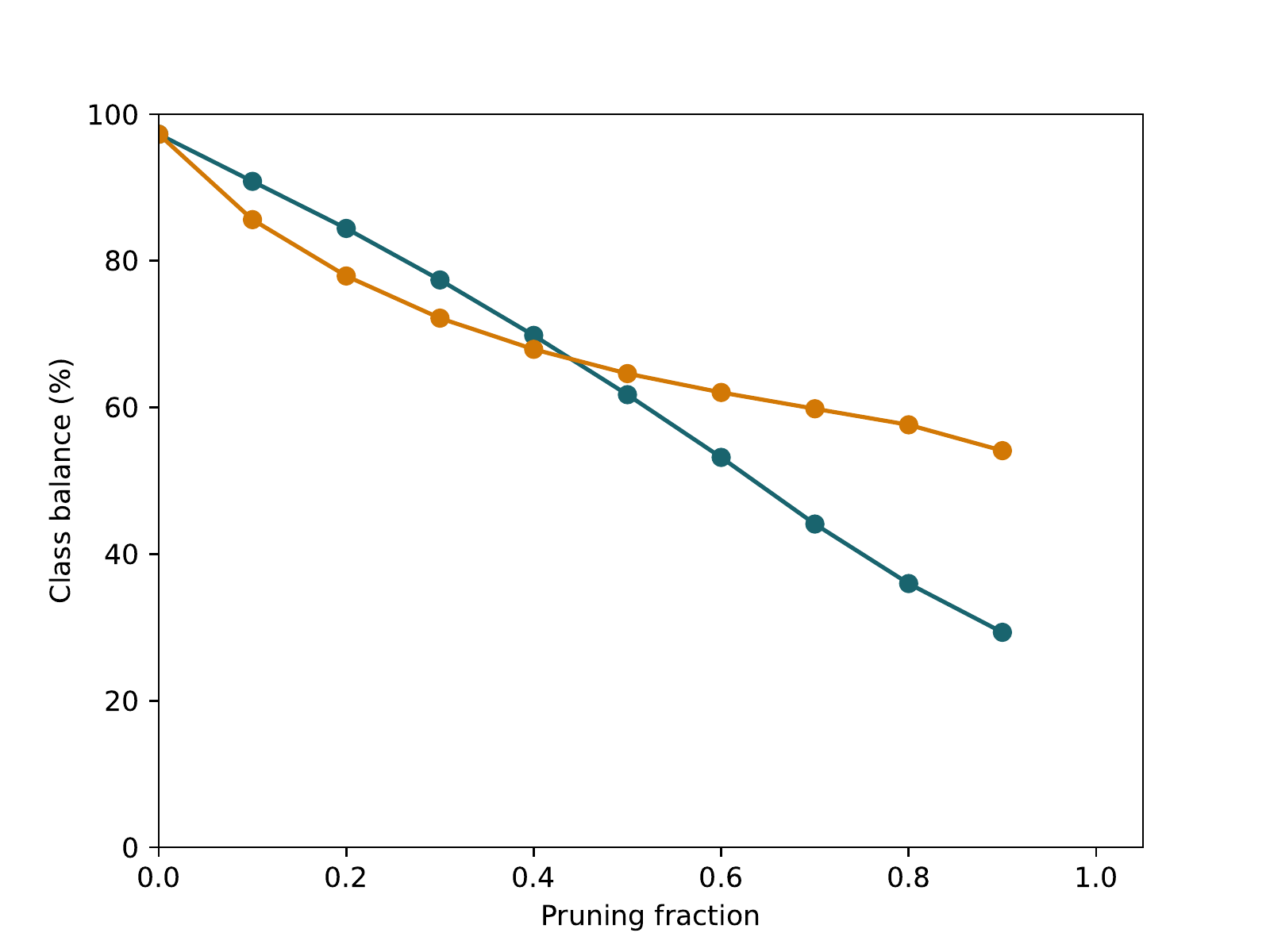}
			\caption{supervised prototypes}
	\end{subfigure}
	\begin{subfigure}{.33\linewidth}
			\centering
			\includegraphics[width=\linewidth]{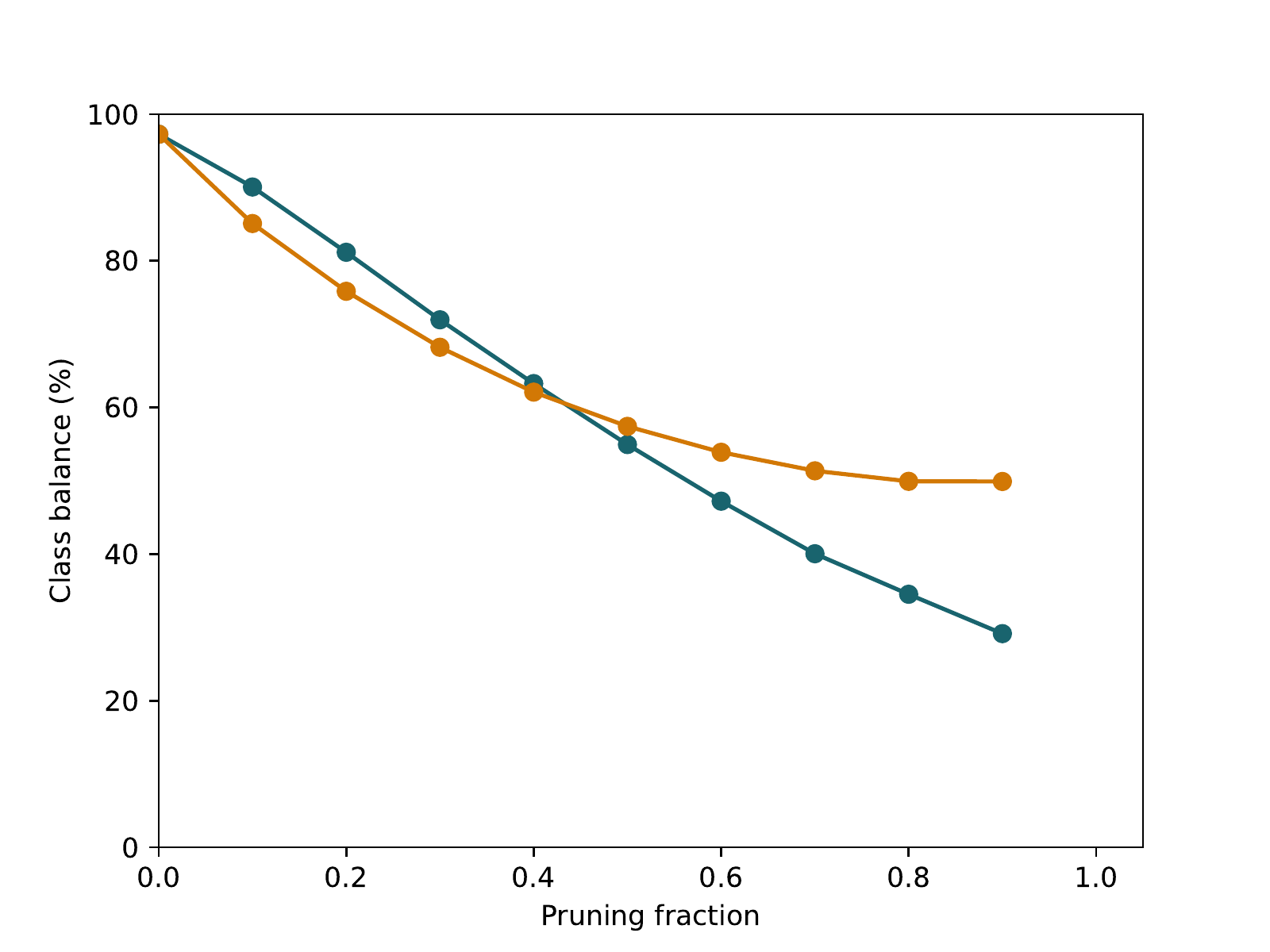}
			\caption{EL2N (1 model)}
	\end{subfigure}
	\begin{subfigure}{.33\linewidth}
			\centering
			\includegraphics[width=\linewidth]{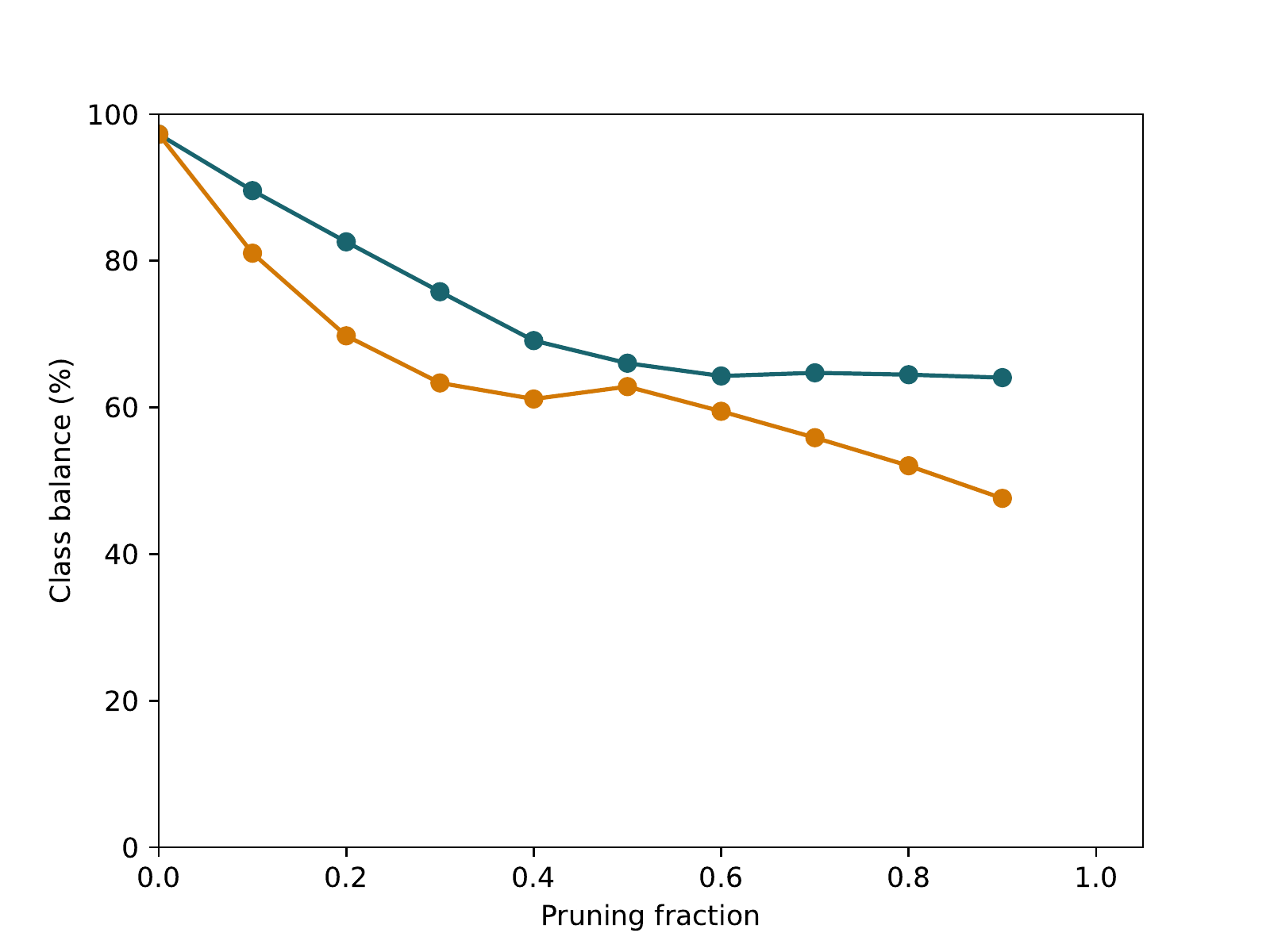}
			\caption{DDD}
	\end{subfigure}
	\begin{subfigure}{.33\linewidth}
			\centering
			\includegraphics[width=\linewidth]{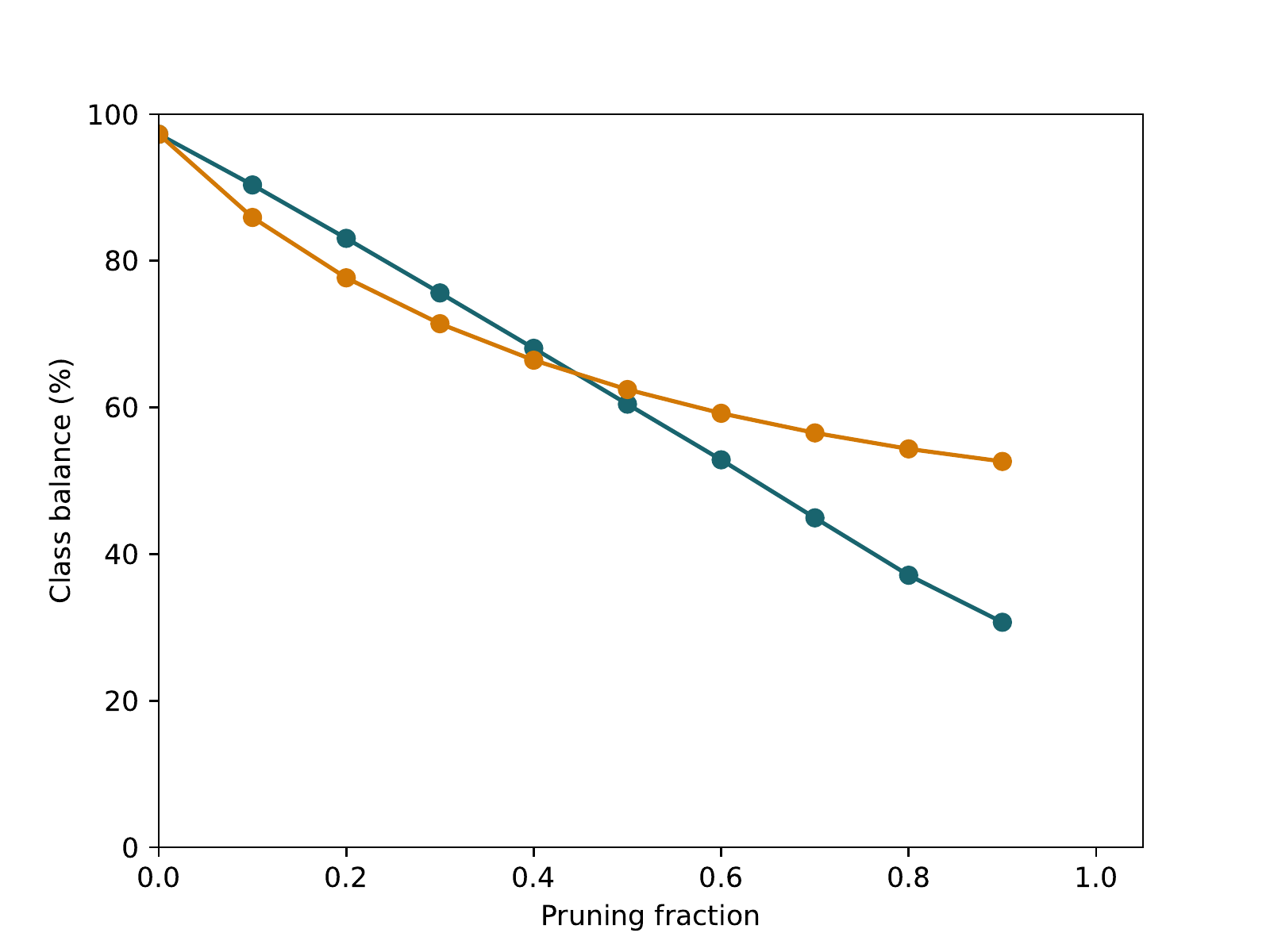}
			\caption{EL2N (20 models)}
	\end{subfigure}
	\begin{subfigure}{.33\linewidth}
			\centering
			\includegraphics[width=\linewidth]{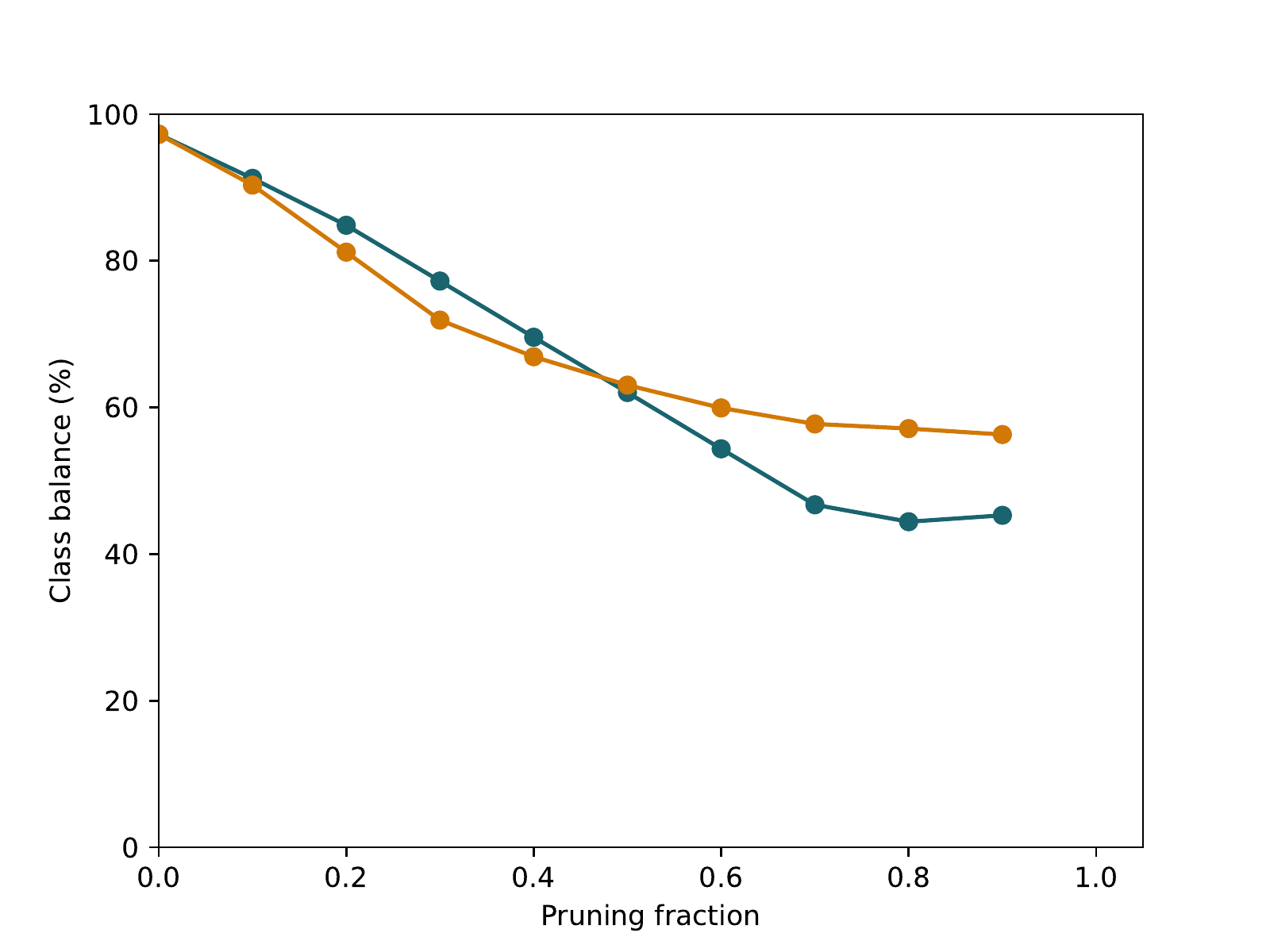}
			\caption{memorization}
	\end{subfigure}
	\begin{subfigure}{.33\linewidth}
			\centering
			\includegraphics[width=\linewidth]{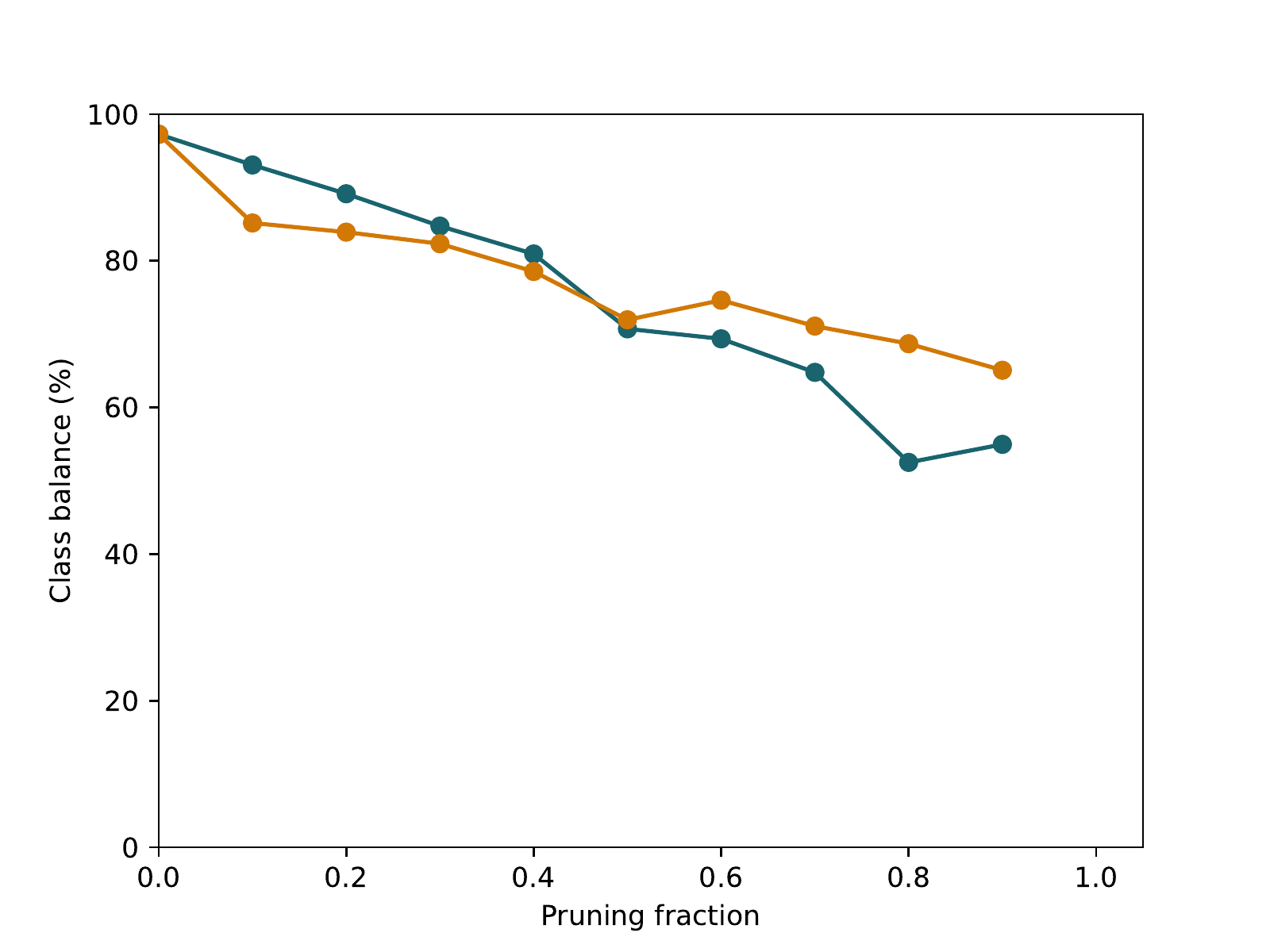}
			\caption{active learning}
	\end{subfigure}
	\begin{subfigure}{.33\linewidth}
			\centering
			\includegraphics[width=\linewidth]{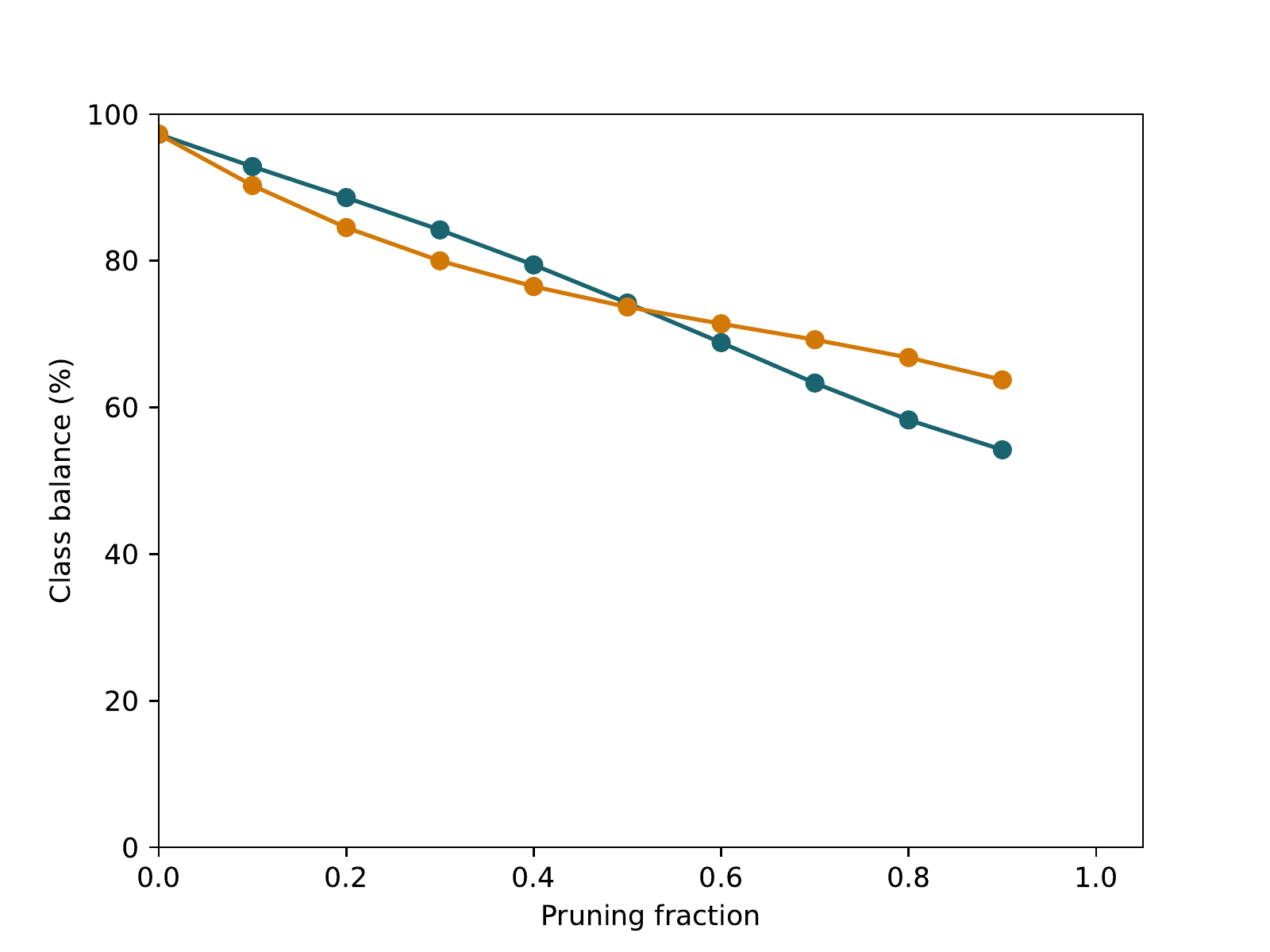}
			\caption{forgetting}
	\end{subfigure}\hfill
	\caption{Pruning amplifies class imbalance. With larger pruning fractions, class balance decreases---both when pruning ``easy'' images (turquoise) and when pruning ``hard'' images (orange). This effect occurs for all pruning metrics except for random pruning (top left). For details on the class imbalance metric see Appendix~\ref{app:class_imbalance}. Pruning fraction refers to the fraction of data pruned, from 0 (no pruning) to 0.9 (keeping only 10\% of the data).}
	\label{fig:class_imbalance}
\end{figure}

In Fig.~\ref{fig:class_imbalance}, we observe that dataset pruning strongly increases class imbalance. This is the case both when pruning away easy images and when pruning away hard images, and the effect occurs for all pruning metrics except, of course, for random pruning. Class imbalance is well-known to be a challenge for deep learning models when not addressed properly \cite{johnson2019survey}. The cause for the amplified class imbalance is revealed when looking at class-conditional differences of metric scores (Figs.~\ref{fig:class_conditional_scores_I} and \ref{fig:class_conditional_scores_II}): The histograms of the class-conditional score distributions show that for many classes, most (if not all) images have very low scores, while for others most (if not all) images have very high scores. This means that as soon as the lowest / highest scoring images are pruned, certain classes are pruned preferentially and thus class imbalance worsens. 

We thus use 50\% class balancing for our ImageNet experiments. This ensures that every class has at least 50\% of the images that it would have when pruning all classes equally (essentially providing a fixed floor for the minimum number of images per class). This simple fix is an important step to address and counteract class imbalance, although other means could be used as well; and ultimately one would likely want to use a self-supervised version of class (or cluster) balancing when it comes to pruning large-scale unlabeled datasets. For comparison purposes, the results for ImageNet pruning \emph{without} class balancing are shown in supplementary Fig.~\ref{fig:pruning_results_unbalanced}.

\begin{figure}[h!]
    \includegraphics[width=\linewidth]{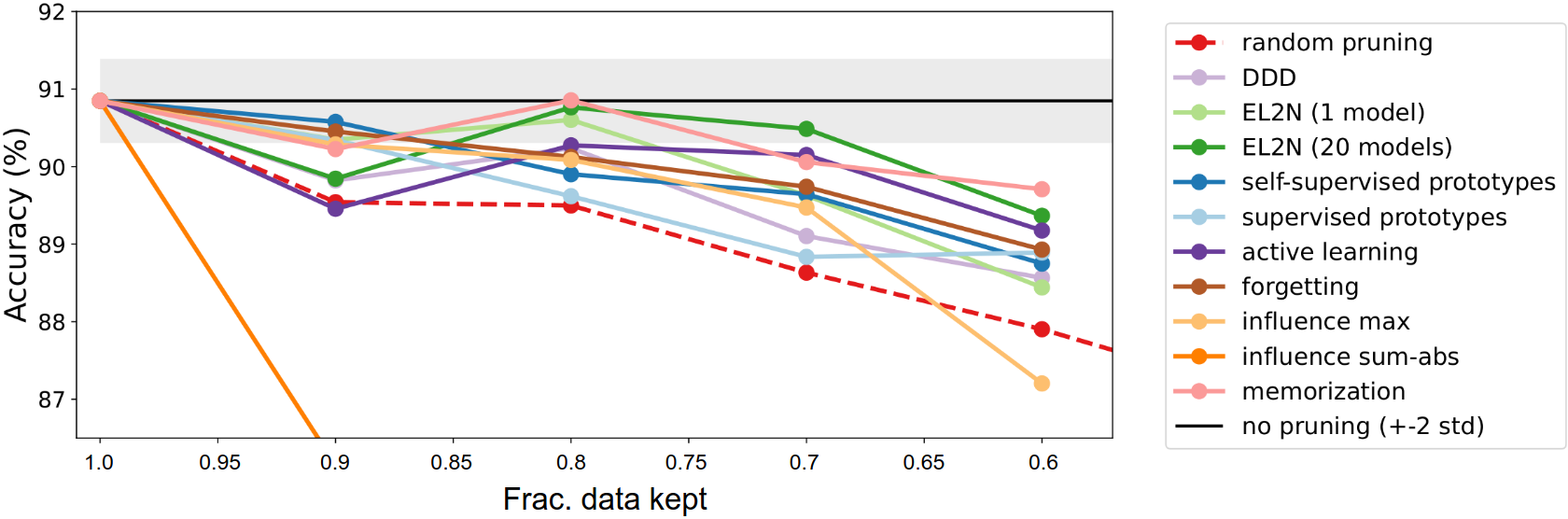}
\caption{ImageNet-1K pruning results for different metrics, compared against random pruning. `Pruning fraction' refers to the fraction of the dataset that is pruned away. Results obtained without any class balancing are worse than results with 50\% class balancing (Fig.~\ref{fig:ImageNet-1K_panel}BC), confirming the finding that vanilla pruning amplifies class imbalance.}
\label{fig:pruning_results_unbalanced}
\end{figure}

\begin{figure*}[h!]
	\begin{subfigure}{1.0\linewidth}
			\centering
			\includegraphics[width=\linewidth]{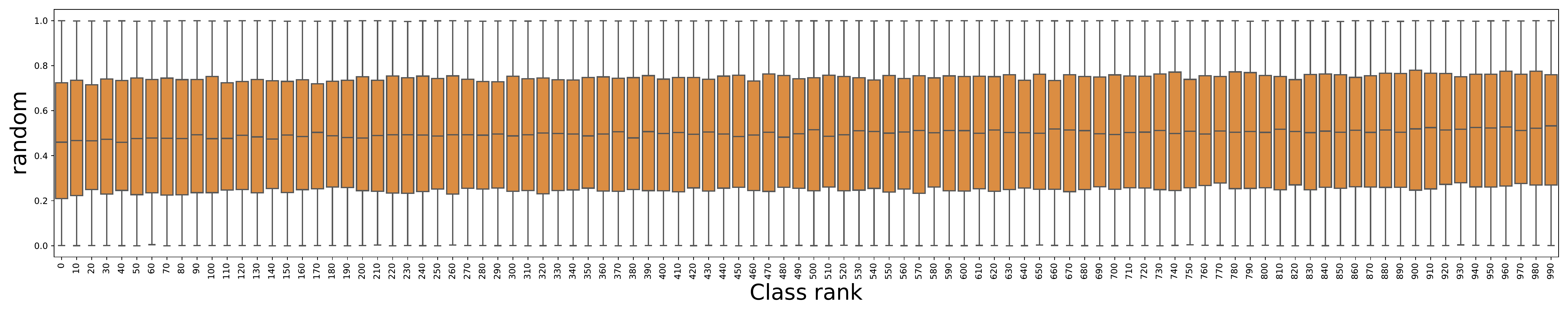}
	\end{subfigure}
	\begin{subfigure}{1.0\linewidth}
			\centering
			\includegraphics[width=\linewidth]{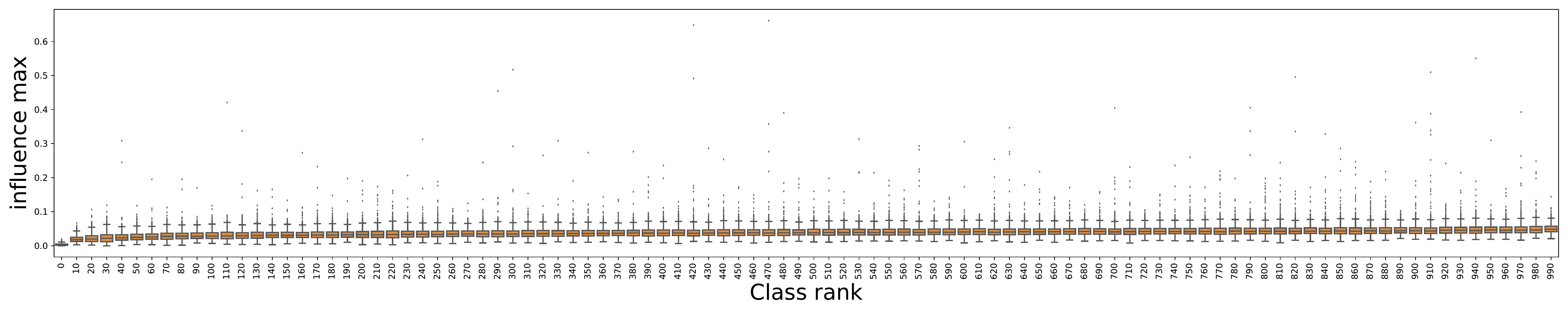}
	\end{subfigure}
	\begin{subfigure}{1.0\linewidth}
			\centering
			\includegraphics[width=\linewidth]{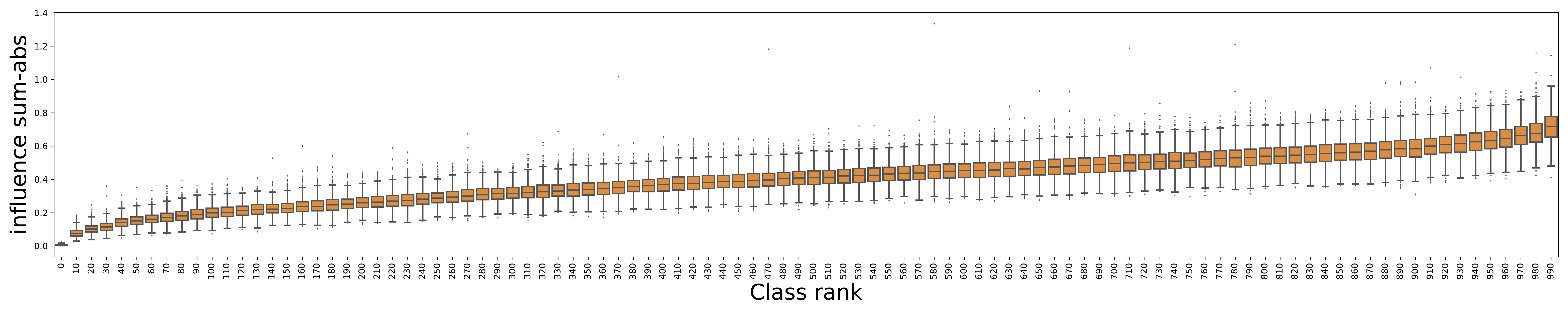}
	\end{subfigure}
	\begin{subfigure}{1.0\linewidth}
			\centering
			\includegraphics[width=\linewidth]{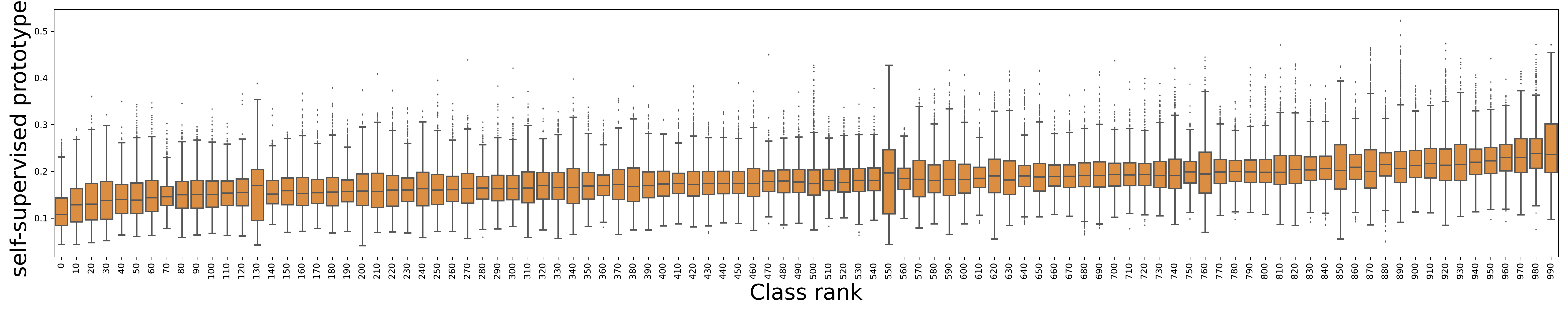}
	\end{subfigure}
	\begin{subfigure}{1.0\linewidth}
			\centering
			\includegraphics[width=\linewidth]{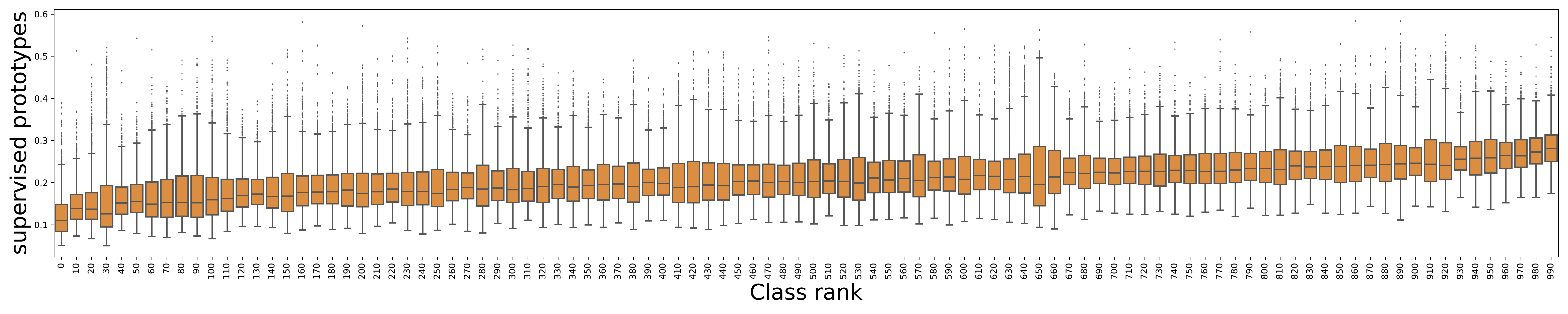}
	\end{subfigure}
	\begin{subfigure}{1.0\linewidth}
			\centering
			\includegraphics[width=\linewidth]{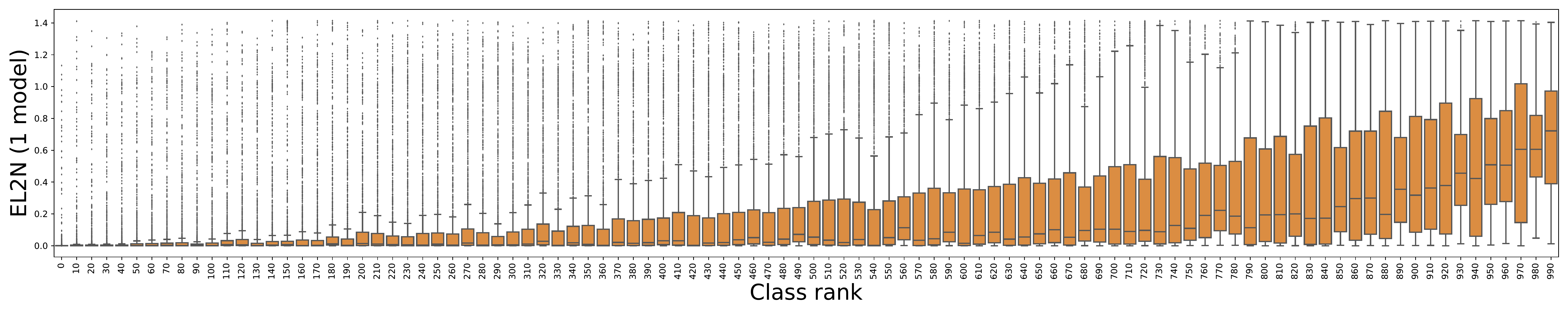}
	\end{subfigure}\hfill
	\caption{How do ImageNet metric scores differ across classes? Class-conditional score distribution histograms across metrics. For the purpose of visualization, only every $10^{th}$ class is shown.}
	\label{fig:class_conditional_scores_I}
\end{figure*}

\begin{figure*}[h!]
	\begin{subfigure}{1.0\linewidth}
			\centering
			\includegraphics[width=\linewidth]{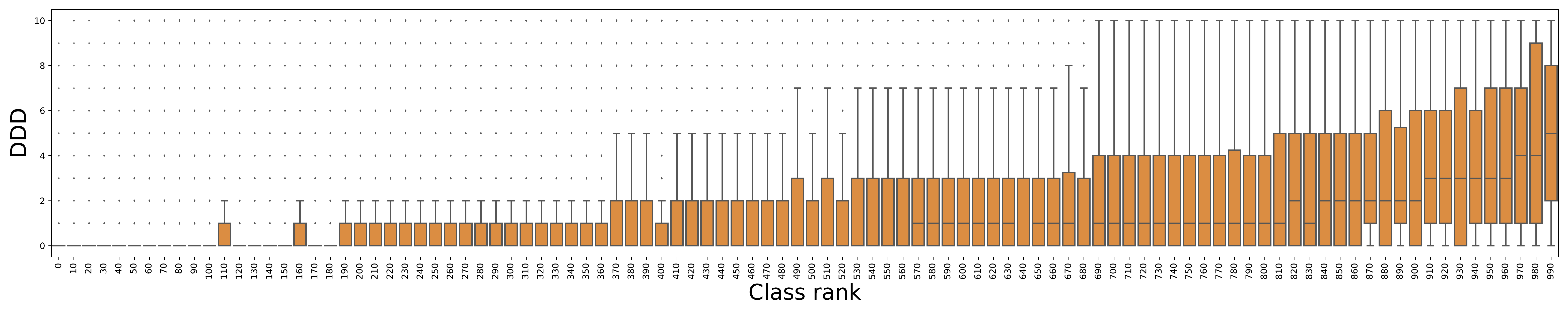}
	\end{subfigure}
	\begin{subfigure}{1.0\linewidth}
			\centering
			\includegraphics[width=\linewidth]{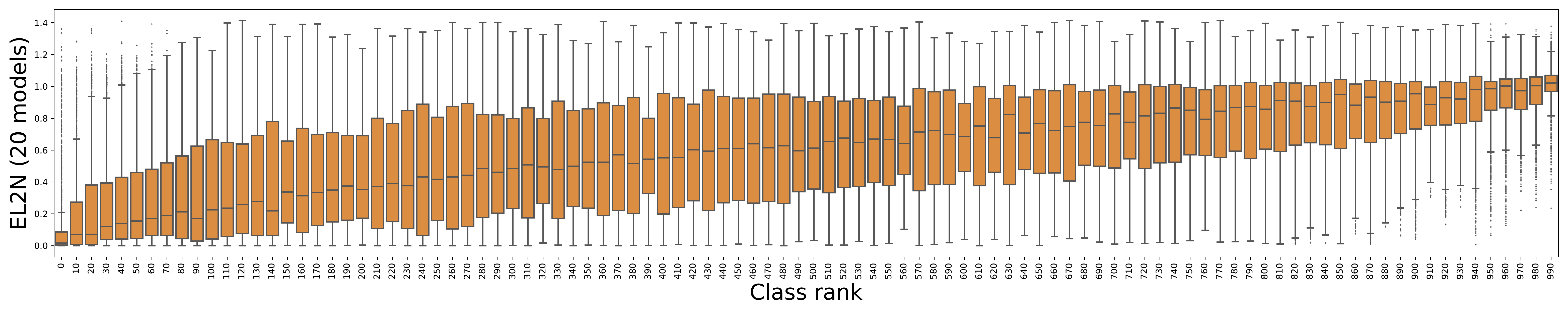}
	\end{subfigure}
	\begin{subfigure}{1.0\linewidth}
			\centering
			\includegraphics[width=\linewidth]{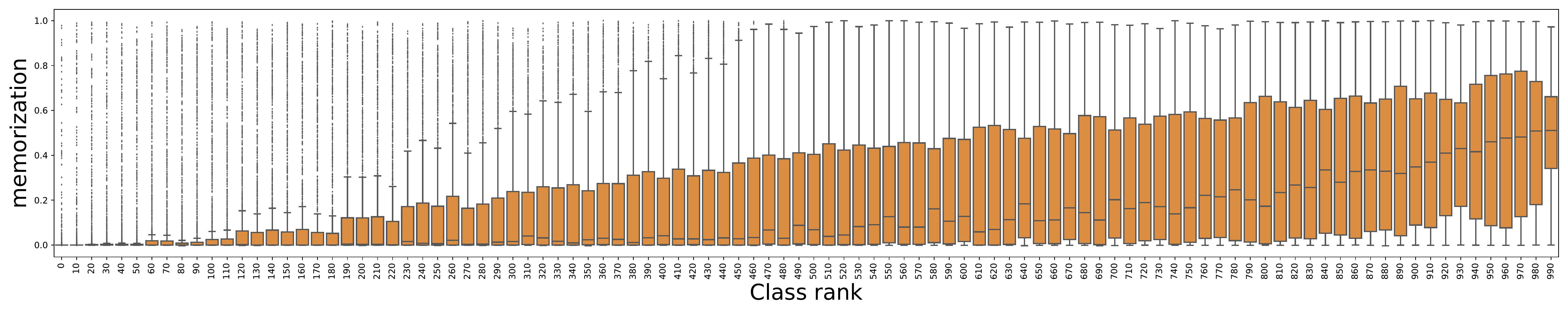}
	\end{subfigure}
	\begin{subfigure}{1.0\linewidth}
			\centering
			\includegraphics[width=\linewidth]{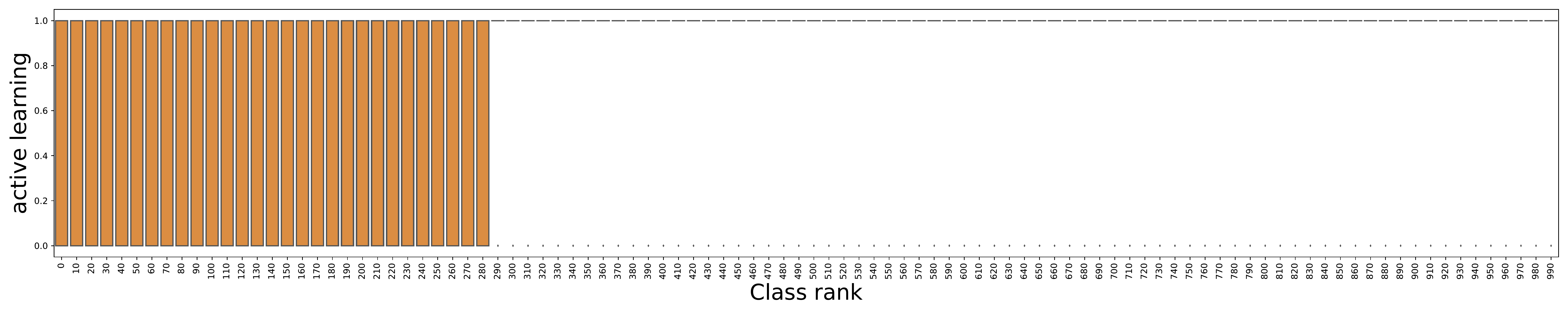}
	\end{subfigure}
	\begin{subfigure}{1.0\linewidth}
			\centering
			\includegraphics[width=\linewidth]{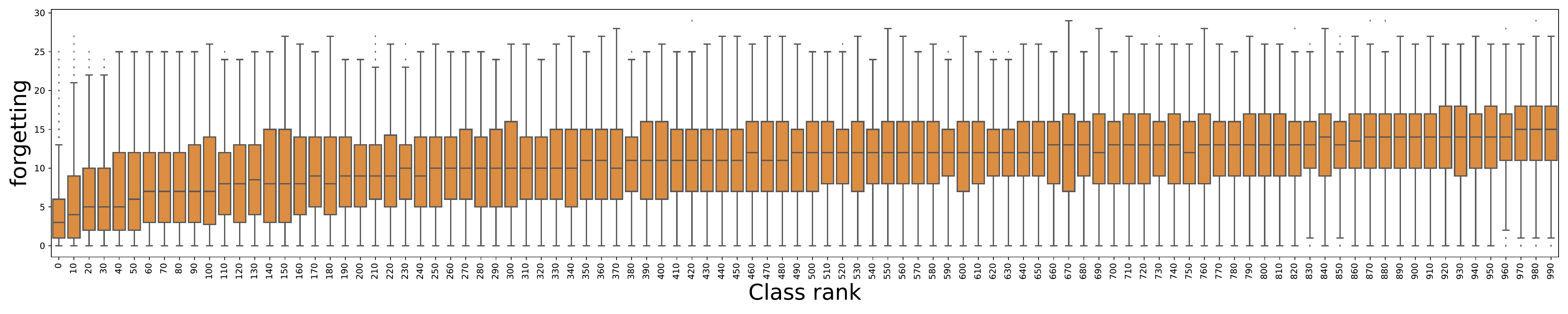}
	\end{subfigure}\hfill	
	\caption{How do ImageNet metric scores differ across classes (continued)? Class-conditional score distribution histograms across metrics. For the purpose of visualization, only every $10^{th}$ class is shown. Please note that the active learning plot is to be taken with a grain of salt since the authors provided a binary score (included/excluded) for ImageNet corresponding to 80\% ``important'' images, thus the scores are either zero or one and a boxplot fit does not apply here.}
	\label{fig:class_conditional_scores_II}
\end{figure*}

\FloatBarrier

\section{Effect of pruning on class-conditional accuracy and fairness}
\label{app:fairness}

\begin{figure*}[h!]
	\begin{subfigure}{.33\linewidth}
			\centering
			\includegraphics[width=\linewidth]{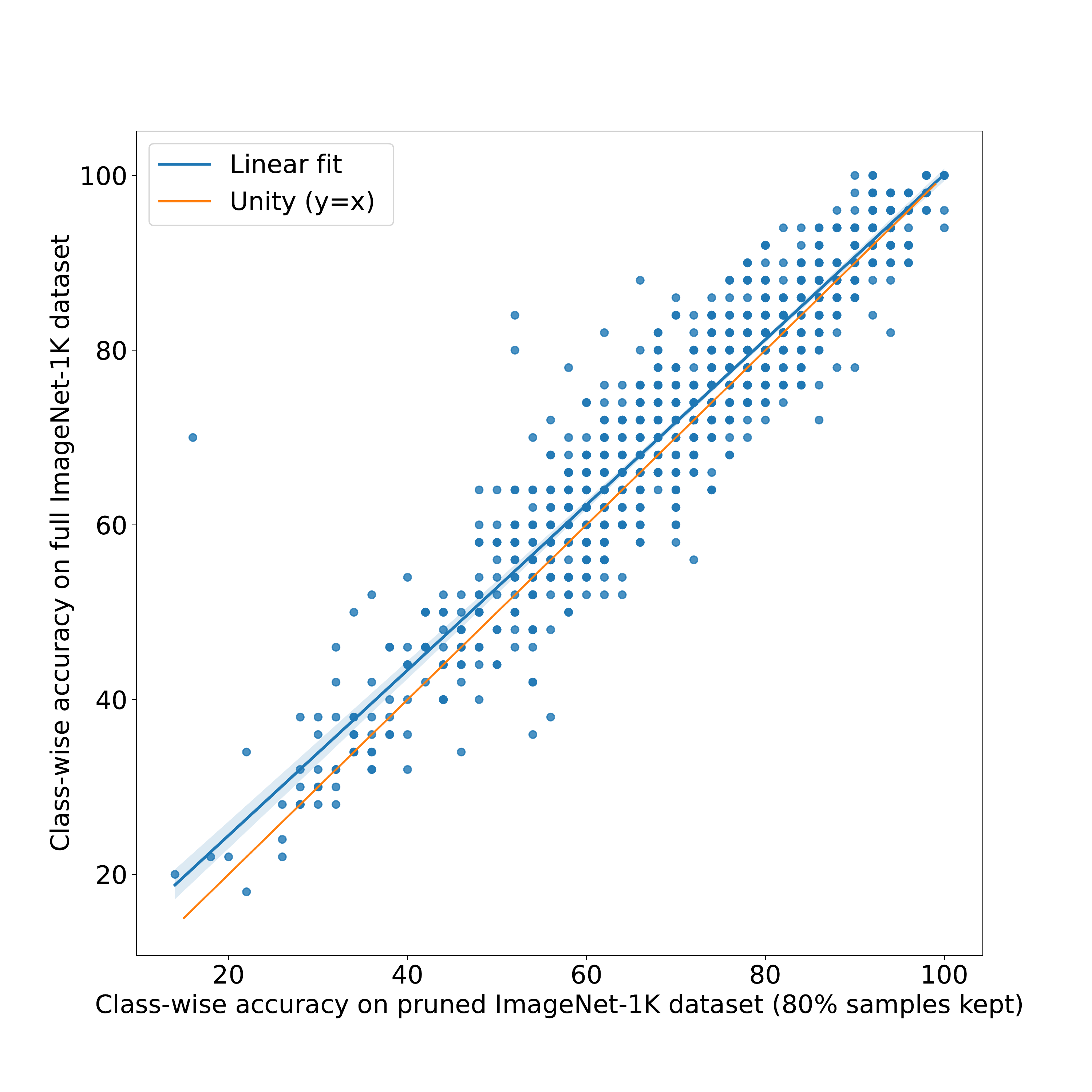}
			\caption{EL2N (20 models)}\label{fig:fairness_a}
	\end{subfigure}
	\begin{subfigure}{.33\linewidth}
			\centering
			\includegraphics[width=\linewidth]{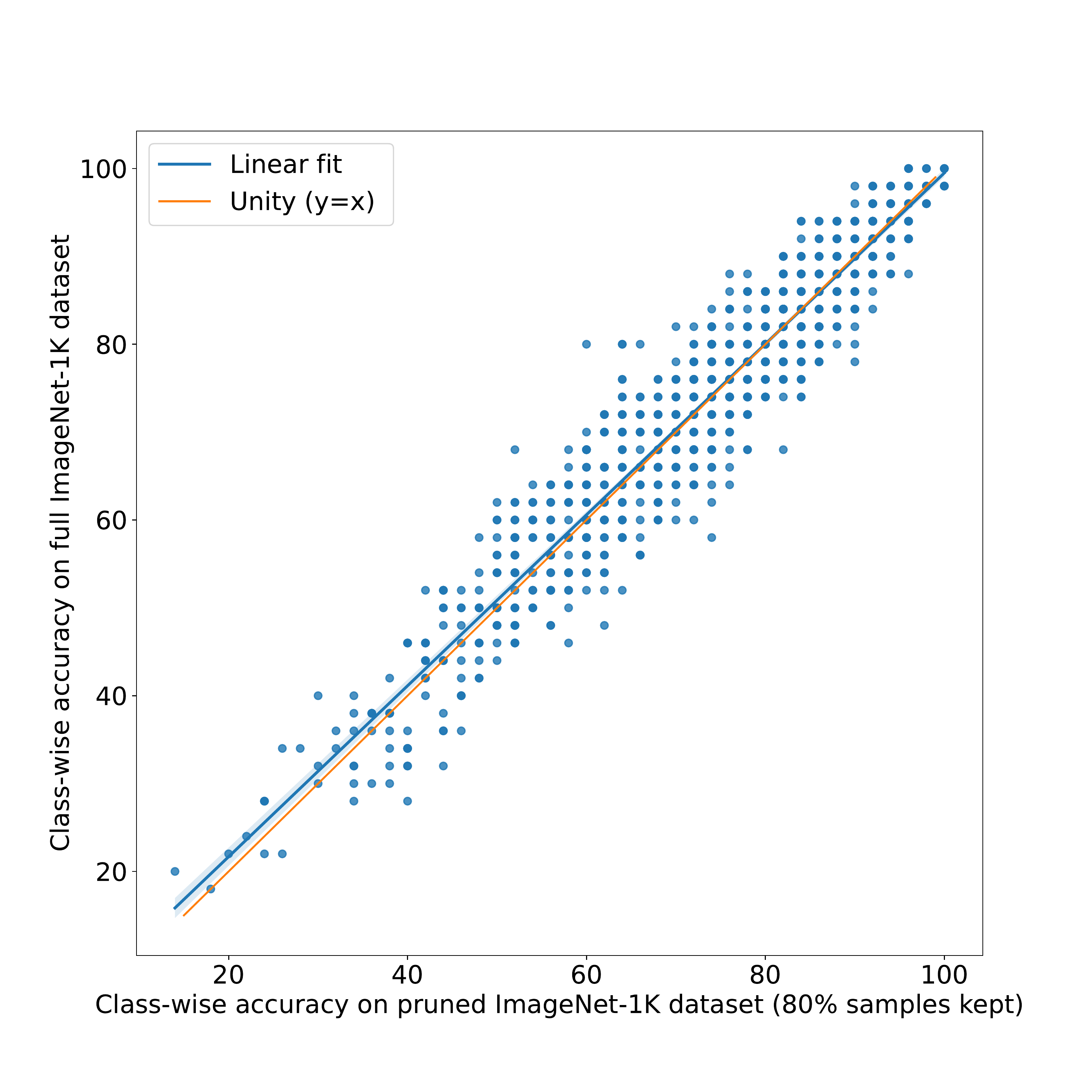}
			\caption{Memorization}\label{fig:fairness_b}
	\end{subfigure}
	\begin{subfigure}{.33\linewidth}
			\centering
			\includegraphics[width=\linewidth]{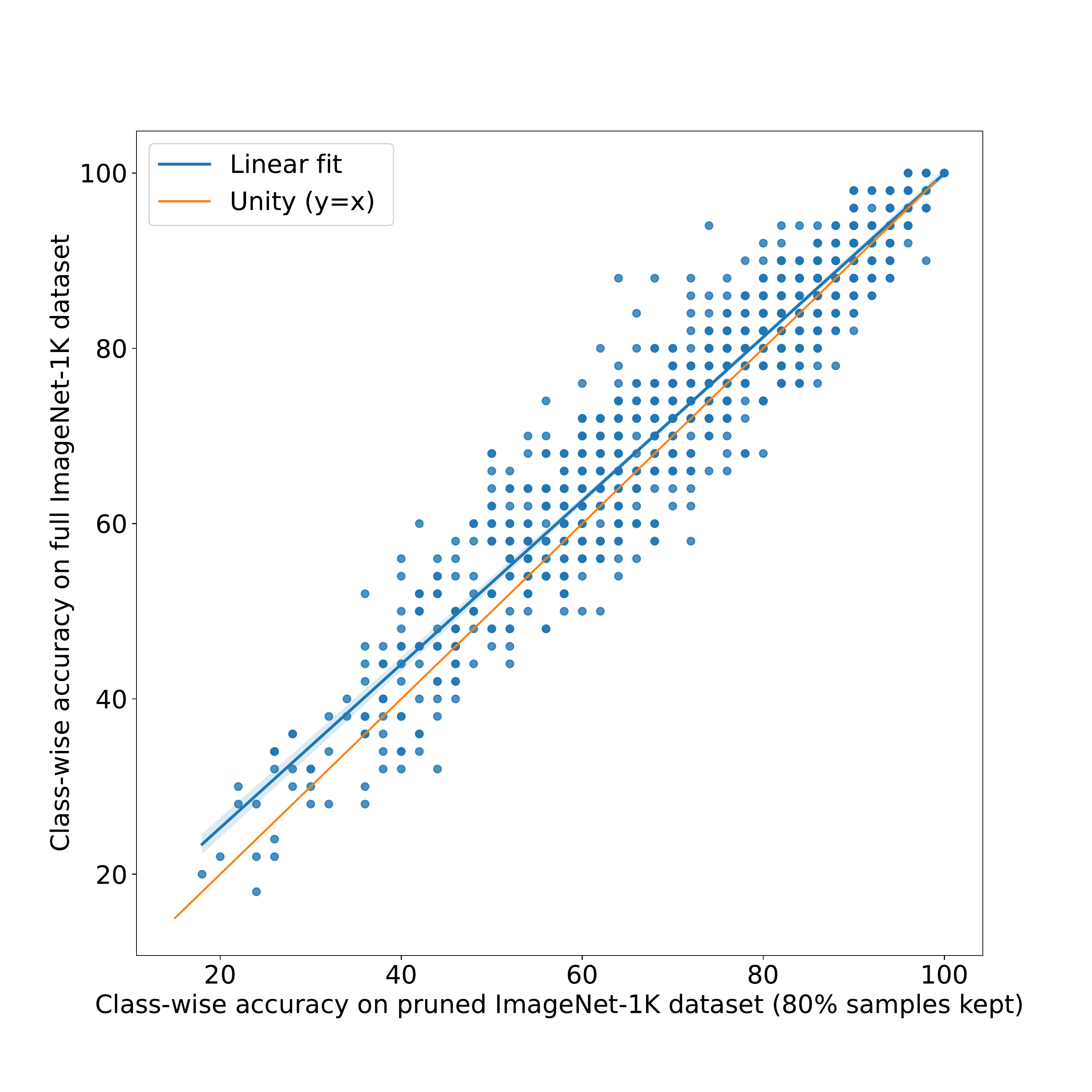}
			\caption{Self-supervised prototypes}\label{fig:fairness_c}
	\end{subfigure}
	\caption{The effect of data pruning on ImageNet test accuracy on individual classes. For all 3 plots, each point corresponds to an ImageNet class, and for all classes, the class specific test accuracy when training on the entire dataset (y-axis) is plotted against the test accuracy when training on the pruned dataset (x-axis).}
	\label{fig:fairness}
\end{figure*}

In order to study the effect of dataset pruning on model fairness, at least with respect to specific ImageNet classes, we compared the class-conditional accuracy of a ResNet-50 model trained on the full ImageNet dataset, versus that of the same model trained on an 80\% subset obtained after pruning. We used two supervised pruning metrics (EL2N, memorization) and one self-supervised pruning metric (self-supervised prototypes) for obtaining the pruned dataset. In all three cases, and across all $1000$ classes, we found that the class-conditional accuracy of the model trained on a pruned subset of the dataset remains quite similar to that of the model trained on the full dataset (Fig.~\ref{fig:fairness}). However, we did notice a very small reduction in class-conditioned accuracy for ImageNet classes that were least accurately predicted by models trained on the entire dataset (blue lines slightly above red unity lines when class-conditioned accuracy is low). This suggests that pruning yields a slight systematic reduction in the accuracy of harder classes, relative to easier classes, though the effect is small.  

While we have focused on fairness with respect to individual ImageNet classes, any ultimate test of model fairness should be conducted in scenarios that are specific to the use case of the deployed model. 
Our examination of the fairness of pruning with respect to individual ImageNet classes constitutes only an initial foray into an exploration of fairness, given the absence of any specific deployment scenario for our current models other than testing them on ImageNet.  We leave a full exploration of fairness in other deployment settings for future work. 

\section{Interaction between data pruning and training duration}
\label{app:interaction_duration}
Throughout the main paper, our ImageNet experiments are based on a setting where the number of training \emph{epochs} is kept constant (i.e.\ we train the same number of epochs on the smaller pruned dataset as on the larger original dataset). This means that data pruning directly reduces the number of \emph{iterations} required to train the model specifically by reducing the size of the dataset. However, this simultaneously places two constraints on model performance: not only training on a smaller data set, but also training for fewer iterations.

We therefore investigate how model performance changes if we train longer, as quantified by a \emph{matched iterations factor}. A matched iterations factor of $0$ corresponds to the default setting used in the paper of training for the same number of epochs (so that a smaller dataset means proportionally fewer training iterations). In contrast a matched iterations factor of $1$ corresponds to training on the smaller pruned dataset for a number of iterations equal to that when training on the larger initial dataset (e.g.\ when pruning away 50\% of the dataset one would train twice as long to match the number of iterations of a model trained on 100\% of the dataset).  Otherwise the matched iterations factor reflects a linear interpolation in the number of training iterations as the factor varies between $0$ and $1$. 

The results are shown in Table~\ref{tab:matched_iterations_factor} and indicate that training longer does indeed improve performance slightly; however, a matched iterations factor of around 0.4--0.6 may already be sufficient to reap the full benefit. Any matched iterations factor strictly smaller than 1.0 comes with reduced training time compared to training a model on the full dataset. 

\begin{table}[h]
\centering
\begin{tabular}{lrrrrrr}
\toprule
matched iterations factor & 0.0 & 0.2 & 0.4 & 0.6 & 0.8 & 1.0\\
\midrule
ImageNet top-5 accuracy & 90.20 & 90.41 & 90.48 & 90.56 & 90.45 & 90.50 \\
\bottomrule
\end{tabular}\\
\caption{Comparing different settings for training longer when pruning: Performance (top-5 acuracy) when pruning away 20\% of ImageNet based on our self-supervised prototype metric. Supervised model training on the pruned dataset performed with a ResNet-50 architecture trained using VISSL without class balancing. Performance tends to increase when training longer (i.e.\ with a larger matched iterations factor). Interestingly, the benefit of training longer may already be achieved with a matched iterations factor of around 0.4--0.6.}
\label{tab:matched_iterations_factor}
\end{table}

\section{Out-of-distribution (OOD) analysis of dataset pruning}
\label{app:OOD_analysis}
Pruning changes the data regime that a model is exposed to. Therefore, a natural question is how this might affect desirable properties beyond IID performance like fairness (see Appendix~\ref{app:fairness}) and out-of-distribution, or OOD, performance which we investigate here. To this end, we use the \href{https://github.com/bethgelab/model-vs-human}{model-vs-human} toolbox \cite{geirhos2021partial} based on data and analyses from \cite{wichmann2017methods,geirhos2018generalisation,geirhos2019imagenettrained,wang2019learning,geirhos2020beyond}. This toolbox is comprised of 17 different OOD datasets, including many image distortions and style changes.

\begin{figure}[h]
	\begin{subfigure}{0.49\linewidth}
		\centering
		\includegraphics[width=\linewidth]{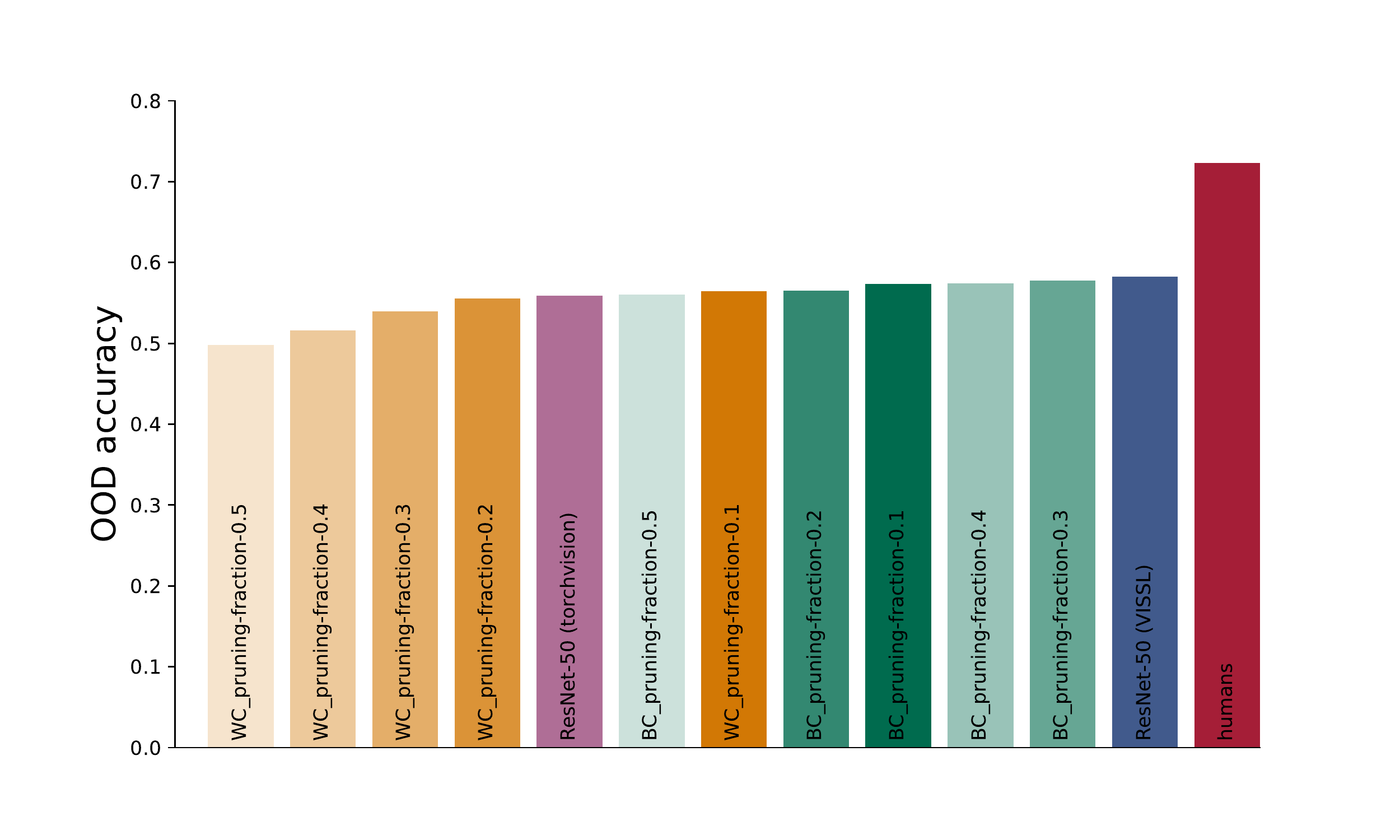}
		\caption{OOD accuracy (higher = better).}
		\label{subfig:benchmark_a}
	\end{subfigure}\hfill
	\begin{subfigure}{0.49\linewidth}
		\centering
		\includegraphics[width=\linewidth]{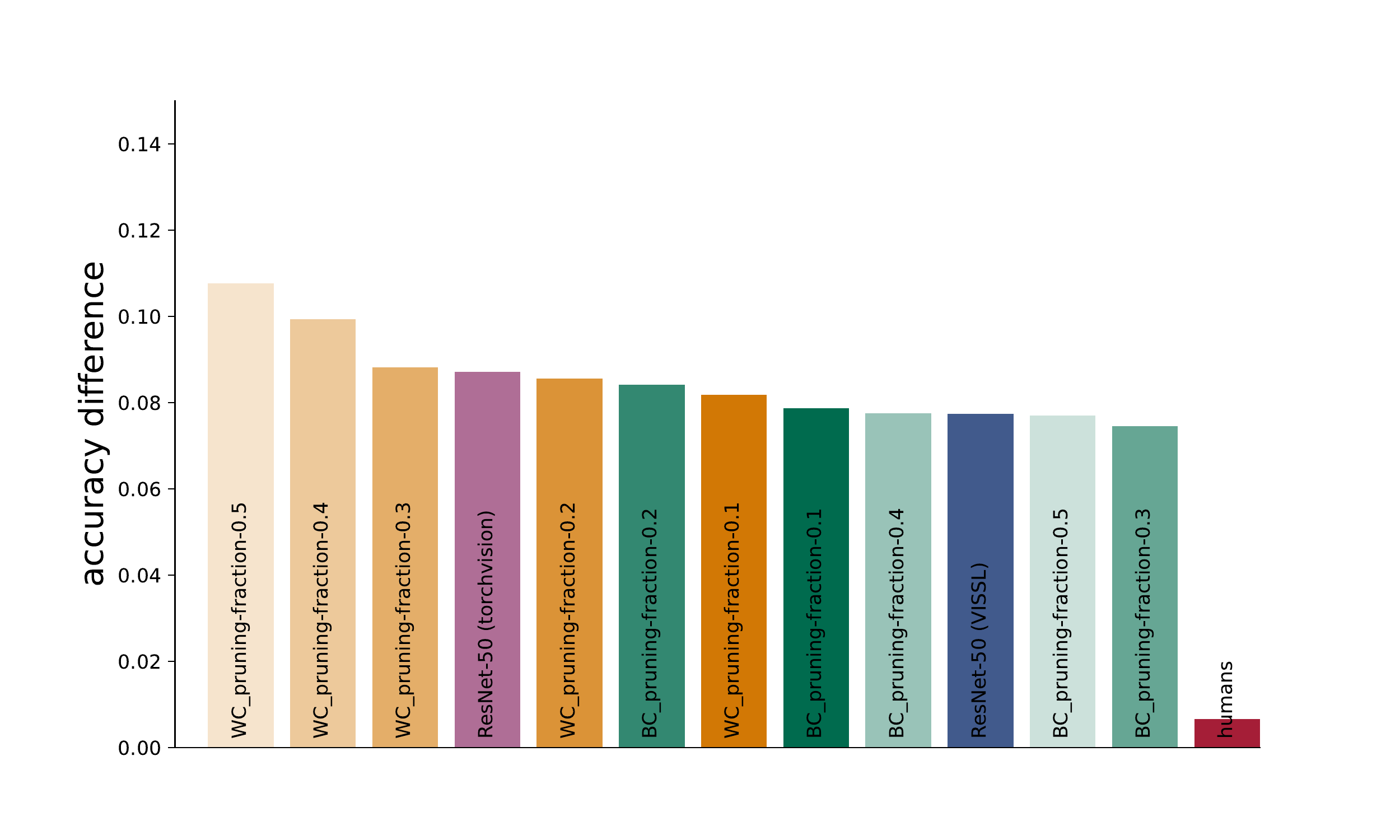}
		\caption{Accuracy difference (lower = better).}
		\label{subfig:benchmark_b}
	\end{subfigure}\hfill
	\begin{subfigure}{0.49\linewidth}
		\centering
		\includegraphics[width=\linewidth]{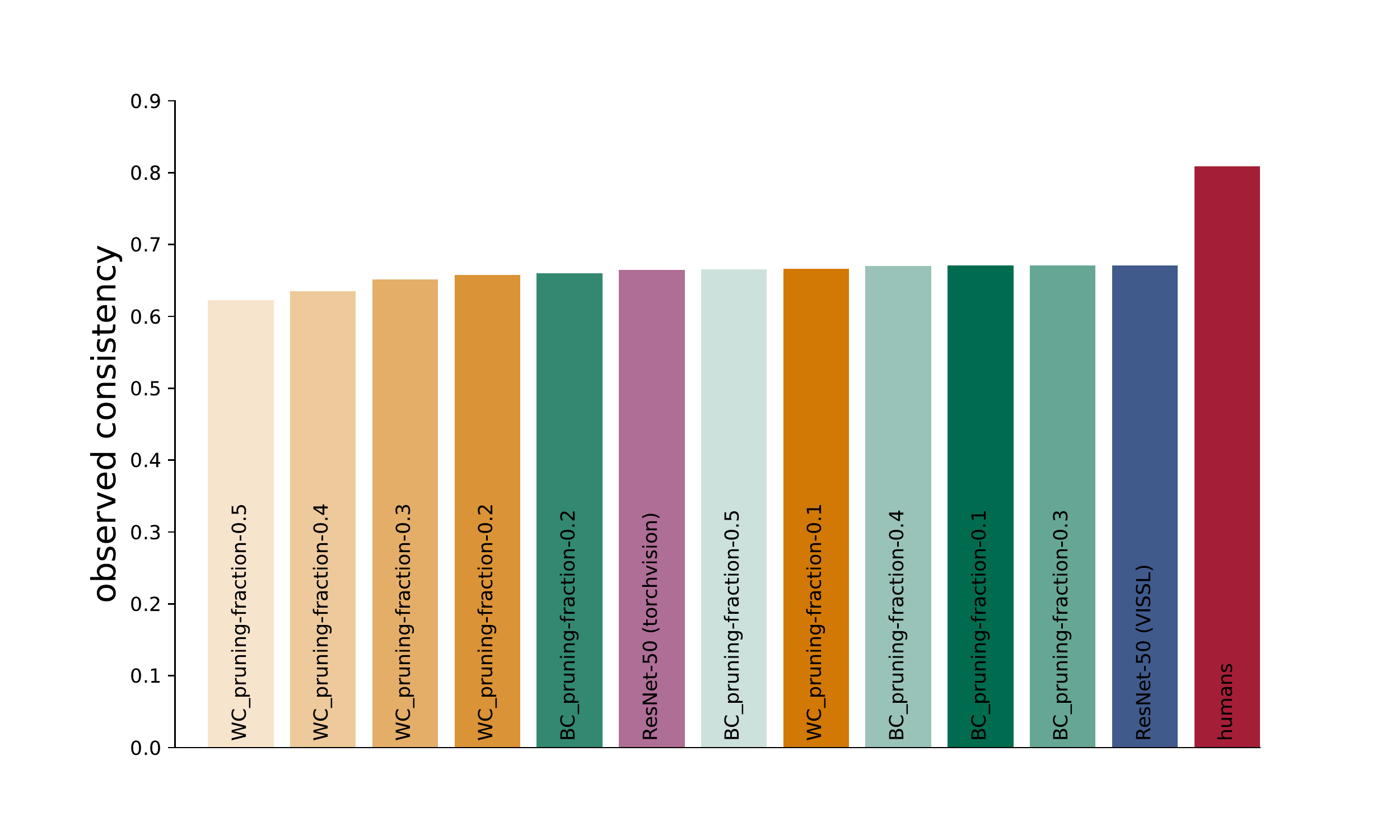}
		\caption{Observed consistency (higher = better).}			\label{subfig:benchmark_c}
	\end{subfigure}\hfill
	\begin{subfigure}{0.49\linewidth}
		\centering
		\includegraphics[width=\linewidth]{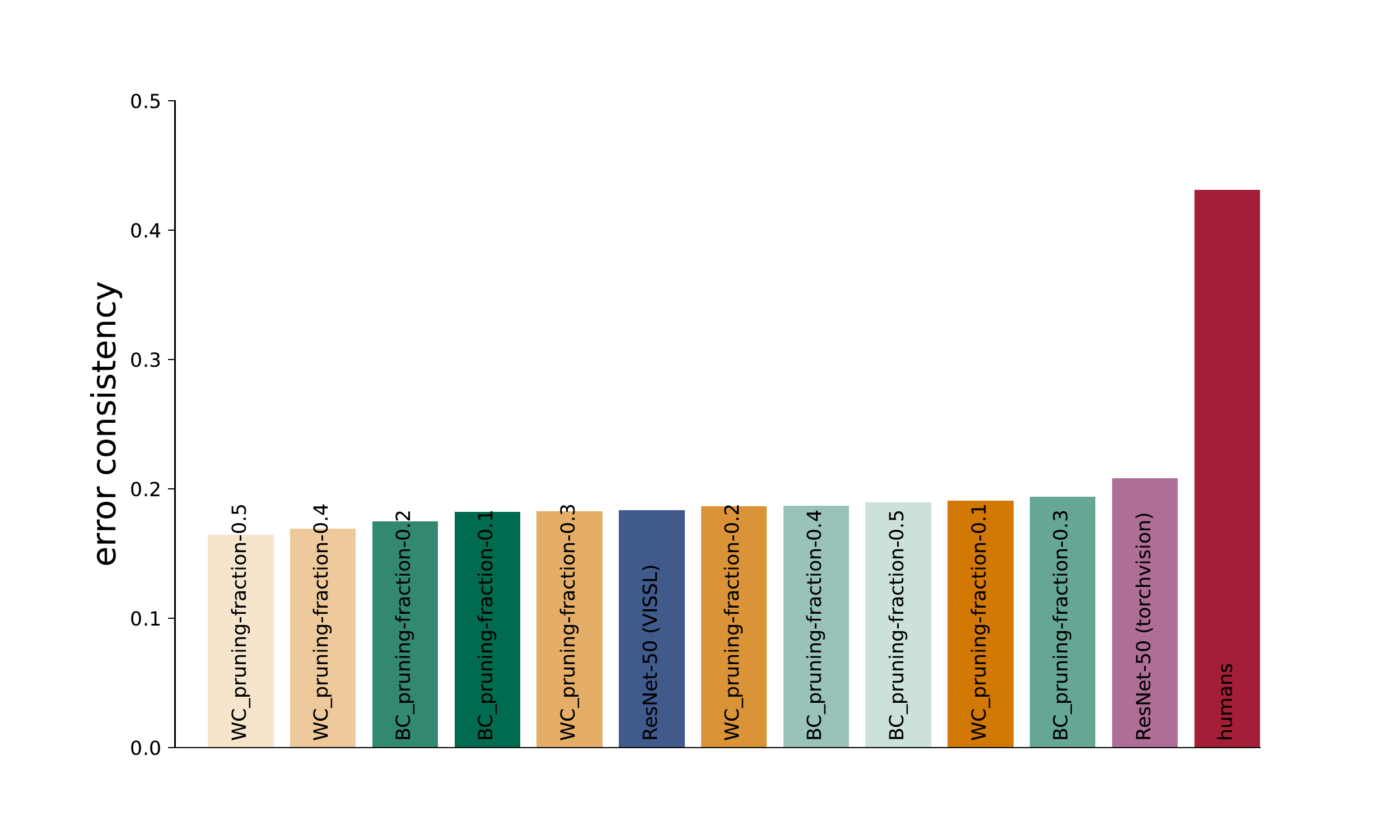}
		\caption{Error consistency (higher = better).}			\label{subfig:benchmark_d}
	\end{subfigure}\hfill
	\caption{OOD benchmark results for different models, aggregated over 17 datasets. All models have a ResNet-50 architecture \cite{he2015delving}. Two baseline models are trained on the full ImageNet training dataset, one using torchvision (purple) and the other using VISSL (blue). Human comparison data is shown in red.}
	\label{fig:benchmark_barplots}
\end{figure}

In Figure~\ref{subfig:benchmark_a}, OOD accuracies averaged across those 17 datasets are shown for a total of 12 models. These models all have a ResNet-50 architecture \cite{he2015delving}. Two baseline models are trained on the full ImageNet training dataset, one using torchvision (purple) and the other using VISSL (blue). Human classification data is shown as an additional reference in red. The remaining 10 models are VISSL-trained on pruned versions of ImageNet using pruning fractions in \{0.1, 0.2, 0.3, 0.4, 0.5\} and our self-supervised prototype metric. A pruning fraction of 0.3 would correspond to ``fraction of data kept = 0.7'', i.e.\ to training on 70\% of ImageNet while discarding the other 30\%. We investigated two different settings: discarding easy examples (the default used throughout the paper), which is denoted as ``Best Case'' (or BC) in the plots; and the reverse setting, i.e.\ discarding hard examples denoted as Worst Case, or WC. (These terms should be taken with a grain of salt; examples are only insofar best- or worst case examples as predicted by the metric, which itself is in all likelihood far less than perfect.)

The results are as follows: In terms of OOD accuracy (Figure~\ref{subfig:benchmark_a}), best-case pruning in green achieves very similar accuracies to the most relevant baseline, the blue ResNet-50 model trained via VISSL. This is interesting since oftentimes, OOD accuracies closely follow IID accuracies except for a constant offset \cite{miller2021accuracy}, and we know from Figure~\ref{fig:ImageNet-1K_panel} that the self-supervised prototype metric has a drastic performance impairment when pruning away 40\% of the data, yet the model ``BC\_pruning-fraction-0.4'' still achieves almost the same OOD accuracy as the baseline trained on the full dataset. \textbf{The core take-away is: While more analyses would be necessary to investigate whether pruning indeed consistently \emph{helps} on OOD performance, it seems safe to conclude that it does not \emph{hurt} OOD performance on the investigated datasets compared to an accuracy-matched baseline.} (The control setting, pruning away hard examples shown in orange, leads to much lower IID accuracies and consequently also lower OOD accuracies.) For reference, the numerical results from Figure~\ref{subfig:benchmark_a} are also shown in Table~\ref{tab:benchmark_table_accurate}.

Figures~\ref{subfig:benchmark_b}, \ref{subfig:benchmark_c} and \ref{subfig:benchmark_d} focus on a related question, the question of whether models show human-like behavior on OOD datasets. Figure~\ref{subfig:benchmark_b} shows that two models pruned using our self-supervised prototype metric somewhat more closely match human accuracies compared to the baseline in blue; Figures~\ref{subfig:benchmark_c} and \ref{subfig:benchmark_d} specifically focus on image-level consistency with human responses. In terms of overall consistency (c), the baseline scores best; in terms of error consistency pruned models outperform the VISSL-trained baseline. For details on the metrics we kindly refer the interested reader to \cite{geirhos2021partial}. Numerical results are again also shown in a Table (Table~\ref{tab:benchmark_table_humanlike}).

\begin{figure}[h]
	\includegraphics[width=\linewidth]{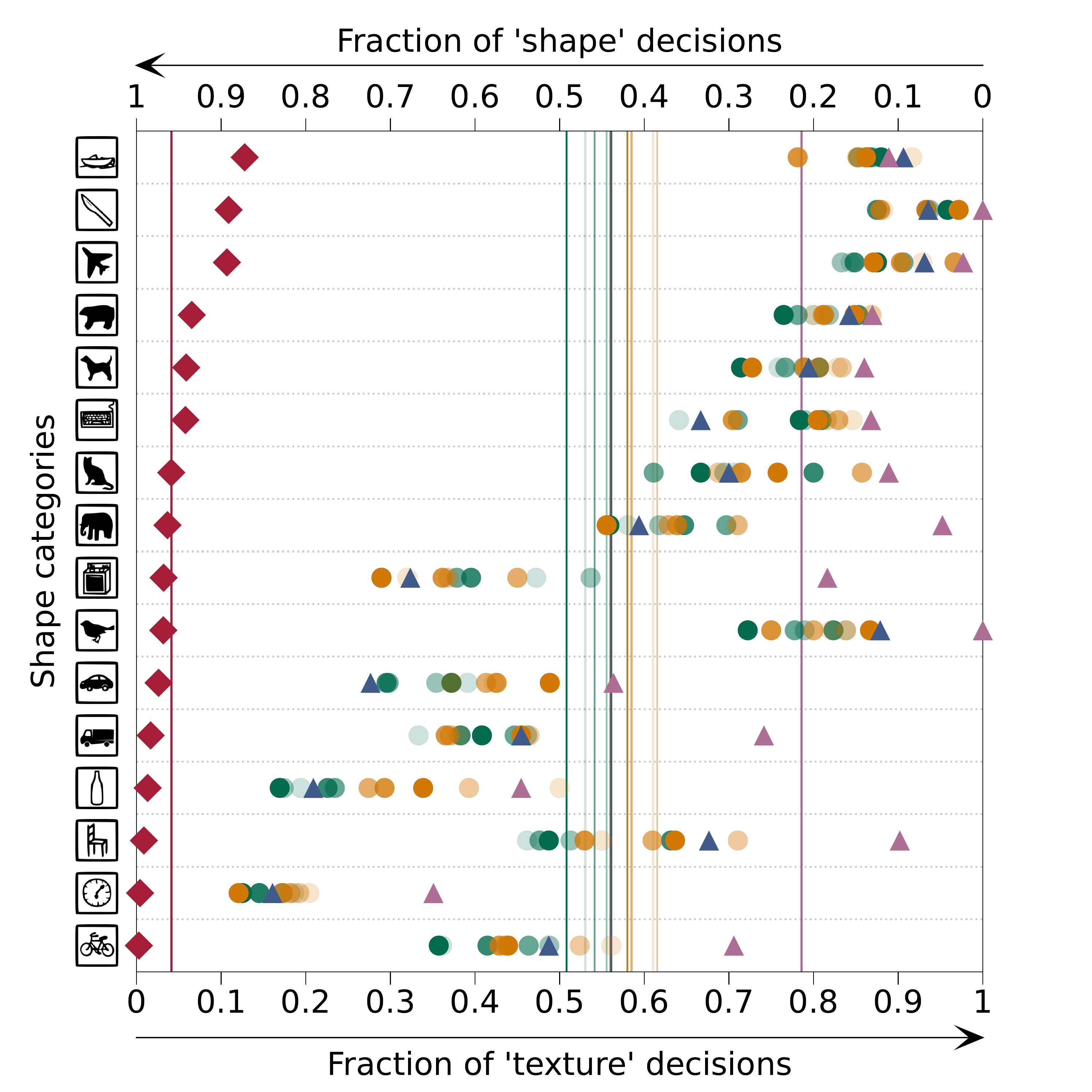}
	\caption{Shape vs.\ texture bias \cite{geirhos2019imagenettrained}: category-level plot. Horizontal lines indicate average shape/texture bias; values to the left lean towards a shape bias while values to the right lean towards a texture bias. For details on the plot see \cite{geirhos2019imagenettrained}; model colors are identical to Figure~\ref{fig:benchmark_barplots}.}
	\label{fig:texture_shape_bias}
\end{figure}

Finally, in Figure~\ref{fig:texture_shape_bias} we observe that best-case pruning leads to slightly higher shape bias as indicated by green vertical lines plotting the average shape bias across categories, which are shifted to the left of the baseline; while worst-case pruning in orange is shifted to the right. An outlier is the torchvision-trained model in purple with a very strong texture bias; we attribute this to data augmentation differences between VISSL and torchvision training.

\newcommand{\figwidth}{0.24\textwidth}
\newcommand{\captionspace}{-1.5\baselineskip}
\newcommand{\captionspaceII}{0.6\baselineskip}
\newcommand{\captionspaceBenchmark}{-0.5\baselineskip}

\begin{table}[h!]
	\caption{Benchmark table of model results for highest out-of-distribution robustness.}
	\label{tab:benchmark_table_accurate}
	\centering
	\begin{tabular}{lrr}
\toprule
                   model & OOD accuracy $\uparrow$ & rank $\downarrow$ \\
\midrule
       ResNet-50 (VISSL) &          \textbf{0.582} &    \textbf{1.000} \\
BC\_pruning-fraction-0.3 &                   0.578 &             2.000 \\
BC\_pruning-fraction-0.4 &                   0.574 &             3.000 \\
BC\_pruning-fraction-0.1 &                   0.574 &             4.000 \\
BC\_pruning-fraction-0.2 &                   0.565 &             5.000 \\
WC\_pruning-fraction-0.1 &                   0.565 &             6.000 \\
BC\_pruning-fraction-0.5 &                   0.560 &             7.000 \\
 ResNet-50 (torchvision) &                   0.559 &             8.000 \\
WC\_pruning-fraction-0.2 &                   0.556 &             9.000 \\
WC\_pruning-fraction-0.3 &                   0.540 &            10.000 \\
WC\_pruning-fraction-0.4 &                   0.516 &            11.000 \\
WC\_pruning-fraction-0.5 &                   0.498 &            12.000 \\
\bottomrule
\end{tabular}
\end{table}

\begin{table}[ht]
	\caption{Benchmark table of model results for most human-like behaviour. The three metrics ``accuracy difference'' ``observed consistency'' and ``error consistency'' (plotted in Figure~\ref{fig:benchmark_barplots}) each produce a different model ranking.}
	\label{tab:benchmark_table_humanlike}
	\centering
	\begin{tabular}{lrrrr}
\toprule
                   model & accuracy diff. $\downarrow$ & obs. consistency $\uparrow$ & error consistency $\uparrow$ & mean rank $\downarrow$ \\
\midrule
BC\_pruning-fraction-0.3 &              \textbf{0.075} &                       0.671 &                        0.194 &         \textbf{1.667} \\
       ResNet-50 (VISSL) &                       0.077 &              \textbf{0.671} &                        0.184 &                  3.667 \\
BC\_pruning-fraction-0.5 &                       0.077 &                       0.666 &                        0.190 &                  4.000 \\
BC\_pruning-fraction-0.4 &                       0.078 &                       0.670 &                        0.187 &                  4.333 \\
WC\_pruning-fraction-0.1 &                       0.082 &                       0.666 &                        0.191 &                  4.667 \\
 ResNet-50 (torchvision) &                       0.087 &                       0.665 &               \textbf{0.208} &                  5.667 \\
BC\_pruning-fraction-0.1 &                       0.079 &                       0.671 &                        0.182 &                  5.667 \\
WC\_pruning-fraction-0.2 &                       0.086 &                       0.658 &                        0.186 &                  7.667 \\
BC\_pruning-fraction-0.2 &                       0.084 &                       0.660 &                        0.175 &                  8.333 \\
WC\_pruning-fraction-0.3 &                       0.088 &                       0.652 &                        0.183 &                  9.333 \\
WC\_pruning-fraction-0.4 &                       0.099 &                       0.635 &                        0.169 &                 11.000 \\
WC\_pruning-fraction-0.5 &                       0.108 &                       0.623 &                        0.165 &                 12.000 \\
\bottomrule
\end{tabular}
\end{table}

\hfill

\end{document}